\documentclass{article}

\usepackage{microtype}
\usepackage{graphicx}
\usepackage{subfigure}
\usepackage{booktabs}
\usepackage[hide=true]{marginalia}
\usepackage{enumitem}
\usepackage{amsmath,amsfonts,bm}
\usepackage{ amssymb }
\usepackage{cancel}

\DeclareMathOperator*{\argminB}{argmin}

\def\eqref#1{eq.~\ref{#1}}
\def\1{\bm{1}}

\def\vx{{\bm{x}}}

\DeclareMathAlphabet{\mathsfit}{\encodingdefault}{\sfdefault}{m}{sl}
\SetMathAlphabet{\mathsfit}{bold}{\encodingdefault}{\sfdefault}{bx}{n}

\def\sD{{\mathbb{D}}}
\usepackage{changepage}
\usepackage{selectp}
\usepackage{comment}
\usepackage{algorithm}
\usepackage[noend]{algpseudocode}
\usepackage{transparent}

\usepackage{hyperref}

\usepackage[accepted]{icml2021}

\icmltitlerunning{On the Predictability of Pruning Across Scales}

\begin{document}

\twocolumn[
\icmltitle{On the Predictability of Pruning Across Scales}



\icmlsetsymbol{equal}{*}

\begin{icmlauthorlist}
\icmlauthor{Jonathan Rosenfeld}{mit}
\icmlauthor{Jonathan Frankle}{mit}
\icmlauthor{Michael Carbin}{mit}
\icmlauthor{Nir Shavit}{mit}
\end{icmlauthorlist}

\icmlaffiliation{mit}{MIT CSAIL}

\icmlcorrespondingauthor{Jonathan Rosenfeld}{jonsr@csail.mit.edu}

\icmlkeywords{}

\vskip 0.3in
]



\printAffiliationsAndNotice{}  % leave blank if no need to mention equal contribution

\begin{abstract}
We show that the error of iteratively magnitude-pruned networks empirically follows a scaling law with interpretable coefficients that depend on the architecture and task.
We functionally approximate the error of the pruned networks, showing it is predictable in terms of an invariant tying width, depth, and pruning level, such that networks of vastly different pruned densities are interchangeable.
We demonstrate the accuracy of this approximation over orders of magnitude in depth, width, dataset size, and density.
We show that the functional form holds (generalizes) for large scale data (e.g., ImageNet) and architectures (e.g., ResNets).
As neural networks become ever larger and costlier to train, our findings suggest a framework for reasoning conceptually and analytically about a standard method for unstructured pruning.
\end{abstract}

\section{Introduction}

For decades, neural network \emph{pruning}---eliminating unwanted parts of a network---has been a popular approach for reducing network sizes or the computational demands of inference \citep{optimalbraindamage,reed,han}.
In practice, pruning can reduce the parameter-counts of contemporary models by 2x \citep{bertpruning} to 5x \citep{Renda2020Comparing} with no increase in error.
More than 80 pruning techniques have been published in the past decade \citep{blalock2020state}, but, despite this enormous volume of research, there remains little guidance on important aspects of pruning.
Consider a seemingly simple question one might ask when using a particular pruning technique:

\textit{Given a family of neural networks (e.g., ResNets on ImageNet of various widths and depths), which family member should we prune (and by how much) to obtain the network with the smallest parameter-count such that error does not exceed some threshold $\epsilon_k$?}

As a first try, we could attempt to answer this question using brute force: %\fTBD{JSR: does this sound like a weak motivation for someone knowing of song? or is it fine since we contrast to him later. JEF: I think it's fine. We cite him soon. }
we could prune every member of a network family (i.e., perform grid search over widths, depth, and pruned densities) and select the smallest pruned network that satisfies our constraint on error.
However, depending on the technique, pruning one network (let alone grid searching) could take days or weeks on expensive hardware.

If we want a more efficient alternative, we will need to make assumptions about pruned networks:
namely, that there is some \emph{structure} to the way that their error behaves.
For example, that pruning a particular network changes the error in a predictable way.
Or that changing its width or depth changes the error when pruning it in a predictable way.
We could then train a smaller number of networks, characterize this structure, and estimate the answer to our question.

We have reason to believe that such structure does exist for certain pruning methods, since there are already techniques that take advantage of it implicitly.
For example, \citet{cai2019once} create a single neural network architecture that can be scaled down to many different sizes;
to choose which subnetwork to deploy, \citeauthor{cai2019once} train an auxiliary, black-box neural network to predict subnetwork performance.
Although this black-box approach implies the existence of structure for this pruning method, it does not reveal this structure explicitly or make it possible to reason analytically in a fashion that could answer our research question.

% The drawbacks of this black-box approach are that (1) it is specific to a particular network and pruning technique and (2) although it implies the existence of structure, it does not 
% \hl{Critically, such a black box approach while very useful in and of itself, is network specific, and does not directly reveal insight into the dependency of the generalization error on different design trade-offs such as network width or depth or level of pruning, or to other pruning algorithms or network types}

For other aspects of deep learning beyond pruning, such structure has been observed---and, further, codified explicitly---yielding insights and predictions in the form of scaling laws.
% Structure has also been observed---and, even further, codified explicitly---for other aspects of deep learning.
\citet{tan2019efficientnet} design the \emph{EfficientNet} family by developing a heuristic for predicting efficient tradeoffs between depth, width, and resolution.
\citet{hestness2017deep} observe a power-law relationship between dataset size and the error of vision and NLP models.
\citet{rosenfeld2020a} use a power law to predict the error of all variations of architecture families and dataset sizes, jointly, for computer vision and natural language processing settings.
\citet{kaplan2020scaling} develop a similar power law for language models that incorporates the computational cost of training.

Inspired by this work, we address our research question about pruning by finding a scaling law to predict the error of pruned networks.
We focus on a pruning method called \emph{iterative magnitude pruning (IMP)}, where weights with the lowest magnitudes are pruned in an unstructured fashion interspersed with re-training to recover accuracy.
This method is a standard way to prune \citep{han} that gets state-of-the-art tradeoffs between error and unstructured density \citep{gale, Renda2020Comparing}.
To the best of our knowledge, this is the first explicit scaling law that holds for pruned networks, let alone entire network families.

To formulate such a predictive scaling law, we consider the dependence of generalization error on the pruning-induced \emph{density} for networks of different depths and widths trained on different dataset sizes.
We begin by developing a functional form that accurately estimates the generalization error of a specific model as it is pruned (Section \ref{sec:single-network}).
%When pruning a network, error behaves in a materially different fashion than when varying its width or depth without pruning; we therefore choose a functional form that is materially different from those used in prior work.\fTBD{Repetitive with what I've added to the previous paragraph}
We then account for other architectural degrees of freedom, expanding the functional form for pruning into a scaling law that jointly considers density alongside width, depth, and dataset size (Section \ref{sec:joint}).
The basis for this joint scaling law is an \emph{invariant} we uncover that describes ways that we can interchange depth, width, and pruning without affecting error.
The result is a scaling law that accurately predicts the error of pruned networks across scales.
And, now that we have established this functional form, fitting it requires only a small amount of data (Section \ref{app:interpolation}), making it efficient to use on new architectures and datasets (Appendix \ref{app:more_arch_alg}).
Finally, we use this scaling law to answer our motivating question (Section \ref{sec:conclusions}).
%The same functional form can accurately estimate the error for both unstructured magnitude pruning \citep{Renda2020Comparing} and SynFlow \citep{tanaka2020pruning} when fit to the corresponding data, suggesting we have uncovered structure that may be applicable to iterative pruning more generally.
In summary, our contributions are as follows:
\begin{itemize}[leftmargin=1.7em,itemsep=0em,topsep=0pt,parsep=1.5pt,partopsep=0em]
    \item We develop a scaling law that accurately estimates the error when pruning a single network with IMP.
    \item We observe and characterize an \emph{invariant} that allows error-preserving interchangeability among depth, width, and pruning density.
    \item Using this invariant, we extend our single-network scaling law into a joint scaling law that predicts the error of all members of a network family at all dataset sizes and all pruning densities.
    \item In doing so, we demonstrate that there is structure to the behavior of the error of iteratively magnitude-pruned networks that we can capture explicitly with a simple functional form and interpretable parameters.
    \item Our scaling law enables a framework for reasoning analytically about IMP, allowing us to answer our motivating question and similar questions about pruning.\fTBD{Say something forward-looking here about the implications for other pruning methods and for applying this in new contexts? JR A: prob not - it is inviting the fire}
\end{itemize}

% This framework provides a compact approach for answering our motivating question and similar design tradeoff questions. 
% Armed with the functional form, which depends on few (5) interpretable parameters, we can analytically reason about tradeoffs. Finally, such a high-agreement scaling law my prove usful in the persuit of a theoretical understanding of the origin of pruning performance and limitations.  

% This framework provides an elegant approach for answering our motivating question and other similar questions: with our functional form for pruning in hand, finding answers becomes an analytical exercise on an optimization problem.\jef{Need to clarify this}
% , and its only required resources are pen and paper.\fTBD{JEF: This is really repetitive with the last bullet point. Replace with:
% The question is cast into an analytical optimization problem which one can directly solve with the benefit of attaining both the sought solution and additional insight.
% ``we can answer our motivating question by analytically solving optimization problems using our scaling law.''}

\section{Experimental Setup}
\label{sec:setup}

\textbf{Pruning.}
We study \emph{iterative magnitude pruning} (IMP) \citep{jankowsky, han}.
IMP prunes by removing a fraction---typically 20\%, as we do here---of individual weights with the lowest magnitudes in an unstructured fashion at the end of training.%
\footnote{We do not prune biases or BatchNorm, so pruning 20\% of weights prunes fewer than 20\% of parameters.}
We choose these weights globally throughout the network, i.e., without regard to specific layers.
We use per-weight magnitude pruning because it is generic, well-studied \citep{han}, and produces state-of-the-art tradeoffs between density and error \citep{gale, blalock2020state, Renda2020Comparing}.

Pruning weights typically increases the error of the trained network, so it is standard practice to further train after pruning to reduce error.
For IMP, we use a practice called \emph{weight rewinding} \citep{frankle2020linear, Renda2020Comparing}, in which the values of unpruned weights are \emph{rewound} to their values earlier in training (in our case, epoch 10) and the training process is repeated from there to completion.
To achieve density levels below 80\%, this process is repeated \emph{iteratively}---pruning by 20\%, rewinding, and retraining---until a desired density level is reached.
For a formal statement of this pruning algorithm, see Appendix \ref{app:pruningalg}.

\textbf{Datasets.}
In the main body of the paper, we study the image classification tasks CIFAR-10 and ImageNet.
Our scaling law predicts the error when training with the entire dataset and smaller \emph{subsamples}.
We include subsampling because it provides a cost-effective way to collect some of the data for fitting our functional form.
To subsample a dataset to a size of $n$, we randomly select $n$ of the training examples without regard to individual classes such that in expectation we preserve the original dataset distribution (we always retain the entire test set).
When performing iterative pruning, we maintain the same subsample for all pruning iterations.
We consider other datasets in Appendix \ref{app:more_arch_alg}.

\textbf{Networks.}
In the main body of the paper, we study ResNets for CIFAR-10 and ImageNet.\footnote{See Appendix \ref{app:resnets} for full details on architectures and hyperparameters. Note that CIFAR-10 and ImageNet ResNets have different architectures \citep{he2016deep}.}
We develop a scaling law that predicts the error (when pruned) of an entire \emph{family} of networks with varying widths and---in the case of the CIFAR-10 ResNets---depths.
To vary width, we multiply the number of channels in each layer by a \emph{width scaling factor}.
To vary depth of the CIFAR-10 ResNets, we vary the number of residual blocks.
We refer to a network by its depth $l$ (the number of layers in the network, not counting skip connections) and its width scaling factor $w$.
We consider other networks in Appendix \ref{app:more_arch_alg}.

\begin{table*}
\begin{center}
{\scriptsize
%\begin{tabular}{@{\ }l@{\ }|@{\ }c@{\ \ }c@{\ }|@{\ }c@{\ }|@{\ }c@{\ }|@{\ }c@{\ }|@{\ }c}
\begin{tabular}{@{\ }l@{\ }|@{\ }c@{\ }|@{\ }c@{\ }|@{\ }c@{\ }|@{\ }c}
\toprule
Network Family &
    %$N_{\text{train}}$ &
    %$N_{\text{test}}$ &
    Densities ($d$) &
    Depths ($l$) &
    Width Scalings ($w$) &
    Subsample Sizes ($n$) \\ \midrule
CIFAR-10 ResNet &
    %50K &
    %10K &
    $0.8^i, i \subseteq \{0, \ldots, 40\}$ (see below) &
    $l \subseteq$ \{8, 14, 20, 26, 50, 98\} (see below) &
    $2^i, i \subseteq \{-4, \ldots, 2\}$ (see below) &
    $\frac{N}{i}$, $i \in \{1, 2, 4, 8, 16, 32, 64\}$ \\
ImageNet ResNet &
    %1.28M &
    %50K &
    $0.8^i, i \subseteq \{0, \ldots, 30\}$ (see below) &
    50 & 
    $2^i, i \subseteq \{-4, \ldots, 0\}$ (see below) &
    $\frac{N}{i}$, $i \in \{1, 2, 4\}$ \\ 
\bottomrule
\end{tabular}
}
\end{center}
\vspace{-4.5mm}
\label{tab:dims}
\caption{The ranges of settings we consider in our experiments in the main body of the paper. We consider all densities $d \in \{0.8^i~|~i \in \mathbb{N}_0\}$ where (1) the network is not disconnected and (2) the network does better than random chance; this value varies based on the configuration $(l, w, n)$ and the random seed. Where neither of these conditions apply, we cap $i$ at 40 (CIFAR-10 ResNets) or 30 (ImageNet ResNets).
We consider all configurations of $(l, w, n)$ before which increasing depth or width of the unpruned network increases test error; configurations beyond this point are overly large for a given dataset subsample.
By this criterion, we use 152 configurations of $(l, w, n)$ for the CIFAR-10 ResNets and 15 for the ImageNet ResNets.
Taking into account all feasible densities, we use a total of 4,301 CIFAR-10 ResNet points and 274 ImageNet ResNet points.
Note that we use these configurations to find and evaluate the functional form. Once we do so, far fewer configurations are needed to fit the functional form for each setting (see Section \ref{app:interpolation}).}
\vspace{-4.5mm}
\end{table*}

\textbf{Notation and terminology.} Throughout the paper, we use the following notation and terminology:
\begin{itemize}[leftmargin=1.3em, topsep=0pt, itemsep=1pt, parsep=0.5pt, partopsep=-1pt]
    \item $\sD_N = \{  \vx_i,y_i \}_{i=1}^{N}$ is a labeled training set with $N$ examples. A \emph{subsample} of size $n$ is a subset of $\sD_N$ with $n$ examples selected uniformly at random.
    \item $l$ and $w$ are, respectively, the depth (i.e., the number of layers, excluding skip connections) and the width scaling factor of a particular network.
    \item Networks that vary by width and depth are a \emph{family}.
    \item $d$ is the \emph{density} of a pruned network (i.e., the fraction of weights that have not been pruned).
    \item $\epsilon \left(d, l, w, n\right)$ is the test error of a network with the specified density, depth, width scaling, and dataset size.
    \item $\epsilon_{np}\left(l, w, n\right) = \epsilon \left(1, l, w, n\right)$ is the test error of the unpruned network with the specified depth, width scaling, and dataset size. When clear from context, we omit $(w,l,n)$ and write $\epsilon_{np}$.
    \item $\hat{\epsilon}(\epsilon_{np}, d \mid l, w, n)$ is an estimate of the error of a pruned model for a scaling law that has been fitted to a specific network with the specified depth, width scaling, and dataset size (Section \ref{sec:single-network}).
    %    denotes the conditional form of our estimate. In contrast to $\hat\epsilon \left(\epsilon_{np}, d, l, w, n\right)$, it allows any parameters within the estimate's definition to be functions of $w$, $l$, and $n$.
    \item $\hat\epsilon \left(\epsilon_{np}, d, l, w, n\right)$ is an estimate of the error of a pruned model with the specified depth, width scaling, and dataset size for a scaling law that has been fitted to a network family (Section \ref{sec:joint}).
    %is an estimate of the error for specified unpruned error and $d$, $l$, $w$, and $n$. 
    % Note the implicit dependency on $l$, $w$, and $n$ in $\epsilon_{np}\left(l, w, n\right) $
\end{itemize}
\vspace{-5pt}

\textbf{Dimensions.}
In developing our scaling laws, we vary four dimensions: dataset subsample size ($n$)
and network degrees of freedom density ($d$), network depth ($l$), and width scaling factor ($w$).
In the main body of the paper, we consider the ranges of these values as specified in Table 1.
See the caption in Table 1 for full details.
We train three replicates of each CIFAR-10 configuration with different seeds.

\section{Modeling the Error of a Pruned Network}
\label{sec:single-network}

Our goal in this section is to develop a functional form that models the error of a member of a network family as it is pruned (using IMP) based on its unpruned error $\smash{\epsilon_{np}(w,l,n)}$.
In other words, we wish to find a function $\smash{\hat\epsilon(\epsilon_{np}, d~|~l, w, n)}$ that predicts the error at each density $d$ for a network with a specific depth $l$, width scaling factor $w$, and dataset size $n$.

\begin{figure*}
\vspace{-1mm}
\centering
\begin{minipage}{0.28\textwidth}
\includegraphics[width=\linewidth,trim={0.2cm 0 0.1cm 0.65cm},clip]{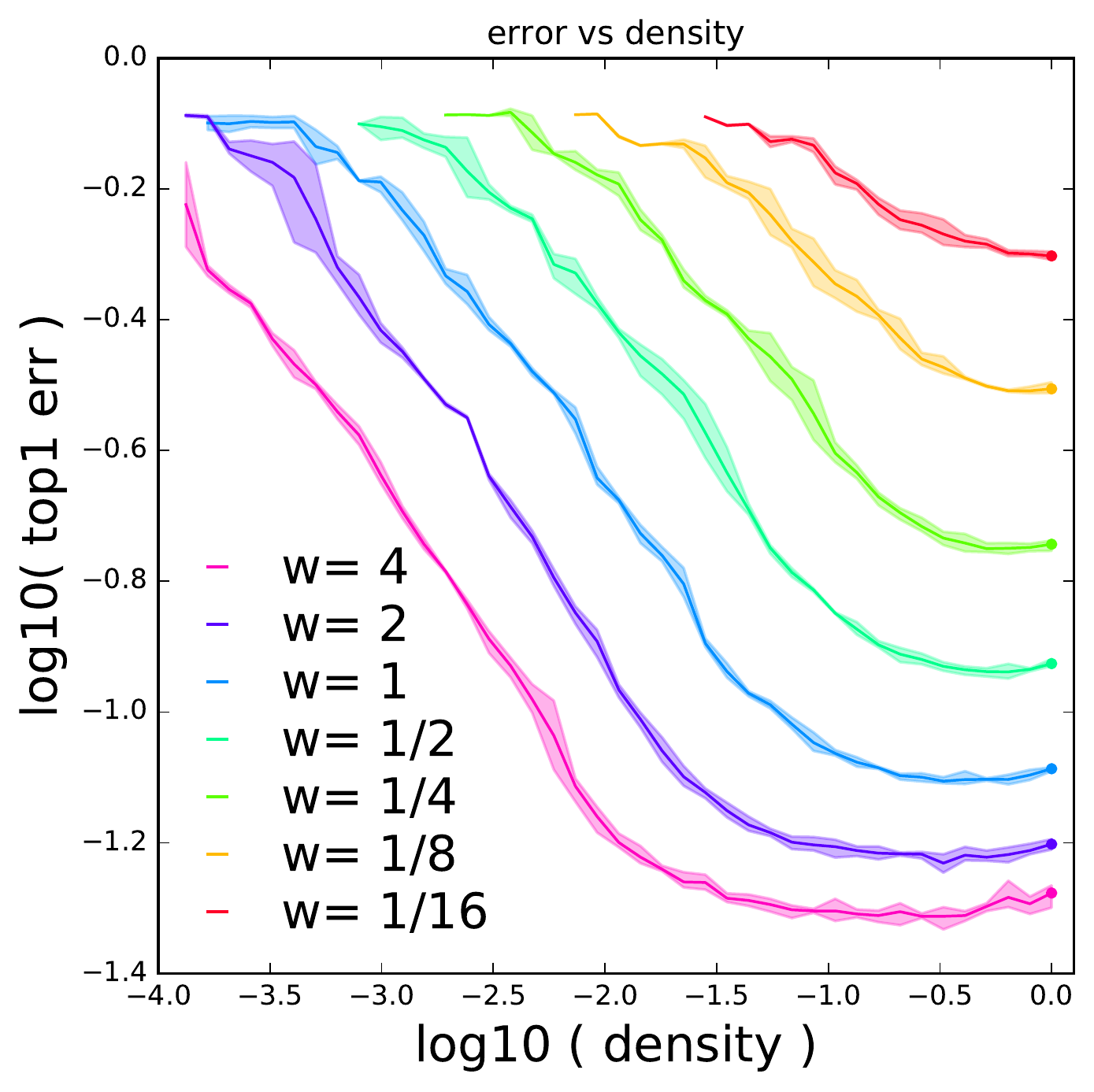}%
\end{minipage}\hfil
\begin{minipage}{0.28\textwidth}
\includegraphics[width=\linewidth,trim={10.8cm 2.1cm 10.9cm 5.65cm},clip]{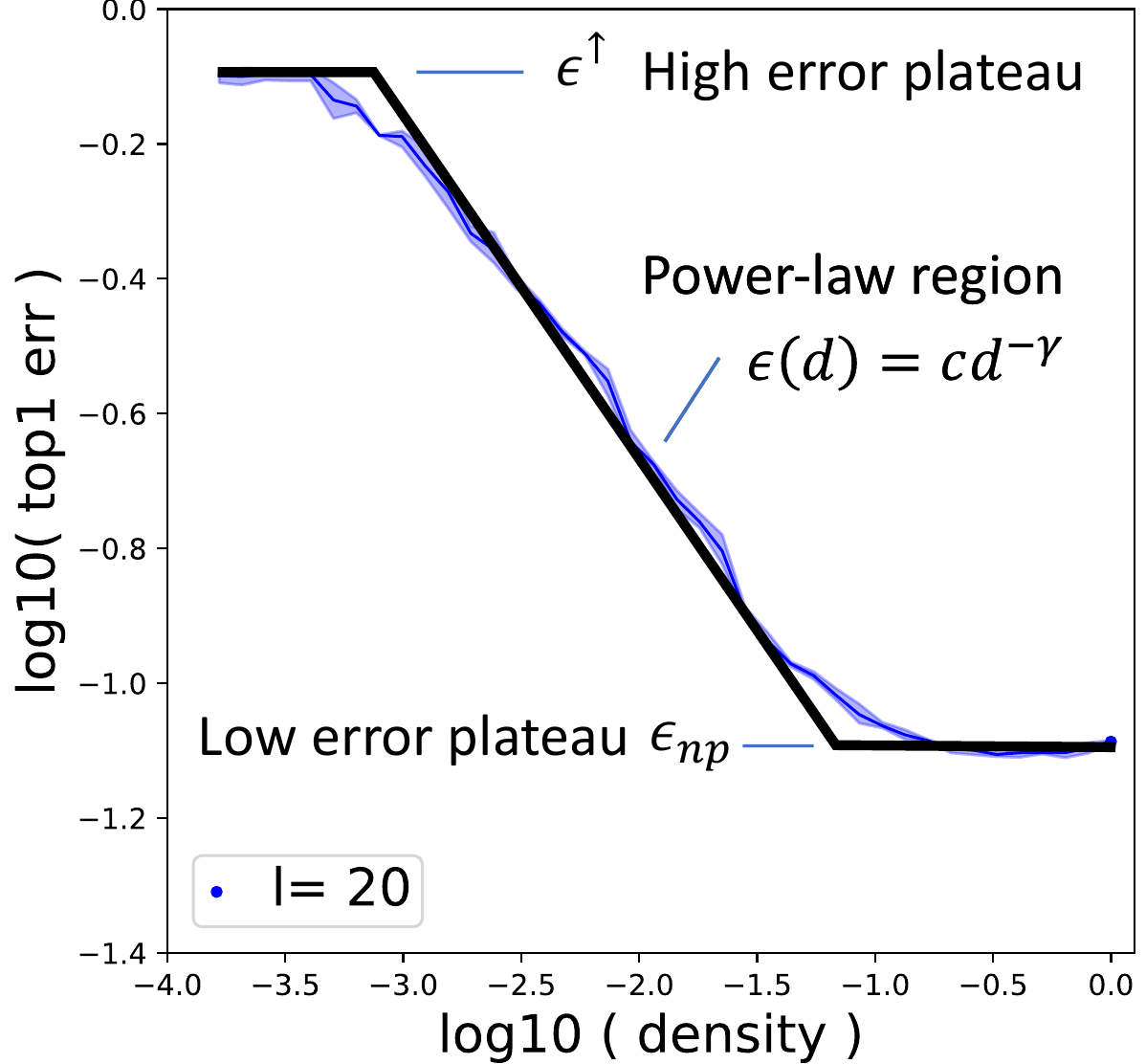}%
\end{minipage}\hfil
\begin{minipage}{0.275\textwidth}
\includegraphics[width=\linewidth,trim={10.8cm 1.9cm 11cm 5.6cm},clip]{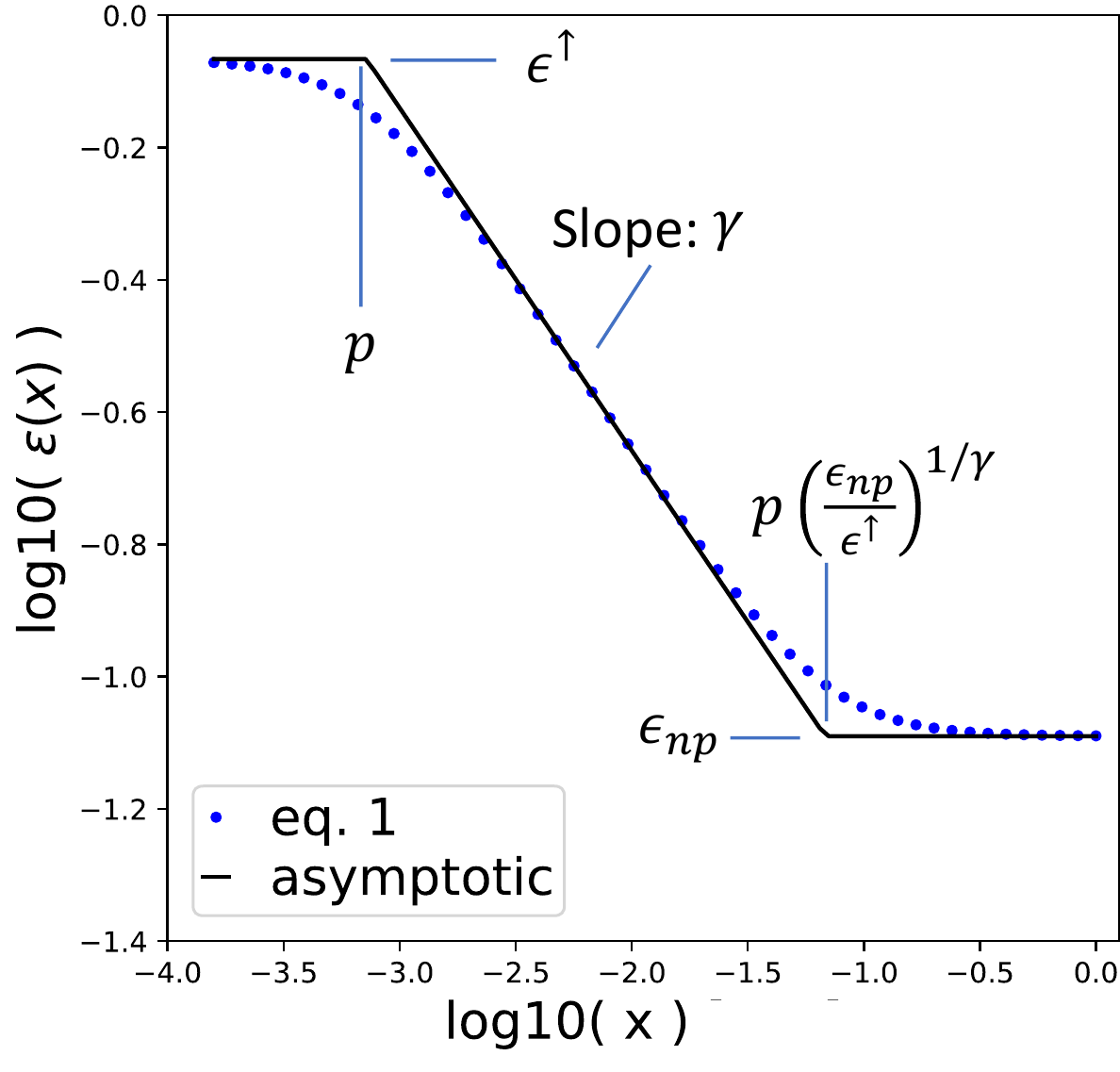}%
\end{minipage}
\vspace{-2.5mm}
\caption{Relationship between density and error when pruning CIFAR-10 ResNets; $w$ varies, $l=20$, $n=N$ (left).
Low-error plateau, power-law region, and high-error plateau when $l=20$, $w=1$, $n=N$ (center).
Visualizing Eq. \ref{eq:prune_density} and the roles of free parameters (right).}
\label{fig:typical_prune_curve}
\end{figure*}

\textbf{Intuition.}
Since IMP prunes 20\% at a time, it produces pruned networks at intermediate densities $\smash{d_k = 0.8^k}$ in the process of pruning to density $\smash{d_K = 0.8^K}$.
In Figure \ref{fig:typical_prune_curve} (left), we plot the error of these pruned networks for CIFAR-10 ResNets with depth $l=20$ and different widths $w$.
All of these curves follow a similar pattern:%
\footnote{The same patterns occur for $l$ and $n$ for CIFAR-10 and $w$ and $n$ for ImageNet (see Appendix \ref{app:sec3-key-observations-alldimensions}). We focus on width for CIFAR-10 here for illustration.}

%\vspace{-5pt}
%\begin{adjustwidth}{2.5mm}{0mm}
\textit{Observation 1: Low-error plateau.} The densest networks (right part of curves) have similar error to the unpruned network: $\smash{\epsilon_{np}(w)}$. We call this the \emph{low-error plateau}.
%\end{adjustwidth}

%\vspace{-5pt}
%\begin{adjustwidth}{2.5mm}{0mm}
\textit{Observation 2: Power-law region.}
When pruned further, error increases linearly on the logarithmic axes of the figure.
Linear behavior on a logarithmic scale is the functional form of a \emph{power law}, where error relates to density via exponent $\gamma$ and coefficient $c$: $\smash{\epsilon(d, w) \approx cd^{-\gamma}}$.
In particular, $\gamma$ is the slope of the line on the logarithmic axes.
%\end{adjustwidth}

%\vspace{-5pt}
%\begin{adjustwidth}{2.5mm}{0mm}
\textit{Observation 3: High-error plateau.} When pruned further, error again flattens; we call this the \emph{high-error plateau} and call the error of the plateau $\smash{\epsilon^\uparrow}$.
%\end{adjustwidth}

%\vspace{-5pt}
Figure \ref{fig:typical_prune_curve} (center) labels these regions for CIFAR-10 ResNet-20 ($w=1$, $n=1$) and shows an approximation of these regions that is piece-wise linear on logarithmic axes.
These observations are our starting point for developing a functional form that estimates error when pruning.

\begin{figure*}
\vspace{-2mm}
\centering
\begin{minipage}{0.28\textwidth}
\includegraphics[width=\linewidth,trim={0 0 0 0.65cm},clip]{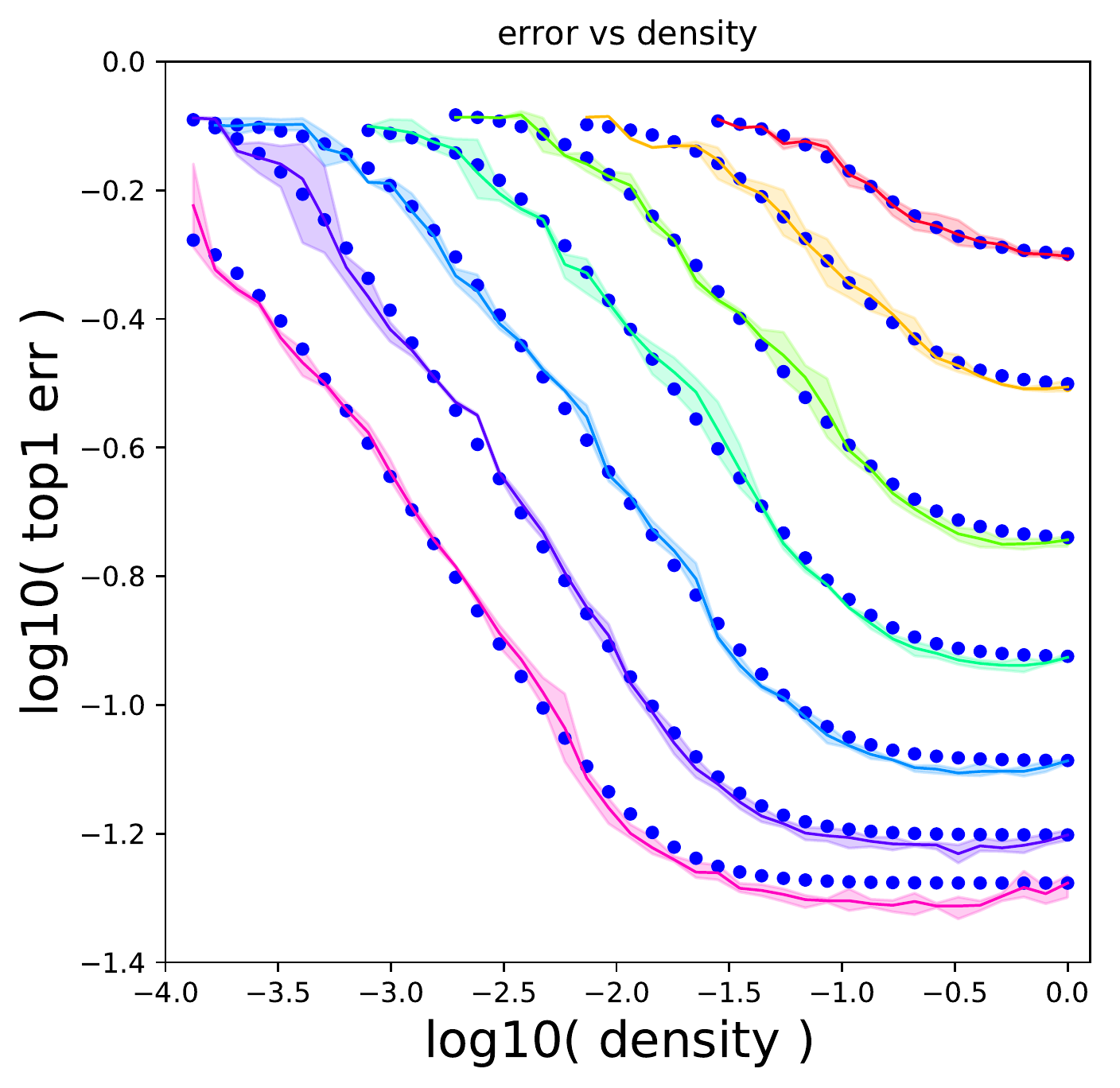}
\end{minipage}\hfil
\begin{minipage}{0.28\textwidth}
\includegraphics[width=\linewidth,trim={0 0 0 0.3cm},clip]{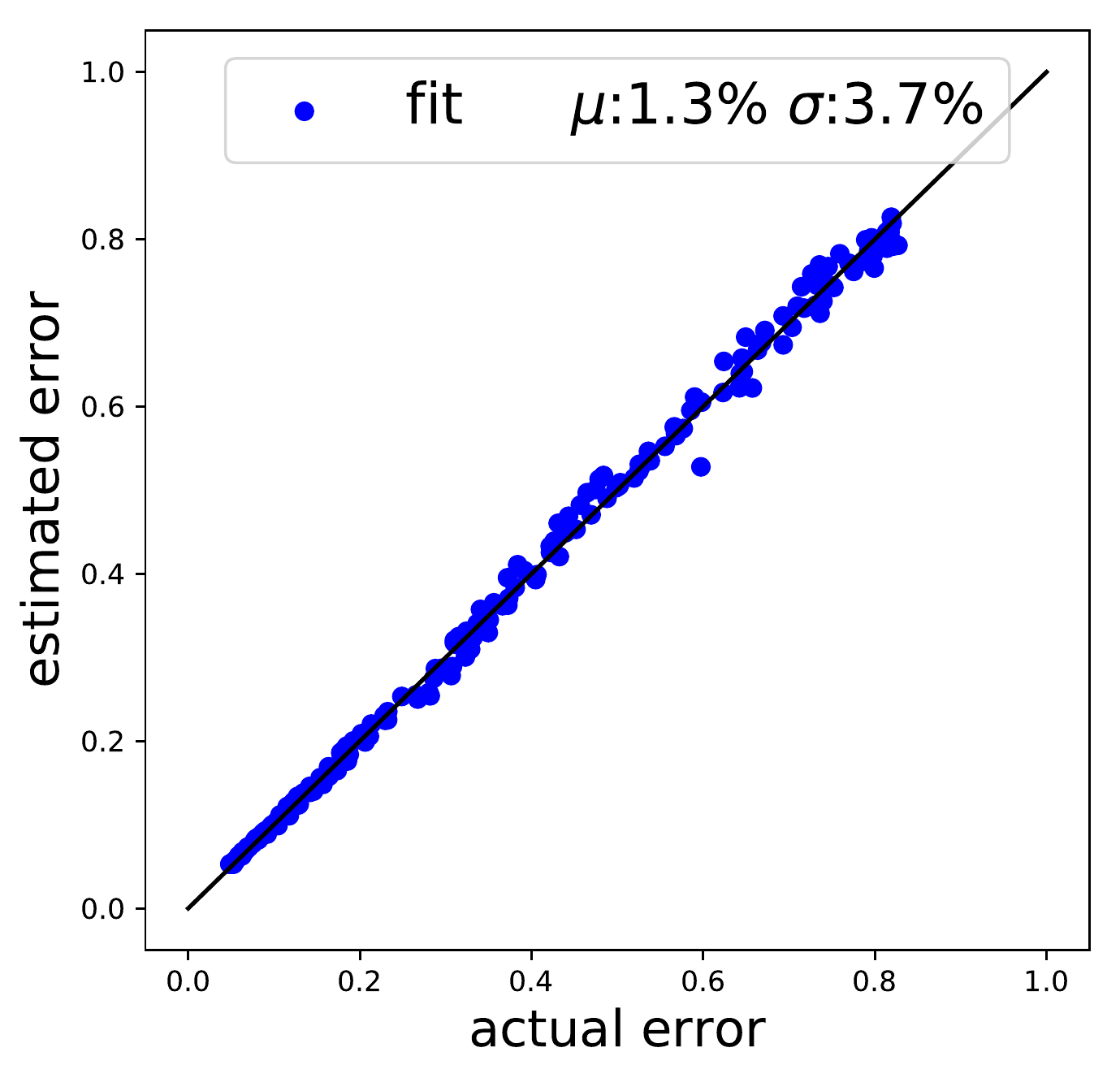}
\end{minipage}\hfil
\begin{minipage}{0.28\textwidth}
\includegraphics[width=\linewidth,trim={0 0 0 0.3cm},clip]{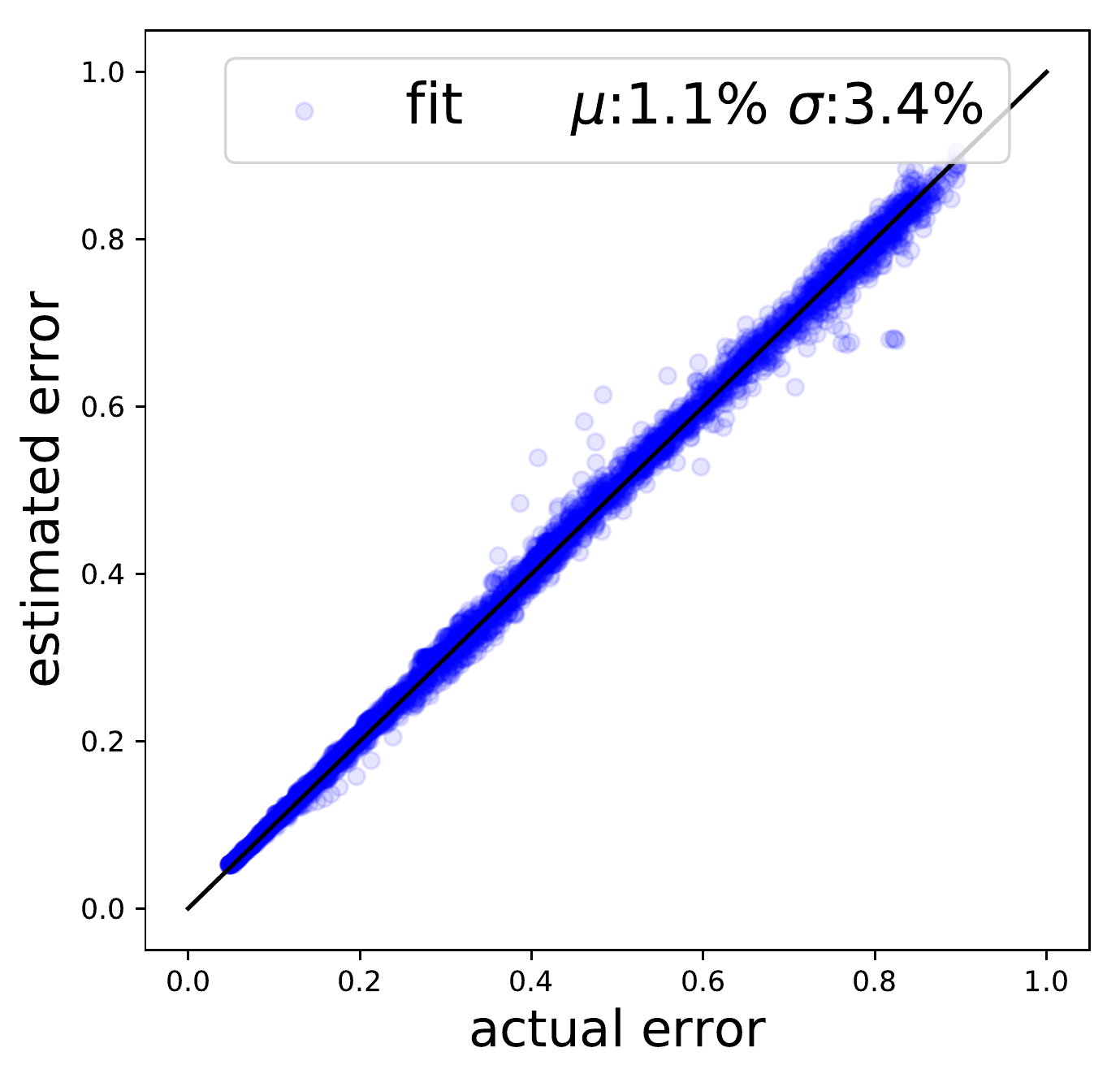}
\end{minipage}
\vspace{-2.5mm}
\caption{Estimated (blue dots) and actual error (solid lines) when pruning CIFAR-10 ResNets; $w$ varies, $l=20$, $n=N$ (left).
Estimated versus actual error for the same networks (center).
Estimated versus actual error for all CIFAR-10 ResNet configurations (right). }
\label{fig:fit_density}
\vspace{-3mm}
\end{figure*}

\textbf{Functional form.}
Our next task is to find a functional form that captures these observations about the relationship between density and error.
In prior work, \citet{rosenfeld2020a} observe that the relationship between width and error shares the same general shape:
it has a region of lower error, a power-law region, and region of higher error.
However, this relationship is different enough from the one we observe (see Appendix \ref{app:difference_in_powerlaws}) to merit an entirely new functional form.

To develop this functional form, we note that the three regions of the curves in Figure \ref{fig:typical_prune_curve} (the low-error plateau, the power-law region, and the high-error plateau) can be described by three power laws: two plateaus with exponent zero and an intermediate region with exponent $\gamma$.
A functional family that arises frequently in systems that exhibit different power-law regions is the \emph{rational family}.
The particular family member we consider is as follows:%
\footnote{The expression $\left\Vert \frac{d-ja}{d-jb} \right\Vert^\gamma = \left(\frac{d^2+a^2}{d^2+b^2}\right)^\frac{\gamma}{2}$ meaning Eq. \ref{eq:prune_density} can be rewritten as {\scriptsize $\epsilon_{np} \left[(d^2+p^2(\epsilon^\uparrow/\epsilon_{np})^{2/\gamma})/(d^2 + p^2) \right]^{\gamma/2}$}}
\vspace{-5pt}
\begin{equation}
\resizebox{0.9\linewidth}{!}{
\label{eq:prune_density}
    $\hat{\epsilon}(\epsilon_{np}, d~|~l, w, n) = \epsilon_{np} \left\Vert \frac{d-jp\left(\frac{\epsilon^\uparrow}{\epsilon_{np}}\right)^{\frac{1}{\gamma}}}{d-j p} \right\Vert^\gamma
    \mbox{where } j = \sqrt{-1}$
}
\end{equation}
\vspace{5pt}

This function's shape is controlled by $\smash{\epsilon_{np}}$, $\smash{\epsilon^\uparrow}$, $\smash{\gamma}$, and $p$ (visualized in Figure \ref{fig:typical_prune_curve}, right).
$\smash{\epsilon_{np}}$ and $\smash{\epsilon^\uparrow}$ are the values of the low and high-error plateaus.
$\smash{\gamma}$ is the slope of the power-law region on logarithmic axes.
$p$ controls the density where the high-error plateau transitions to the power-law region. 

\textbf{Fitting.}
To fit $\smash{\hat \epsilon(\epsilon_{np}, d~|~l, w, n)}$ to actual data $\smash{\epsilon(d, l, w, n)}$, we estimate values for the free parameters $\smash{\epsilon^\uparrow}$, $\smash{\gamma}$, and $p$ by minimizing the relative error $\delta \triangleq \frac{\hat{\epsilon}(\epsilon_{np}, d|l, w, n) - \epsilon(d, l, w, n)}{\epsilon(d, l, w, n)}$ using least squares regression. 
The fit is performed separately for each configuration $(l, w, n)$ for all densities, resulting in per-configuration estimates of $\smash{\hat\epsilon^\uparrow}$, $\smash{\hat\gamma}$, and $\smash{\hat p}$.
 
\textbf{Evaluating fit.}
For a qualitative view,\footnote{Error is a 4-dimensional function, so we can only qualitatively examine 2D projections. All such projections are in Appendix \ref{app:more_fits}.} we plot the actual error\footnote{We compute the error as the mean across three replicates with different random seeds and dataset subsamples.} $\epsilon(d, l, w, n)$ and the estimated error $\hat \epsilon(\epsilon_{np}, d~|~l, w, n)$ as a function density for CIFAR-10 ResNets of varying widths (Figure \ref{fig:fit_density}, left). 
% (In Appendix \ref{app:more_arch_alg}, we plot the same data for other networks and datasets and for SynFlow.)
Our estimated error appears to closely follow the actual error.
The most noticeable deviations occur at large densities, where the error decreases slightly on the low-error plateau whereas we treat it as flat (see Section \ref{sec:design}).

Quantitatively, we measure the extent to which estimated error departs from the actual error using the mean $\mu$ and standard deviation $\sigma$ of the relative deviation $\delta$.
Figure \ref{fig:fit_density} (center) compares the estimated and actual errors for the networks in Figure \ref{fig:fit_density} (left);
Figure \ref{fig:fit_density} (right) shows the same comparison for  all configurations of $l$, $w$, and $n$ on CIFAR-10 and the more than 4,000 pruned ResNets that result.
The relative deviation on all configurations has mean $\mu<2\%$ and standard deviation $\sigma<4\%$; this means that, if the actual error is $10\%$, the estimated error is $9.8 \pm 0.4\%$ ($\hat{\epsilon} = (1-\delta) \epsilon $).

\section{Jointly Modeling Error For All Dimensions}
\label{sec:joint}

In Section \ref{sec:single-network}, we found a functional form $\hat\epsilon(\epsilon_{np}, d~|~l, w, n)$ (Eq. \ref{eq:prune_density}) that accurately predicts the error when pruning a \emph{specific} member of a network family (with depth $l$ and width $w$) trained with a dataset of size $n$.
The parameters governing Equation \ref{eq:prune_density} ($\smash{\epsilon^\uparrow}$, $p$, and $\smash{\gamma}$ ) varied between and depended on the specific configuration of $l$, $w$, $n$.
However, we are interested in a single \emph{joint} scaling law $\hat\epsilon(\epsilon_{np}, d, l, w, n)$ that, given the unpruned network error $\epsilon_{np}(l, w, n)$, accurately predicts error across \emph{all} dimensions: all members of a network family that vary in depth and width, all densities, and all dataset sizes.
Importantly, the parameters of this scaling law must be constants as a function of all dimensions. In this section, we develop this joint scaling law.

\begin{figure}
\vspace{-1mm}
\centering
\begin{minipage}{.24\textwidth}
\includegraphics[width=\linewidth,trim={0 0 0 0.7cm},clip]{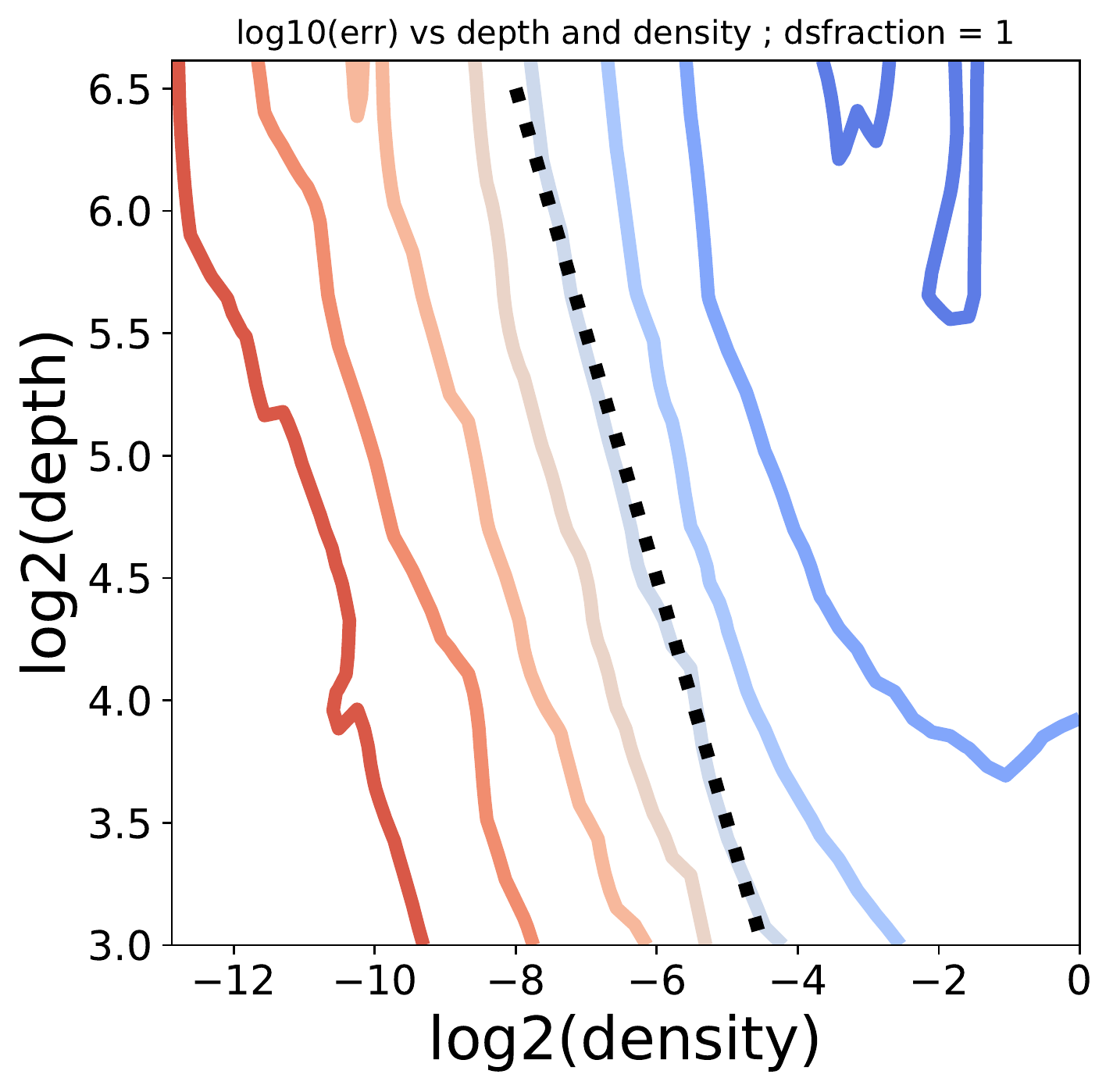}
\end{minipage}%
\begin{minipage}{0.24\textwidth}
\includegraphics[width=\linewidth,trim={0 0 0 0.7cm},clip]{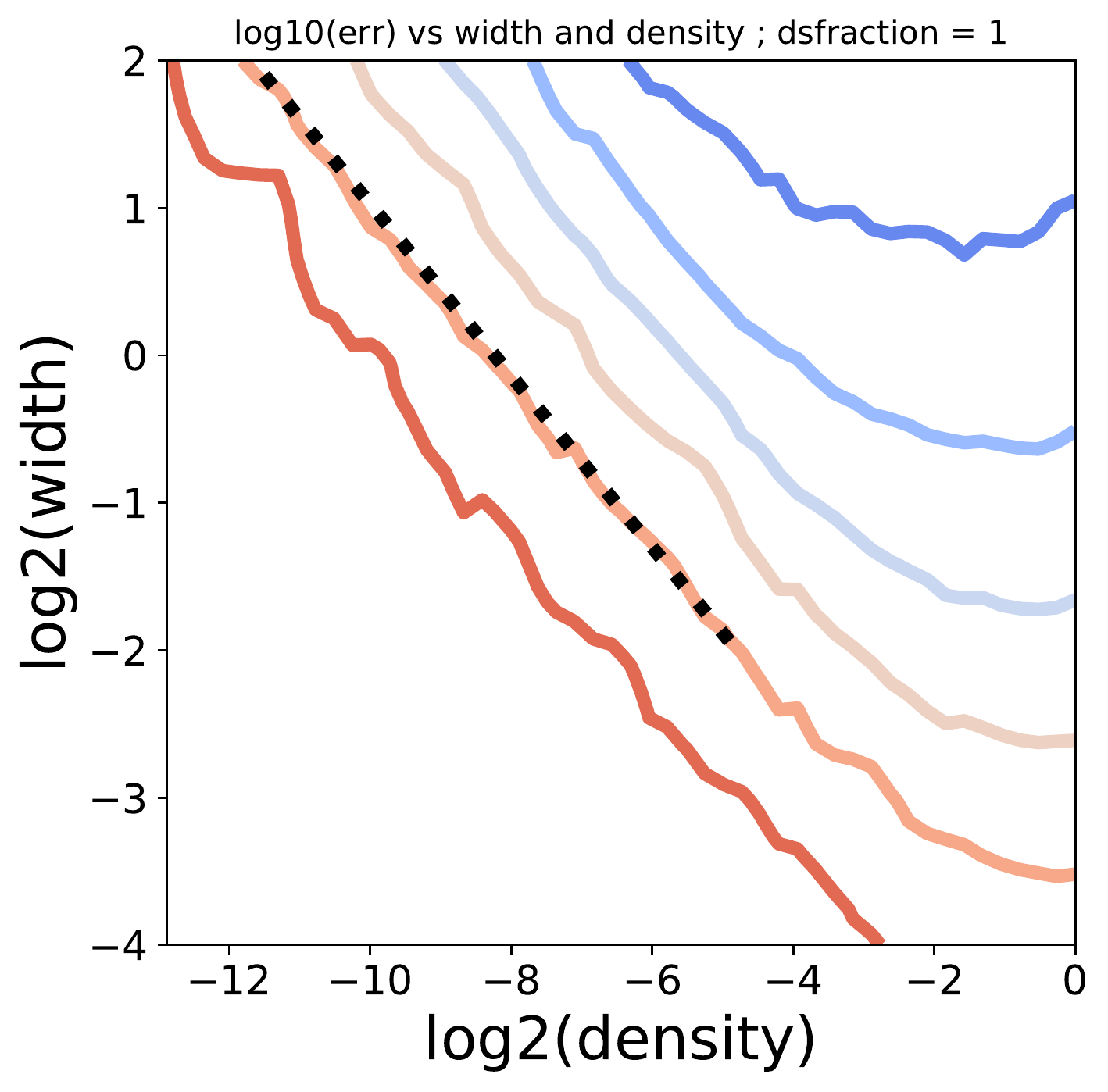}
\end{minipage}
\vspace{-1mm}
\caption{Projections of $\epsilon(d, l, w, n)$ onto two-dimensional planes for the CIFAR-10 ResNets, showing contours of constant error. For low enough densities, the contours have linear slopes on the logarithmic axes---depicted by a reference black-dotted line. The density/depth plane (left).  The density/width plane (right).}
\vspace{-4mm}
\label{fig:basic_form}
\end{figure}

% \vspace{-0.3cm}
\textbf{Intuition: the error-preserving invariant.}
Our desired scaling law $\hat\epsilon(\epsilon_{np}, d, l, w, n)$ will be a four-dimensional function of $d$, $w$, $l$, and $n$. 
To develop an intuition for the relationship between these inputs, we study the interdependence between density and depth or width by examining two-dimensional projections of the actual error $\epsilon(d, l, w, n)$ in Figure \ref{fig:basic_form}.
These plots display contours of constant error as density and depth or width vary. 

Consider the projection onto the plane of density and depth (Figure \ref{fig:basic_form}, left).
The constant-error contours are linear except for in the densest networks, meaning each contour traces a power-law relationship between $d$ and $l$.
In other words, we can describe all combinations of densities and widths that produce error $\epsilon_v$ using $l^\phi d = v$, where $v$ is a constant at which network error is $\epsilon_v$ and $\phi$ is the slope of the contour on the logarithmic axes.
The contours of density and width also have this pattern (Figure \ref{fig:basic_form}, right), meaning we can describe a similar relationship $w^\psi d = v'$.
Finally, we can combine these observations about depth and width into the expression $l^\phi w^\psi d = v''$.

We refer to the expression $l^\phi w^\psi d$ as the \emph{error-preserving invariant}, and we denote it $m^*$.
This invariant captures the observation that there exist many interchangeable combinations of depth, width, and density that achieve the same error and tells us which combinations do so. For example, networks of vastly different densities reach the same error if we vary $l$ and $w$ according to the invariant.

\textbf{Functional form.}
The invariant allows us to convert the functional form $\hat\epsilon(\epsilon_{np}, d~|~l, w, n)$ for a specific $l$, $w$, and $n$ from Section \ref{sec:single-network} into a joint functional form $\hat\epsilon(\epsilon_{np}, d, l, w, n)$ for all $l$, $w$, and $n$.
Rewriting the definition of the invariant, $d = \frac{m^*}{l^\phi w^\psi}$.
We can substitute this for $d$ in the functional form from Section \ref{sec:single-network}.
Finally, by rewriting $p$ as $\frac{p'}{l^\phi w^\psi}$ and canceling, we arrive at the following expression:
\vspace{-14pt}

\begin{scriptsize}
\begin{align}
\label{eq:intermediate_state}
& \hat{\epsilon}(\epsilon_{np}, d~|~l, w, n) = \epsilon_{np} \left\Vert \frac{m^*-jp'\left(\frac{\epsilon^\uparrow}{\epsilon_{np}}\right)^{\frac{1}{\gamma}}}{m^*-j p'} \right\Vert^\gamma
\nonumber = \\ & \epsilon_{np} \left\Vert \frac{l^\phi w^\psi d-jp'\left(\frac{\epsilon^\uparrow}{\epsilon_{np}}\right)^{\frac{1}{\gamma}}}{l^\phi w^\psi d-j p'} \right\Vert^\gamma = \hat{\epsilon}(\epsilon_{np}, d, l, w, n)
\end{align}
\end{scriptsize}
\vspace{-14pt}

which is the joint functional form $\smash{\hat{\epsilon}(\epsilon_{np}, d, l, w, n)}$ of all four dimensions $d$, $l$, $w$, and $n$.
Critically, for this to be a useful joint form, the free parameters $\smash{e^\uparrow, p', \text{ and } \gamma}$ must be constants shared across all possible values of $d$, $l$, $w$, and $n$. We will assume this is the case and directly quantify how well this assumption holds in the evaluation section below.

\begin{figure*}[t]
\vspace{-2mm}
\centering
\begin{minipage}{0.28\textwidth}
\includegraphics[width=\linewidth,trim={0cm 0 0 0.6cm},clip]{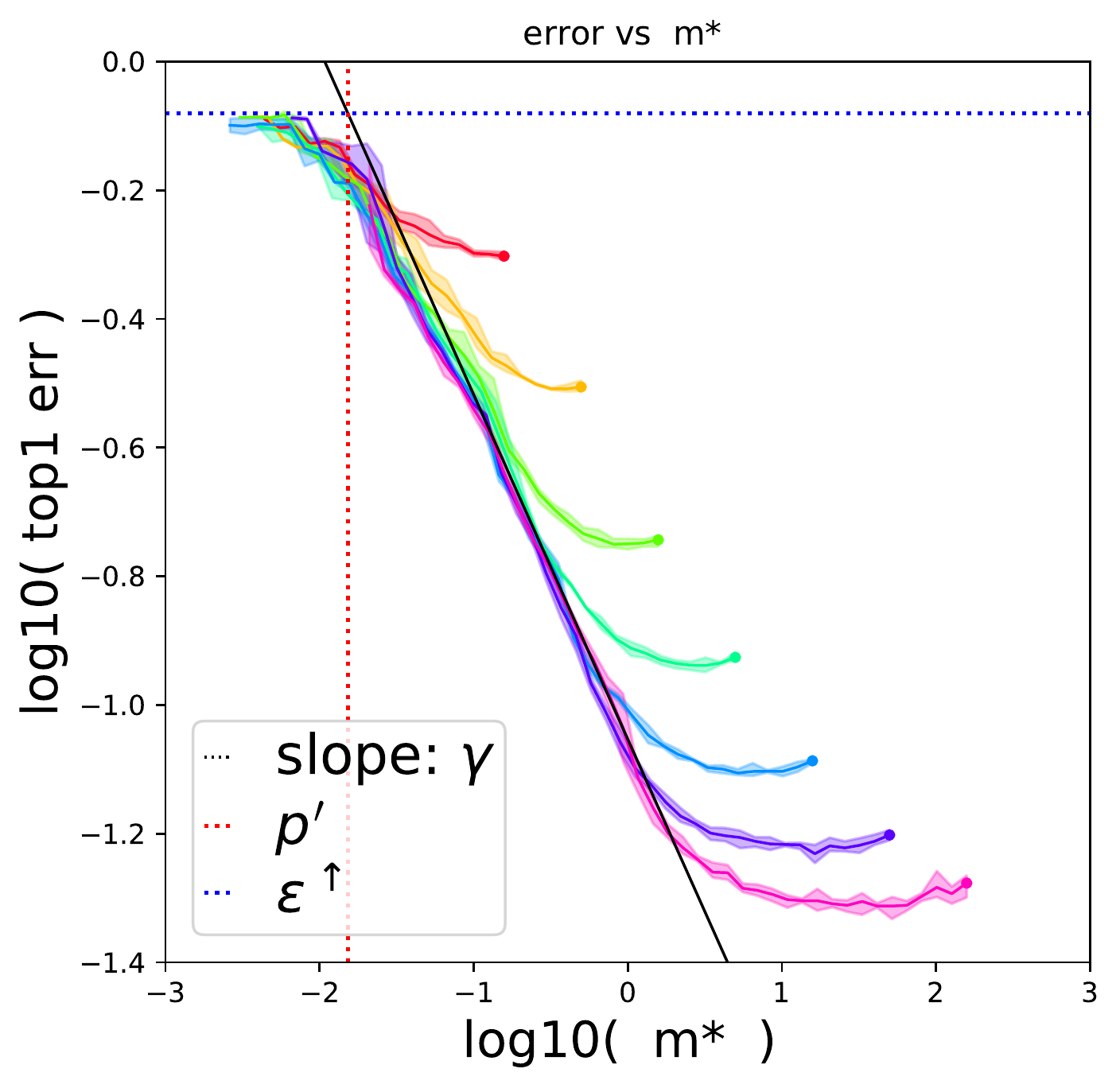}
\end{minipage}\hfil
\begin{minipage}{0.245\textwidth}
  \includegraphics[width=\linewidth,trim={2cm 0 0 0.6cm},clip]{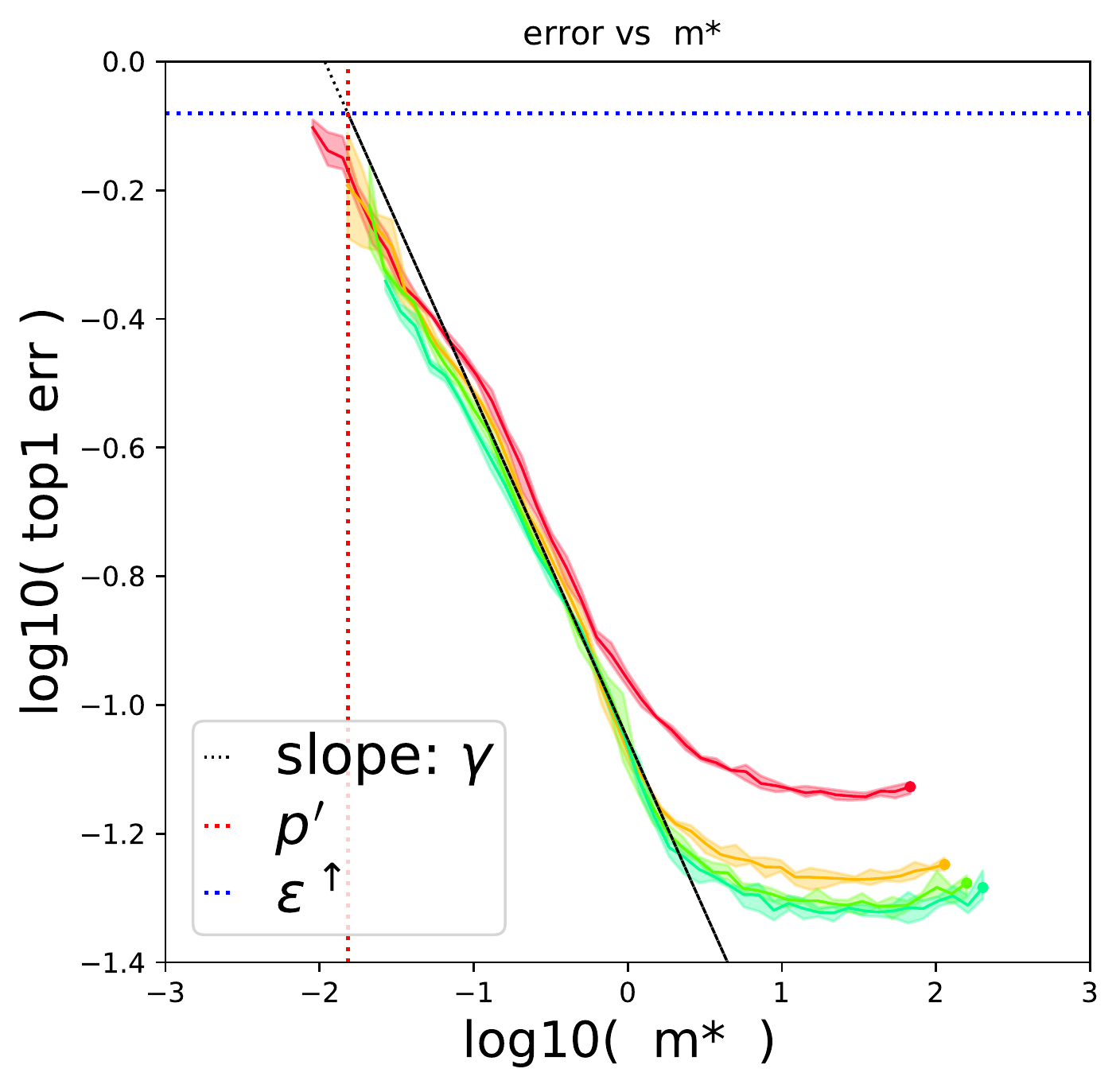}
\end{minipage}\hfil
\begin{minipage}{0.245\textwidth}
  \includegraphics[width=\linewidth,trim={2cm 0 0 0.6cm},clip]{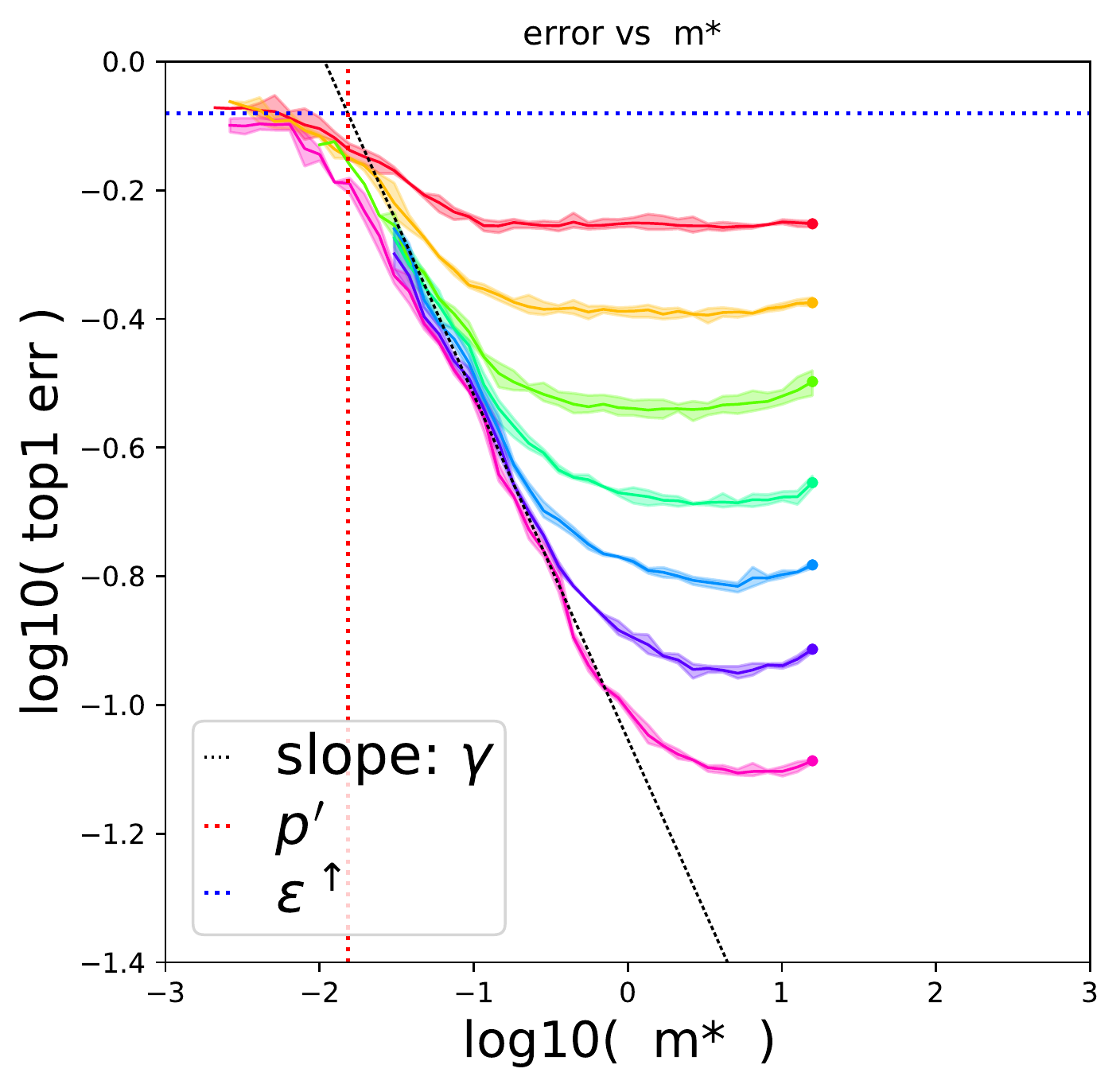}
\end{minipage}
\vspace{-2mm}
\caption{Relationship between $m^*$ and error when pruning CIFAR-10 ResNets and varying $w$ (left, $l=20$, $n=N$),  $l$ (center, $w=1$, $n=N$), $n$ (right, $l=20$, $w=1$). We annotate $\gamma$, $\epsilon^\uparrow$, and $p'$; they qualitatively appear to take on similar values in all cases, an observation that we use to inform the design of our joint scaling law. }
\label{fig:cifar_joint_observations}
\end{figure*}

For qualitative intuition as to why this is a reasonable assumption, consider the relationship between $m^*$ and the test error of pruned networks as we vary depth, width, and dataset size (Figure \ref{fig:cifar_joint_observations}).
Across all projections, the annotated $\smash{e^\uparrow}$ (error of the high-error plateau), $\gamma$ (slope of the power-law region) and $\smash{p'}$ (value of $\smash{m^*}$ where the high-error plateau transitions to the power-law region) appear the same.

The preceding discussion addresses how we handle $l$, $w$, and $d$ in our joint scaling law.
We address dataset size $n$ in Eq. \ref{eq:intermediate_state} implicitly through the way that it affects $\epsilon_{np}$, and we validate that this is a reasonable choice through the evaluation below.
We retain the explicit form $\hat\epsilon(...,n)$ to stress that the lack of explicit dependency on $n$ is non-trivial and was not known prior to our work.

\begin{figure*}[t]
    \centering
        \begin{minipage}{0.28\textwidth}
            \centering
            \includegraphics[width=\linewidth,trim={0.2cm 0 0 0.3cm},clip]{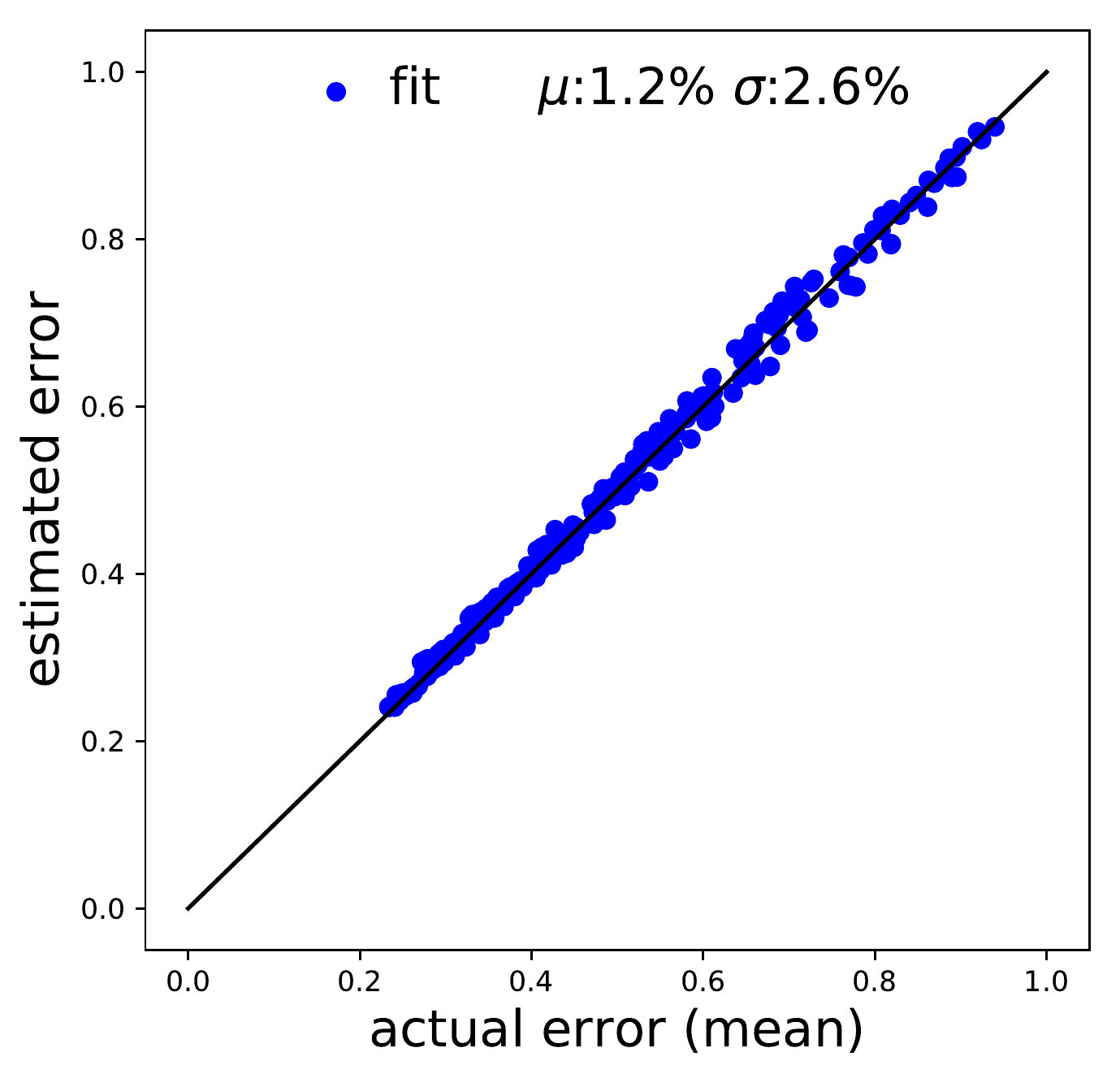}
            \label{fig:fit_imagent}
        \end{minipage}\hfil
        \begin{minipage}{0.28\textwidth}
            \centering
            \includegraphics[width=\linewidth,trim={0.2cm 0 0 0.3cm},clip]{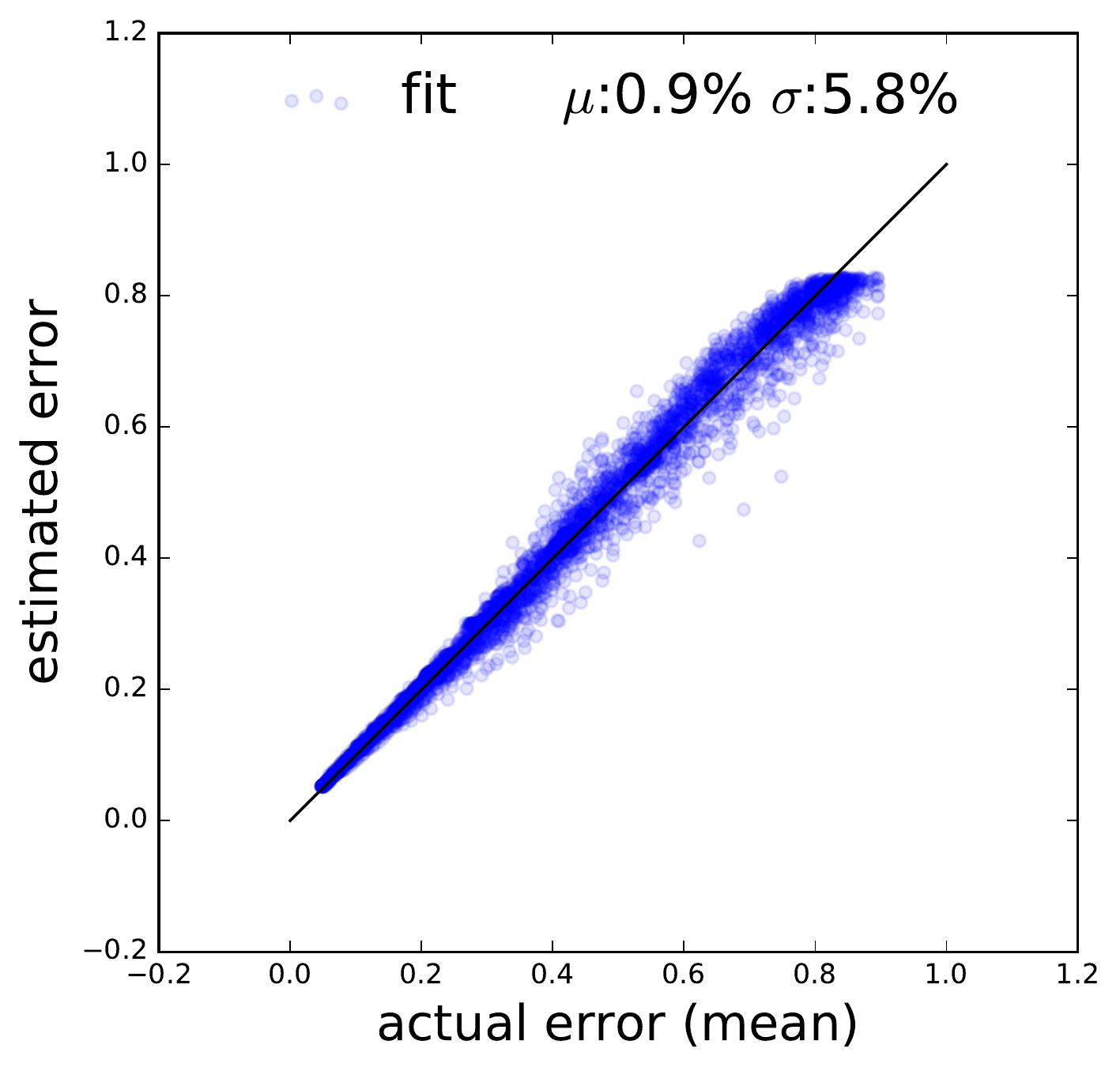}
            \label{fig:fit_corr_all}
        \end{minipage}\hfil
        \begin{minipage}{0.28\textwidth}
            \centering
            \includegraphics[width=\linewidth,trim={0.2cm 0 0 0.3cm},clip]{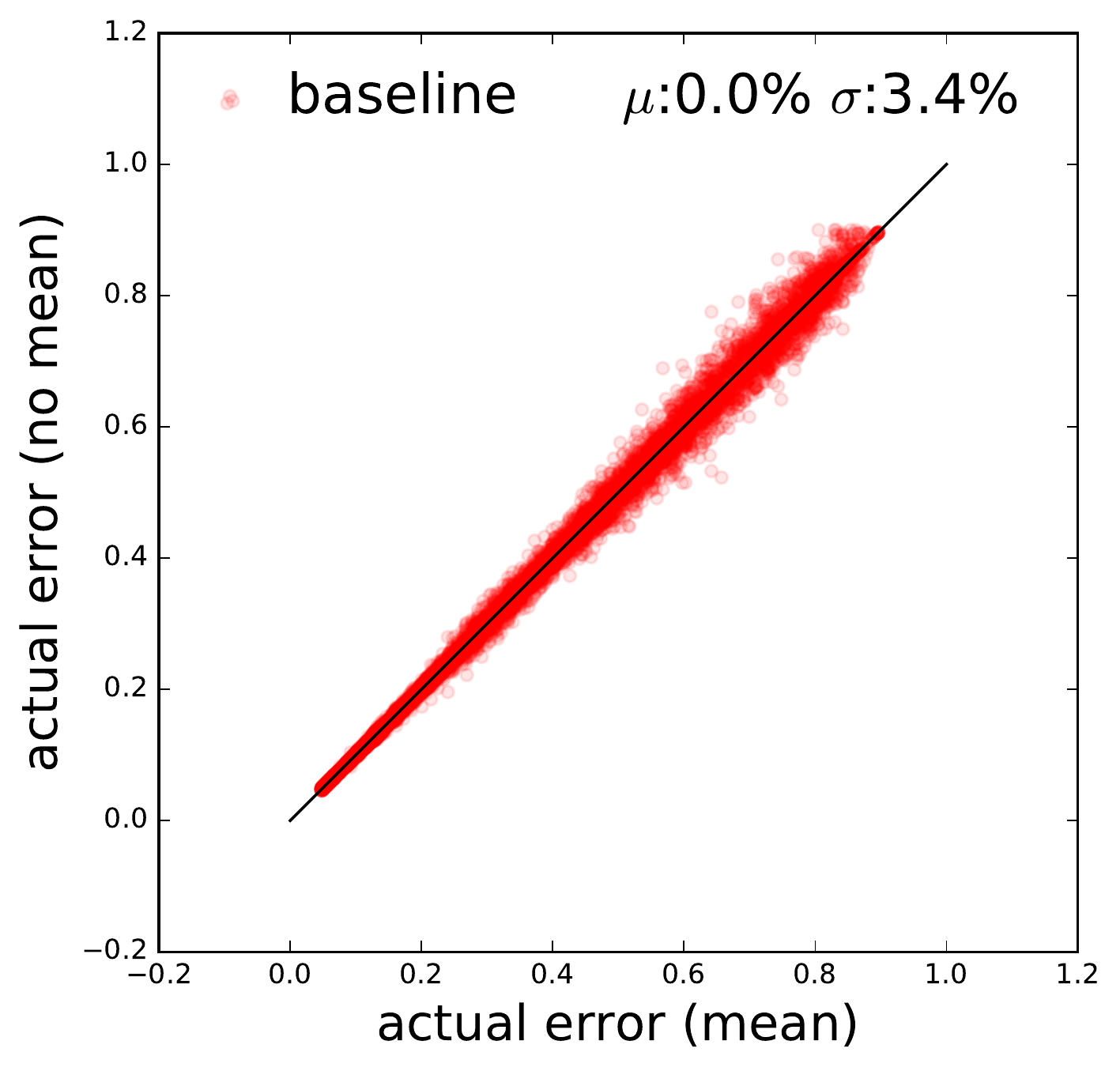}
            \label{fig:cifar_baseline_joint}
        \end{minipage}
\vspace{-6mm}
\caption{
Estimated versus mean actual error for all configurations $(d,w,n)$ for ImageNet (left) and $(d,w,l,n)$ for CIFAR-10 (center). 
The variation in error when running the same experiment on CIFAR-10 three times with different random seeds (right).}
\label{fig:joint_fit1}
\vspace{-2mm}
\end{figure*}

\textbf{Fitting.}
To fit $\hat \epsilon(\epsilon_{np}, d, l, w, n)$ to the actual data $\epsilon(d, l, w, n)$, we estimate values for the free parameters $\smash{\epsilon^\uparrow, \gamma, p', \phi \text{ and } \psi}$ by minimizing the relative error $\delta \triangleq \frac{\hat{\epsilon}(\epsilon_{np}, d, l, w, n) - \epsilon(d, l, w, n)}{\epsilon(d, l, w, n)}$ using least squares regression. The fit is performed jointly over all configurations of $d, l, w,$ and $n$, resulting in joint estimates of  $\smash{\hat\epsilon^\uparrow, \hat\gamma, \hat p, \hat\phi, \text{ and } \hat\psi}$.
One can also perform a partial fit for a subset of dimensions (e.g., just $d$, $l$, and $n$) by omitting $\smash{\phi}$ and/or $\smash{\psi}$ (see Appendix \ref{app:more_fits}).

\textbf{Evaluating fit.}
In Figure \ref{fig:joint_fit1}, we plot the actual error $\epsilon(d, l, w, n)$ and the estimated error $\smash{\hat \epsilon(\epsilon_{np}, d, l, w, n)}$ for
the CIFAR-10 ResNets (all widths, depths and dataset sizes)
and ImageNet ResNets (all widths and dataset sizes for depth 50).
As in Section \ref{sec:single-network}, our estimated error appears to closely follow the actual error. 
Deviations arise mainly at high densities where error decreases below $\epsilon_{np}$ and low densities approaching high error saturation.

We again quantify the fit of the estimated error using the mean $\mu$ and standard deviation $\sigma$ of the relative deviation $\delta$.
The relative deviation on the joint scaling laws for the CIFAR-10 and ImageNet networks has a mean $\mu<2\%$ and standard deviation of $\sigma<6\%$.

To contextualize these results, Figure \ref{fig:joint_fit1} (right) quantifies the variation in error we see over multiple replicates of the CIFAR-10 experiments due to using different random seeds.
It plots the minimum, maximum, and mean errors across the three replicates we ran.%
\footnote{We only ran a single replicate of the ImageNet experiments due to the significant cost of collecting data.}
The variation across trials has a standard deviation of $\sigma=3.4\%$, sizeable relative to the estimation error of $\sigma=5.8\%$ for the joint scaling law. This indicates that a significant portion of our error may stem from measurement noise.

The functional form has just five parameters and obtains an accurate fit on 4,301 points on CIFAR-10 and 274 points on ImageNet, suggesting it is a good approximation.
In Appendix \ref{app:more_arch_alg}, we show that it achieves a similarly good fit for additional architectures and datasets.
In Section \ref{app:interpolation}, we show that, although we use a large number of points to develop and evaluate our functional form here, it is possible to get a good fit with far fewer points and the fit has low sensitivity to the choice of points.

\section{Sensitivity of Fit to Number of Points}
\label{app:interpolation}

In Section \ref{sec:joint}, we showed that our scaling law was accurate when we fit it on all of the available data.
Now that we possess the functional form and know that it can accurately model the behavior of IMP, we study the amount of data necessary to obtain a stable,\footnote{Stability is defined as a small change in output relative to a change in input. The requirement here is that a change in choice of points leads to a small expected change in estimation accuracy.} accurate fit.
This question is especially relevant when the functional form is applied to new settings---new networks, datasets, or pruning algorithms---and we must collect new data to do so.
The functional form has only five parameters, suggesting that few experiments will be necessary to obtain an accurate fit.

\textbf{Experiments.}
To evaluate the effect of the number of points on the stability and accuracy of the fit, we randomly sample varying numbers of points, fit the scaling law to those points, and evaluate the quality of the fit over all points.
We sample these points in two ways.

\textit{Experiment 1.} Randomly sample $T$ networks $(w, l, n, d)$.
This experiment evaluates the stability and accuracy of the fit when naively varying the number of points.

\textit{Experiment 2.} Randomly sample $T$ network configurations $(w, l, n)$ and include all densities $d$ for each configuration. This experiment captures the specific use case of IMP, where obtaining data at density $d$ requires obtaining all densities $d' > d$. As such, we anticipate that data will be obtained by iteratively pruning a small number of configurations $(w, l, n)$ to low density.

\begin{figure}
\centering
\begin{minipage}{0.24\textwidth}
    \includegraphics[width=\linewidth]{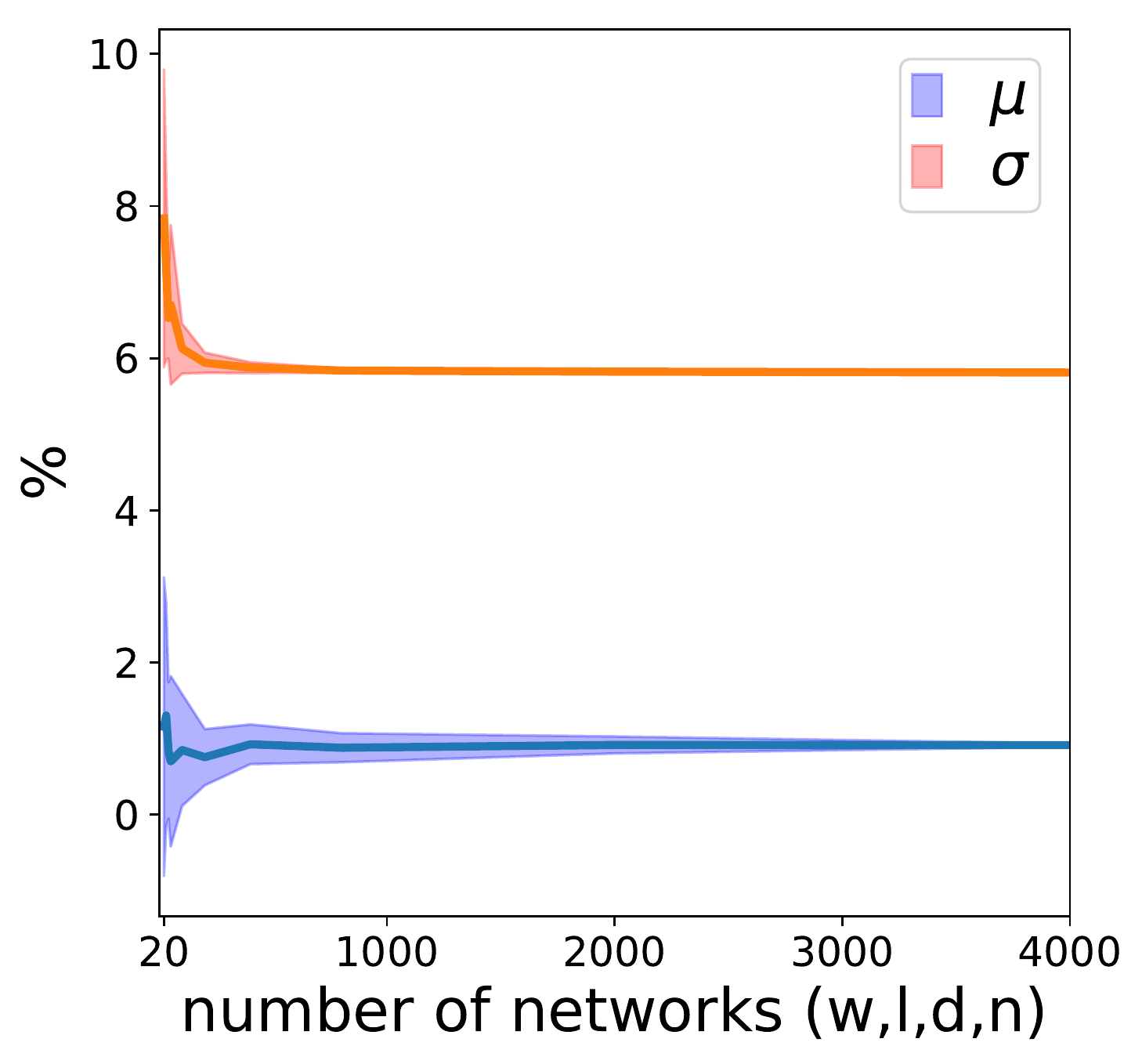}
\end{minipage}%
\begin{minipage}{0.24\textwidth}
    \includegraphics[width=\linewidth]{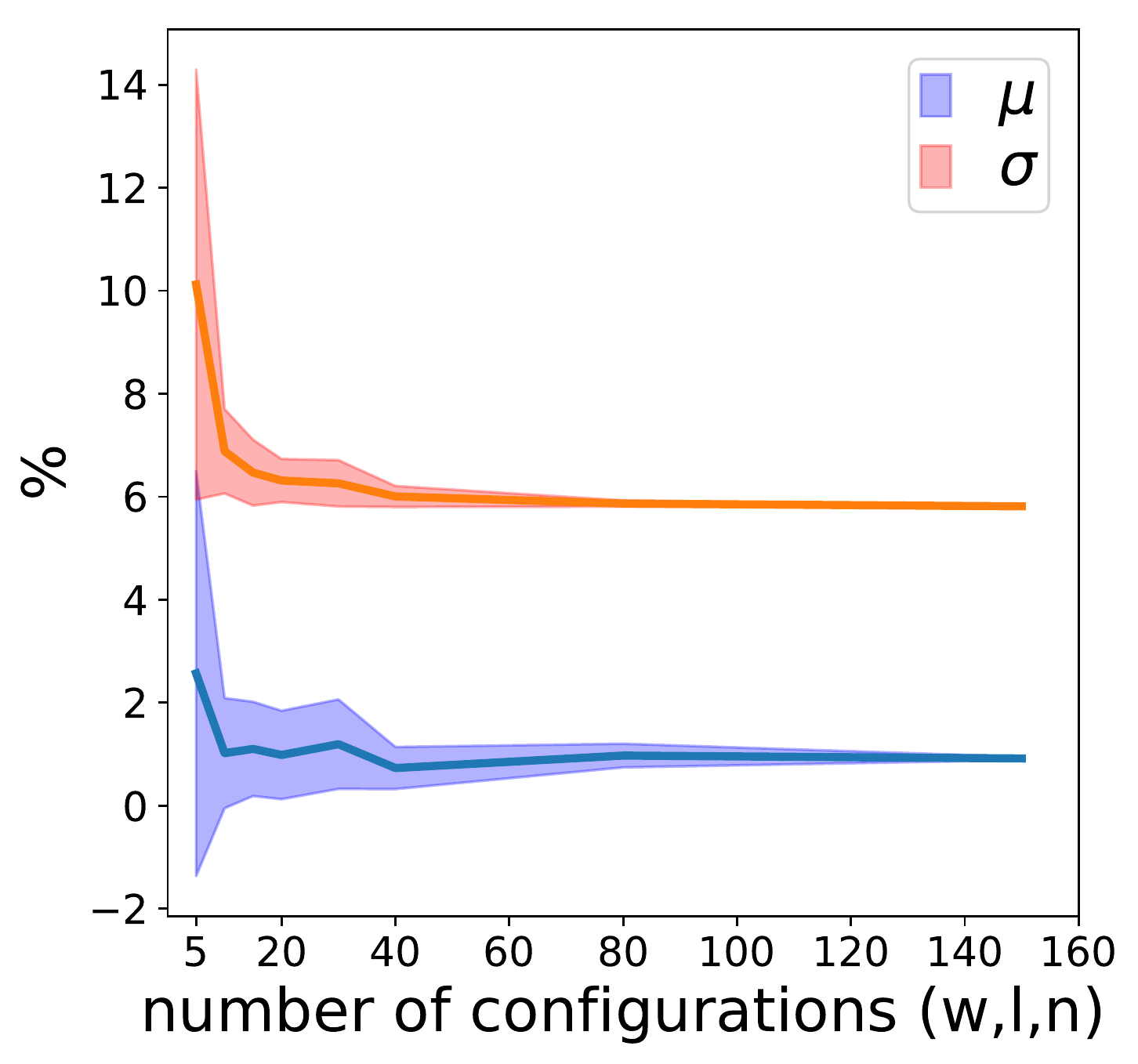}
\end{minipage}

\caption{The effect of the number of points used to fit our scaling law (on the CIFAR-10 ResNets) on $\mu$ and $\sigma$. Left: experiment 1 from Section \ref{app:interpolation} (random points $w$, $l$, $d$, $n$). Right: experiment 2 from Section \ref{app:interpolation} (random configurations $w$, $l$, $n$ and all densities).}
%\vspace{-3mm}
\label{fig:stability1}
\end{figure}

\textbf{Results.}
We perform each experiment for many different values of $T$ on the CIFAR-10 ResNets pruned with IMP.
We repeat the experiment at each value of $T$ 30 times with different samples of points and report the mean and standard deviation of $\mu$ and $\sigma$ for the fit.
Experiments 1 and 2 respectively appear in Figure \ref{fig:stability1} left and right.
The shaded areas represent one standard deviation from the mean in each direction.
On Experiment 1, when just 40 networks $(w, l, d, n)$ are available, the standard deviation on both $\mu$ and $\sigma$ is just one percentage point.
On Experiment 2, when just 15 random configurations of $(w, l, n)$ are available at all densities, we similarly achieve standard deviation below 1\%.
In both cases, as the number of networks increases, the standard deviation decreases further.

These results show that, now that our scaling law is known, it is possible to obtain an accurate (and stable) estimation using far less data than we used to evaluate the quality of the fit in Section \ref{sec:joint}.
This implies that translating our scaling law to new settings will be far less data-intensive than developing and evaluating it in the first place.
Moreover, the results in this section reflect a particularly naive way of selecting points: doing so randomly; we made no effort to ensure that, for example, the networks represented a diverse range of widths, depths, dataset sizes, and densities. 
By selecting these networks in a strategic way, it may be possible to further reduce the number of networks necessary to obtain a similarly accurate fit.

\section{Discussion: Selecting a Functional Form}
\label{sec:design}

We have shown that our proposed functional form $\hat\epsilon(d, l, w, n)$ accurately approximates the error when pruning families of neural networks.
In this section, we discuss some of the key criteria that led us to select this particular functional form.
We intend this section to provide insight into our choices in the context of the broader design space and to highlight opportunities for further refinement.

\begin{figure*}
\centering
\begin{minipage}{.35\linewidth}
    \centering
    \includegraphics[width=.8\linewidth,trim={0 0 0 0.65cm},clip]{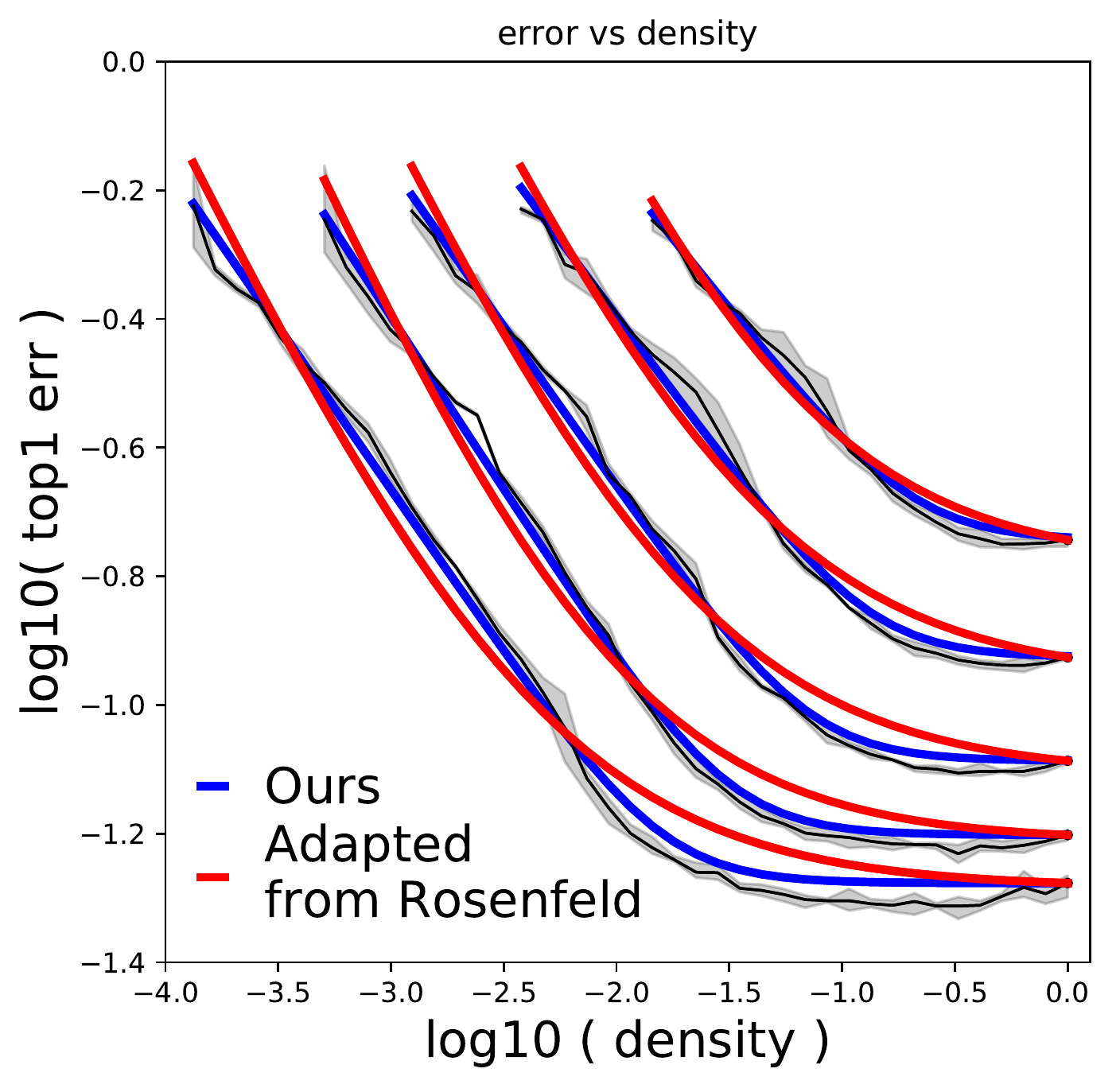}
    \label{fig:comparison}
    \vspace{-2mm}
    \caption{Estimated error from our scaling law (blue) and the scaling law adapted from \citet{rosenfeld2020a} (red) for CIFAR-10 ResNet as $w$ varies. Actual error is in black. Our scaling law better captures transitions between the low-error plateau and the power-law region.}
\end{minipage}%
\hfil
\begin{minipage}{0.6\linewidth}
\vspace{-1mm}
\begin{minipage}{.48\linewidth}
    \includegraphics[width=\linewidth,trim={10.5cm 1.9cm 11cm 5.6cm},clip]{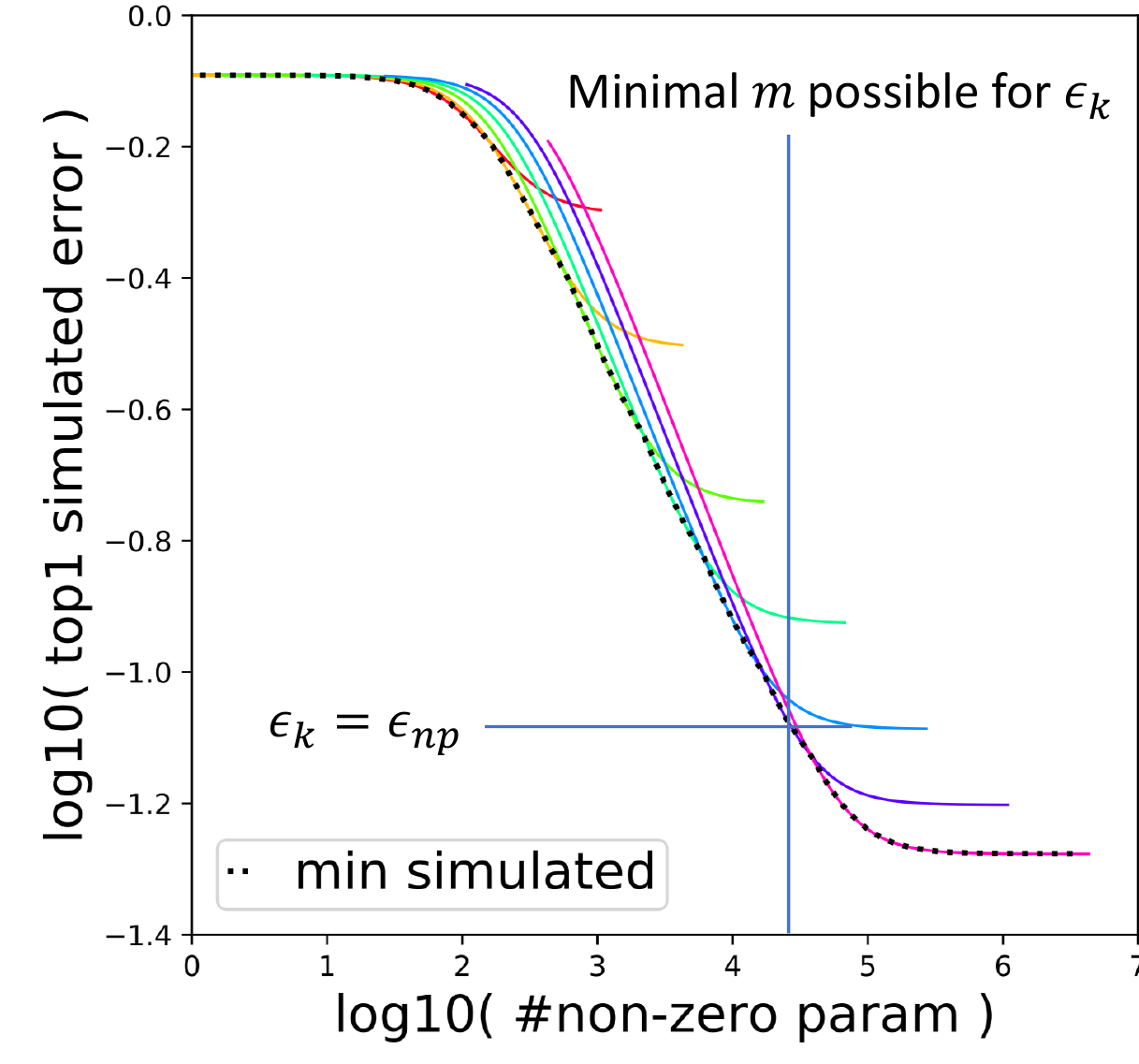}
\end{minipage}
% \hfill
\begin{minipage}{.48\linewidth}
    \includegraphics[width=\linewidth,trim={10.5cm 1.9cm 11cm 5.6cm},clip]{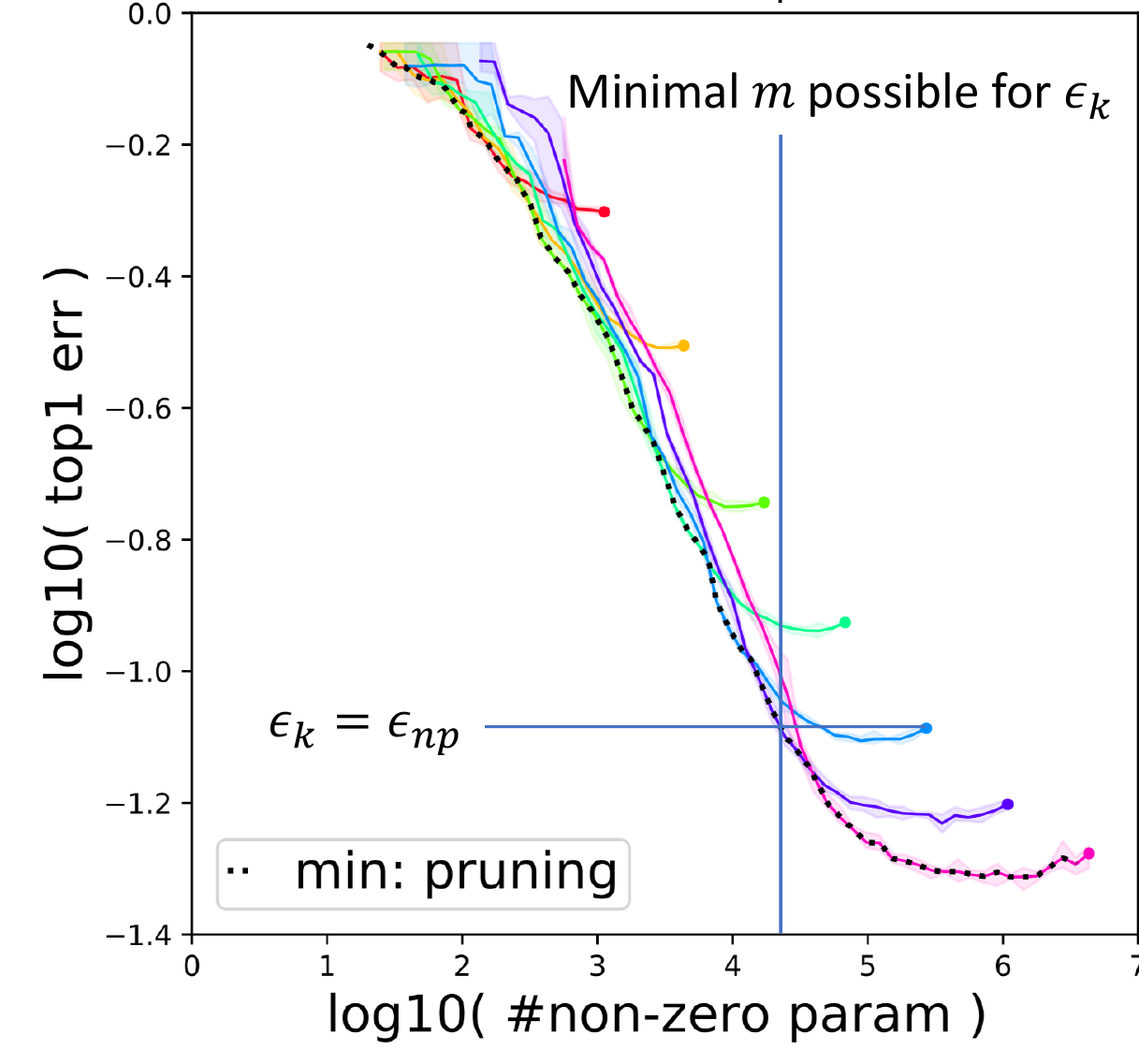}
\end{minipage}
    \vspace{-2mm}
    \caption{
    Estimated error as width varies for the CIFAR-10 ResNets  (left).
    Actual error as width varies for the CIFAR-10 ResNets (right).
    The dotted black line is the minimal number of parameters necessary to reach each error $\epsilon_k$ among all of the pruned networks.
    Reaching this point requires starting with a particular lower-error network (purple) and pruning until error increases to $\epsilon_k$.
    Starting too large (pink) will miss this point.
    }
    \label{fig:misc}
    \end{minipage}
   % \vspace{-4mm}
\end{figure*}

\textbf{Criterion 1: Transitions.}
In Section \ref{sec:single-network}, we observe that, when pruning a neural network with IMP, error has a low-error plateau, a power-law region, and a high-error plateau.
Between these regions are \emph{transitions} where error varies smoothly from one region to the next.
Matching the shape of these transitions was a key consideration for selecting our function family.
To illustrate the importance of properly fitting the transitions, Figure \ref{fig:comparison} shows two possible functional families for fitting the relationship between density and error for the CIFAR-10 ResNets.
Actual error is in black, and the functional form from Section \ref{sec:single-network} is in blue.
In red is the fit for a functional form adapted from the one that \citet{rosenfeld2020a} use to model the relationship between width and error.
The difference between these functional families is the way they model transitions, and the one we choose in this paper better models the transitions in our setting.
For further discussion of this comparison, see Appendix \ref{app:difference_in_powerlaws}.

\textbf{Criterion 2: Few, interpretable parameters.}
Selecting a functional form is not merely a curve-fitting exercise.
We seek the underlying structure that governs the relationships between $d, l, w, n$, and error in a manner akin to a law of physics.
As such, our functional form should have a small number of parameters that are \emph{interpretable}.
In our functional form (Eq. \ref{eq:intermediate_state}), each parameter has a clear meaning. 
The parameters 
$\epsilon^\uparrow$, $p'$, and $\gamma$ control the high-error plateau, the transition to the power-law region, and the slope of the power-law region.
$\phi$ and $\psi$ control the interchangeability of width and depth with density.
We approximate error over multiple orders of magnitude and 4,301 configurations of ResNet-20 on CIFAR-10 with just five parameters, indicating we have distilled key information about the behavior of pruning into our functional form.

\textbf{Sources of systemic error and limitations.}
By seeking to minimize the number of parameters in our functional form, we leave some phenomena unmodeled.
In particular, there are two phenomena we have chosen \emph{not} to model that introduce systemic error.
First, the low-error plateau is not a plateau.
Error often improves slightly at high densities before returning to $\epsilon_{np}$ during the transition to the power-law region.
Our model treats the region as flat and treats error as monotonically increasing as density decreases.
This source of error accounts for a bias of $\sim1\%$ relative error in our estimation (Appendix \ref{app:magic-one-percent}).
Second, we model both transitions (between the power-law region and each plateau) with a single shape and the same transition rate.
If we treated them separately and used higher-order terms in the rational form, we could potentially reduce some of the residual error in our estimation at the cost of additional complexity.

\section{Implications and Conclusions}
\label{sec:conclusions}

Our main contribution is a functional form $\hat \epsilon(\epsilon_{np}, d, l, w, n)$ that accurately predicts the error when pruning members of a network family using IMP.
There are several broader implications of our ability to characterize pruning in this way.
The mere existence of this functional form means there is indeed structure to the way pruning affects error. 
Although prior work \citep{cai2019once} has implicitly relied on the existence of structure for a different pruning method, we are the first to explicitly describe such structure.
This functional form enables a framework in which we can reason conceptually and analytically about pruning.
In doing so, we can make new observations about pruning that are non-obvious or costly to exhaustively demonstrate empirically.
For example, recall our motivating question:

\textit{Given a family of neural networks, which should we prune (and by how much) to obtain the network with the smallest parameter-count such that its error does not exceed some threshold $\epsilon_k$?}

This is an increasingly common question as the community begins to think about pruning as a way of seeking the optimal pruned member of a \emph{family} of networks rather than as a technique applied to a specific network in isolation.
At its heart, the question is really an optimization problem: find the configuration of $d$, $l$, and $w$ that minimizes parameter-count $m$ subject to an error constraint: $\smash{\argminB_{w,l,d} m \text{   s.t.   } \hat{\epsilon}=\epsilon_k}$. For ResNets, the parameter-count $m$ is proportional to $(dlw^2)$.%
\footnote{Increasing the depth linearly increases the number of parameters, but increasing the width quadratically increases the number of convolutional filters and thereby the parameter-count.}
Hence, this yields the following optimization problem:
\vspace{-3mm}

{
\small
\begin{align*}
    & l,w,d = \argminB_{l,w,d} lw^2d \text{~~~~~s.t.~~~~~} \\
    & \epsilon_{np} \left\Vert  \frac{l^\phi w^\psi d-jp'(\epsilon^\uparrow/\epsilon_{np})^{1/\gamma}}{l^\phi w^\psi d-j p'} \right\Vert^\gamma=\epsilon_k
\end{align*}
}

\vspace{-2mm}
This optimization problem is solvable directly without running any further experiments.

Studying this optimization problem reveals a useful insight about---in this case---the CIFAR-10 ResNets.
In the pruning literature, it is typical to report the minimum density where the pruned network matches the error $\epsilon_{np}(l, w)$ of the unpruned network  \citep{han}.
However, our scaling law suggests  this is not the smallest model to achieve error $\epsilon_{np}(l, w)$.
Instead, it is better to train a larger network with depth $l'$ and width $w'$ and prune until error reaches $\epsilon_{np}(l, w)$, despite the fact that error will be higher than $\epsilon_{np}(l', w')$.
This analytic result parallels and extends the findings of \citet{li2020train} on NLP tasks.
However, unlike \citeauthor{li2020train}, our scaling law suggests starting too large is detrimental for the CIFAR-10 ResNets, leading to a higher parameter-count at error $\epsilon_k$.

Figure \ref{fig:misc} (left) illustrates this behavior concretely: it shows the error predicted by our scaling law for CIFAR-10 ResNets with varying widths.
The dotted black line shows the minimal parameter-count at which we predict it is possible to achieve each error.
Importantly, none of the low-error plateaus intersect this black dotted line, meaning a model cannot be minimal until it has been pruned to the point where it increases in error.
This occurs because the transitions of our functional form are gradual.
On the other hand, if we start with a model that is too large, it will no longer be on the black line when it has been pruned to the point where its error reaches $\epsilon_{np}(l, w)$; this behavior occurs because error decreases as a function of the invariant $m^*$ rather than the parameter-count $m$ and because $m~~\cancel{\propto}~~m^*$.
In Figure \ref{fig:misc} (right), we plot the same information from the actual CIFAR-10 data and see the same phenomena occur in practice.
The difference between the estimated and actual optimal parameter count is no more than 25\%.\fTBD{Need to reference some evidence}

Looking ahead, there are many directions for future work.
%Now that we have the function form, we can fit it for a new network family or a new pruning algorithm with a small number of configurations (Appendix \ref{app:interpolation}).
% to extend the generality---and thereby, utility---of this framework.
% For example, by studying other classes of networks and tasks, we can confirm that our scaling law applies in these settings or revise it to include this new data.
% By studying other pruning methods, we can further understand which aspects of our framework (e.g., the scaling law itself, the invariant, the design approach) generalize.
% Finally, we can explore fitting our scaling law to small-scale settings and extrapolating upwards to larger networks and datasets, which would make it possible to reason analytically without ever having to train a large-scale network.
Further studying sources of systematic error (transition shape and error improvements on the low-error plateau) is a promising avenue for making it possible to \emph{extrapolate} from small-scale settings to large-scale settings (see Appendix \ref{app:more_extrapolations} for a forward-looking discussion). 
Furthermore, while we focus on CIFAR-10 and ImageNet ResNets in the main body, it is important to understand the generality of our functional form for other networks and tasks (see Appendix \ref{app:more_arch_alg}).
Finally, now that we have described the structure of the error of IMP-pruned networks, it will be valuable to study the nature of scaling laws that capture the behavior of the plethora of other pruning methods that achieve different tradeoffs between parameter-count and error.

\section*{Acknowledgements}

Jonathan Rosenfeld was funded by the Efficient NN Model Scaling grant from Analog Devices Incorporated. This work was partially supported by NSF grants CCF-1563880 and IIS-1607189

We gratefully acknowledge the support of Google, which provided us with the TPU resources necessary to conduct our experiments through the TPU Research Cloud. %In particular, we express our gratitude to Zak Stone.

We gratefully acknowledge the support of IBM, which provided us with access to the GPU resources necessary to conduct our experiments on CIFAR-10 through the MIT-IBM Watson AI Lab. %In particular, we express our gratitude to David Cox and John Cohn.

This work was supported in part by cloud credits from the MIT Quest for Intelligence.

This work was supported in part by the Office of Naval Research (ONR N00014-17-1-2699).

This work was supported in part by DARPA award \#HR001118C0059.

We thank Yonatan Belinkov and Roy Shilkrot for their constructive comments.

\bibliography{main}
\bibliographystyle{icml2021}

\newpage

\appendix

\onecolumn

\section*{Contents of Appendices}

\textbf{Appendix \ref{app:pruningalg}.} Details on the IMP pruning algorithm.

\textbf{Appendix \ref{app:resnets}.} Details on the models, datasets, and training hyperparameters.

\textbf{Appendix \ref{app:sec3-key-observations-alldimensions}.} Full data for the observations we make in Section \ref{sec:single-network}.

\textbf{Appendix \ref{app:more_fits}.} Partial fits (e.g., just width, depth, or dataset size) for the joint scaling law in Section \ref{sec:joint}.

\textbf{Appendix \ref{app:more_arch_alg}.} Demonstrating that our functional form applies to additional networks and datasets.

\textbf{Appendix \ref{app:more_extrapolations}.} A discussion of extrapolation for our scaling law.

\textbf{Appendix \ref{app:comparison-to-rosenfeld}.} A more detailed comparison between our scaling law and that of \citet{rosenfeld2020a} following up on Section \ref{sec:design}.

\textbf{Appendix \ref{app:magic-one-percent}.} How we computed the effect of error dips on our estimator in Section \ref{sec:design}.

\newpage

\section{Formal Statement of Iterative Magnitude Pruning}
\label{app:pruningalg}

{
\begin{algorithm}[H]
    \small
    \caption{Iterative Magnitude Pruning (IMP) with weight rewinding to epoch 10 and $N$ iterations.}
    \begin{algorithmic}[1]
    \State Create a neural network with randomly initialized weights $W_0 \in \mathbb{R}^d$ and initial pruning mask $m = 1^{d}$
    \State Train $W_0$ to epoch 10, resulting in weights $W_{10}$
    \For{$n \in \{1, \ldots, N\}$}
    \State Train $m \odot W_{10}$ (the element-wise product of $m$ and $W_{10}$) to final epoch $T$ and weights $m \odot W_{T, n}$
    \State Prune the 20\% of weights in $m \odot W_{T, n}$ with the lowest magnitudes. $m[i] = 0$ if $W_{T, n}[i]$ is pruned
    \EndFor
    \State Return $m$ and $W_{T, n}$ 
    \end{algorithmic}
    \label{alg:imp}
\end{algorithm}
}

% \subsection{SynFlow}

% Unlike IMP, SynFlow is a pruning algorithm that prunes neural networks \emph{before} any training has taken place \citep{tanaka2020pruning}.
% To do so, SynFlow computes the ``synaptic strengths'' of each connection and prunes those weights with the lowest synaptic strengths (see Algorithm \ref{alg:synaptic-strength} below for the details on computing the synaptic strengths).

% Importantly, SynFlow prunes iteratively. It prunes a small number of weights, recalculates the synaptic strengths once those weights have been fixed to zero, and then prunes again.
% To prune to sparsity $s$, SynFlow iteratively prunes from sparsity $s^{n-1 \over 100}$ to sparsity $s^{n \over 100}$ for $n \in \{1, \ldots, 100\}$.

% After pruning, SynFlow trains the network normally using the standard hyperparameters. SynFlow computes the synaptic strengths as follows:\footnote{\texttt{https://github.com/ganguli-lab/Synaptic-Flow}}
% \begin{algorithm}[H]
%     \small
%     \caption{Computing the synaptic strengths of a network with weights $W$.}
%     \begin{algorithmic}[1]
%     \State Replace all weights $w \in W$ with their magnitudes $|w|$.
%     \State Forward propagate an input of all 1's
%     \State Take the sum of the logits $R$.
%     \State The synaptic strength for each weight $w$ is the gradient ${dR \over dw}$.
%     \end{algorithmic}
%     \label{alg:synaptic-strength}
% \end{algorithm}

% Note that this algorithm leads to exploding activations on deeper networks, so we do not vary network depths in any of our experiments involving SynFlow.

\newpage
\section{Experimental Details} \label{app:resnets}

\subsection{ResNets}

We study the residual networks (ResNets) designed by \citet{he2016deep} for CIFAR-10 and ImageNet.
ResNets for CIFAR-10 are composed of an initial convolutional layer, three sets of $B$ residual blocks (each with two convolutional layers and a skip connection), and a linear output layer.
The sets of blocks have 16, 32, and 64 convolutional channels, respectively.

ResNets for ImageNet and TinyImageNet are composed of an initial convolutional layer, a max-pooling layer, four sets of residual blocks (each with three convolutional layers and a skip connection), and a linear output layer.
The sets of blocks have 64, 128, 256, and 512 convolutional channels, respectively.
On ImageNet, we use a ResNet with 50 layers. On TinyImageNet, we use a ResNet with 18 layers. Both choices are standard for these datasets.

We place batch normalization before the ReLU activations.

To vary the width of the networks, we multiply the number of convolutional channels by the width scaling factor $w$.
To vary the depth of the CIFAR-10 ResNets, we vary the value of $B$.
The depth $l$ of the network is the total number of the layers in the network, not counting skip connections.

\subsection{VGG Networks}

We study the VGG-16 variant of the VGG networks for CIFAR-10 as provided by the OpenLTH repository.\footnote{\texttt{github.com/facebookresearch/open\_lth}}
The network is divided into five sections, each of which is followed by max pooling with kernel size 2 and stride 2.
The sections contain 3x3 convolutional layers arranged as follows:

\begin{center}

\begin{tabular}{c c c}
\toprule
Section & Width & Layers \\
\midrule
1 & 64 & 2 \\
2 & 128 & 2 \\
3 & 256 & 3 \\
4 & 512 & 3 \\
5 & 512 & 3 \\
\bottomrule
\end{tabular}
\end{center}

The network has ReLU activations and batch normalization before each activation.
To vary the width of VGG-16, we multiply each of the per-segment widths by the width scaling factor $w$.

\subsection{DenseNets}

We study the densely connected residual networks (DenseNets) designed by \citet{he2016deep} for CIFAR-10.
DenseNets for CIFAR-10 are composed of an initial convolutional layer, four sets of dense blocks, and a linear output layer.
Between the sets of blocks are transition layers of 1x1 convolutions and an average pooling operation that downsamples the image by 2x.
Each block comprises a 1x1 convolution that increases the channel count by 4x and a 3x3 block that decreases it to a fixed constant size $g$; this output is then concatenated to the input of the block.
As such, if the input to the block has $n$ channels, the output of the block has $n + g$ channels.
We use DenseNet-121, which has sets of 6, 12, 24, and 16 blocks.
$g$ is set to 16 but is multiplied by the width scaling factor $w$ to modify the width.

\subsection{Training Hyperparameters}

We train CIFAR-10 and SVHN ResNets and VGG-16 for 160 epochs with a batch size of 128.
The initial learning rate is 0.1, and it drops by an order of magnitude at epochs 80 and 120.
We optimize using SGD with momentum (0.9).
We initialize with He uniform initialization.
CIFAR-10 data is augmented by normalizing, randomly flipping left and right, and randomly shifting by up to four pixels in any direction (and cropping afterwards).
SVHN data is not augmented.

We train CIFAR-10 DenseNets with the same hyperparameters but for 200 epochs (with learning rate drops at 130 and 165 epochs).
We train SVHN DenseNets with the same hyperparameters but for 100 epochs (with learning rate drops at 70 and 85 epochs).

We train ImageNet ResNets for 90 epochs with a batch size of 1024.
The initial learning rate is 0.4, and it drops by an order of magnitude at epochs 30, 60, and 80.
We perform linear learning rate warmup from 0 to 0.4 over the first 5 epochs.
We optimize using SGD with momentum (0.9).
We initialize with He uniform initialization.
Data is augmented by normalizing, randomly flipping left and right, selecting a random aspect ratio between 0.8 and 1.25, selecting a random scaling factor between 0.1 and 1.0, and cropping accordingly.

We train TinyImageNet ResNets identically except for that we train them for 200 epochs (with learning rate drops at 100 and 150 epochs) and a learning rate of 0.2.
Augmentation is identical to ImageNet.

\subsection{Dimensions}

We use the following dimensions for the additional experiments. We select configurations using the same methodology as in Table 1.

\begin{center}
{\scriptsize
%\begin{tabular}{@{\ }l@{\ }|@{\ }c@{\ \ }c@{\ }|@{\ }c@{\ }|@{\ }c@{\ }|@{\ }c@{\ }|@{\ }c}
\begin{tabular}{@{\ }l@{\ }|@{\ }c@{\ }|@{\ }c@{\ }|@{\ }c@{\ }|@{\ }c}
\toprule
Network Family &
    %$N_{\text{train}}$ &
    %$N_{\text{test}}$ &
    Densities ($d$) &
    Depths ($l$) &
    Width Scalings ($w$) &
    Subsample Sizes ($n$) \\ \midrule
CIFAR-10/SVHN ResNet &
    %50K &
    %10K &
    $0.8^i, i \subseteq \{0, \ldots, 40\}$ &
    $l \subseteq$ \{8, 14, 20, 26, 50, 98\} &
    $2^i, i \subseteq \{-4, \ldots, 2\}$ &
    $\frac{N}{i}$, $i \in \{1, 2, 4, 8, 16, 32, 64\}$ \\
CIFAR-10/SVHN VGG &
    $0.8^i, i \subseteq \{0, \ldots, 37\}$ &
    16 &
    $2^i, i \subseteq \{-4, \ldots, 0\}$ &
    $\frac{N}{i}$, $i \in \{1\}$ \\
CIFAR-10/SVHN DenseNet &
    $0.8^i, i \subseteq \{0, \ldots, 50\}$ &
    121 &
    $2^i, i \subseteq \{-4, \ldots, 1\}$ &
    $\frac{N}{i}$, $i \in \{1\}$ \\
TinyImageNet ResNet &
    $0.8^i, i \subseteq \{0, \ldots, 50\}$ &
    18 &
    $2^i, i \subseteq \{-6, \ldots, 0\}$ &
    $\frac{N}{i}$, $i \in \{1\}$ \\
ImageNet ResNet &
    %1.28M &
    %50K &
    $0.8^i, i \subseteq \{0, \ldots, 30\}$ &
    50 & 
    $2^i, i \subseteq \{-4, \ldots, 0\}$ &
    $\frac{N}{i}$, $i \in \{1, 2, 4\}$ \\ 
\bottomrule
\end{tabular}
}
\end{center}

\clearpage

\section{Full Data for Key Observations in Section \ref{sec:single-network}}
\label{app:sec3-key-observations-alldimensions}

\begin{figure}[h!]
\centering % <-- added
\begin{minipage}{0.34\textwidth}
\includegraphics[width=\linewidth,trim={0.2cm 0 0.1cm 0.65cm},clip]{figures/pruning_curves_widths.pdf}
\end{minipage}\hfil % <-- added
\begin{minipage}{0.295\textwidth}
  \includegraphics[width=\linewidth,trim={2cm 0 0 0.6cm},clip]{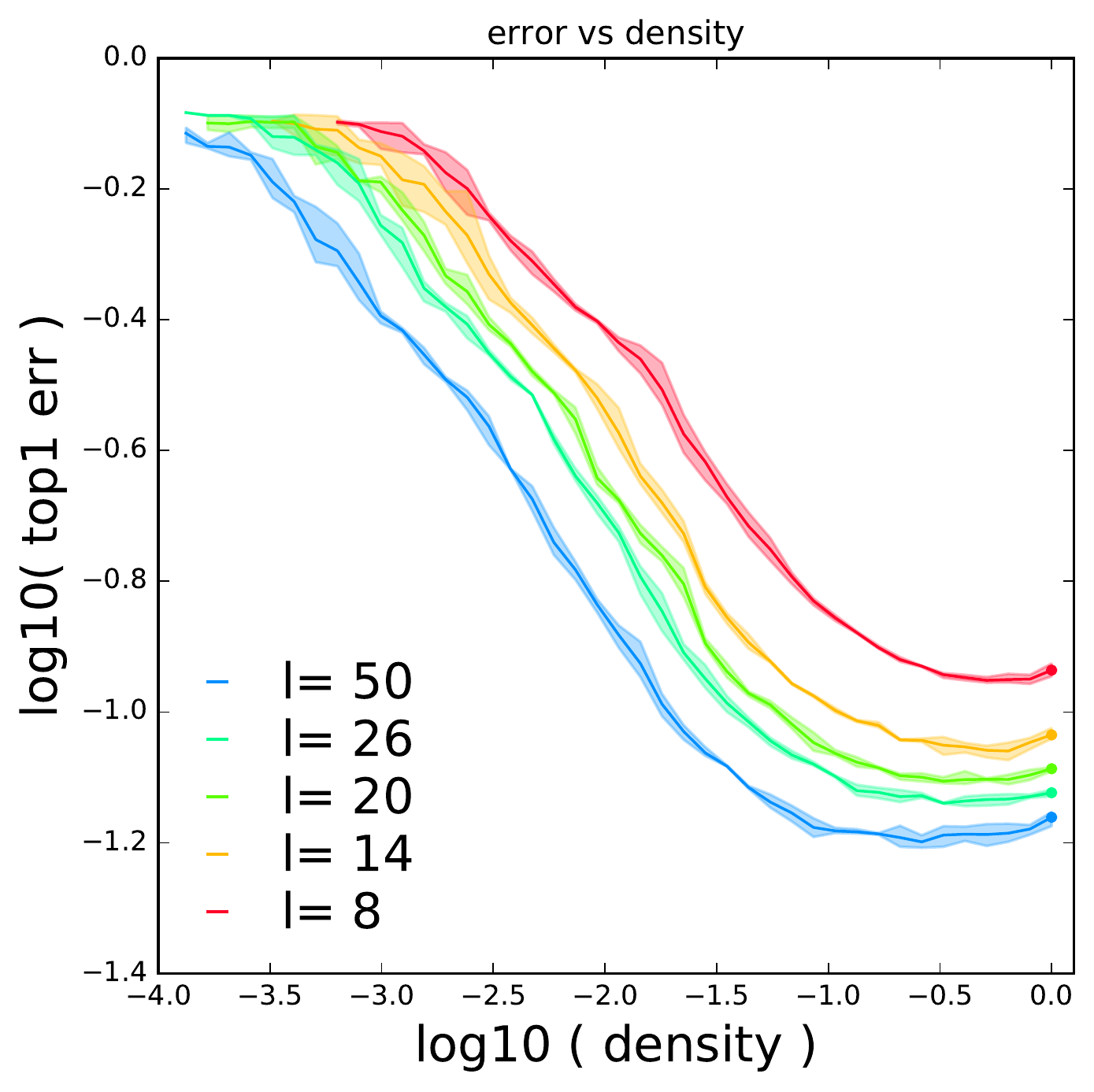}
\end{minipage}\hfil % <-- added
\begin{minipage}{0.295\textwidth}
  \includegraphics[width=\linewidth,trim={2cm 0 0 0.6cm},clip]{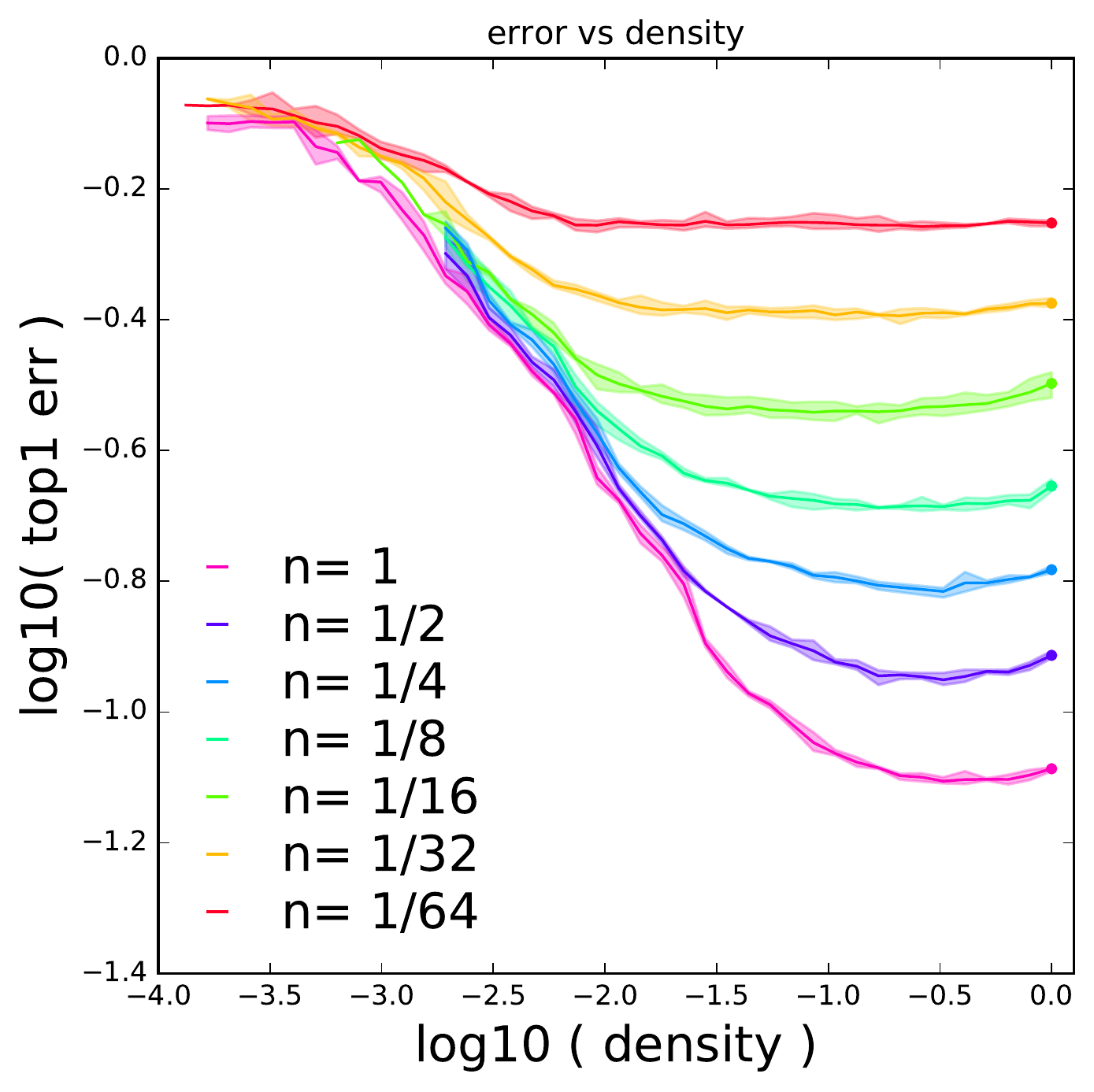}
\end{minipage}
\caption{Relationship between density and error when pruning CIFAR-10 ResNets and varying $w$ (left, $l=20$, $n=N$),  $l$ (center, $w=1$, $n=N$), $n$ (right, $l=20$, $w=1$) }
\vspace{0mm}
\label{fig:cifar_observations_all}
\end{figure}

\begin{figure}[h!]
\centering % <-- added
\begin{minipage}{0.34\textwidth}
\includegraphics[width=\linewidth,trim={0.2cm 0 0.1cm 0.65cm},clip]{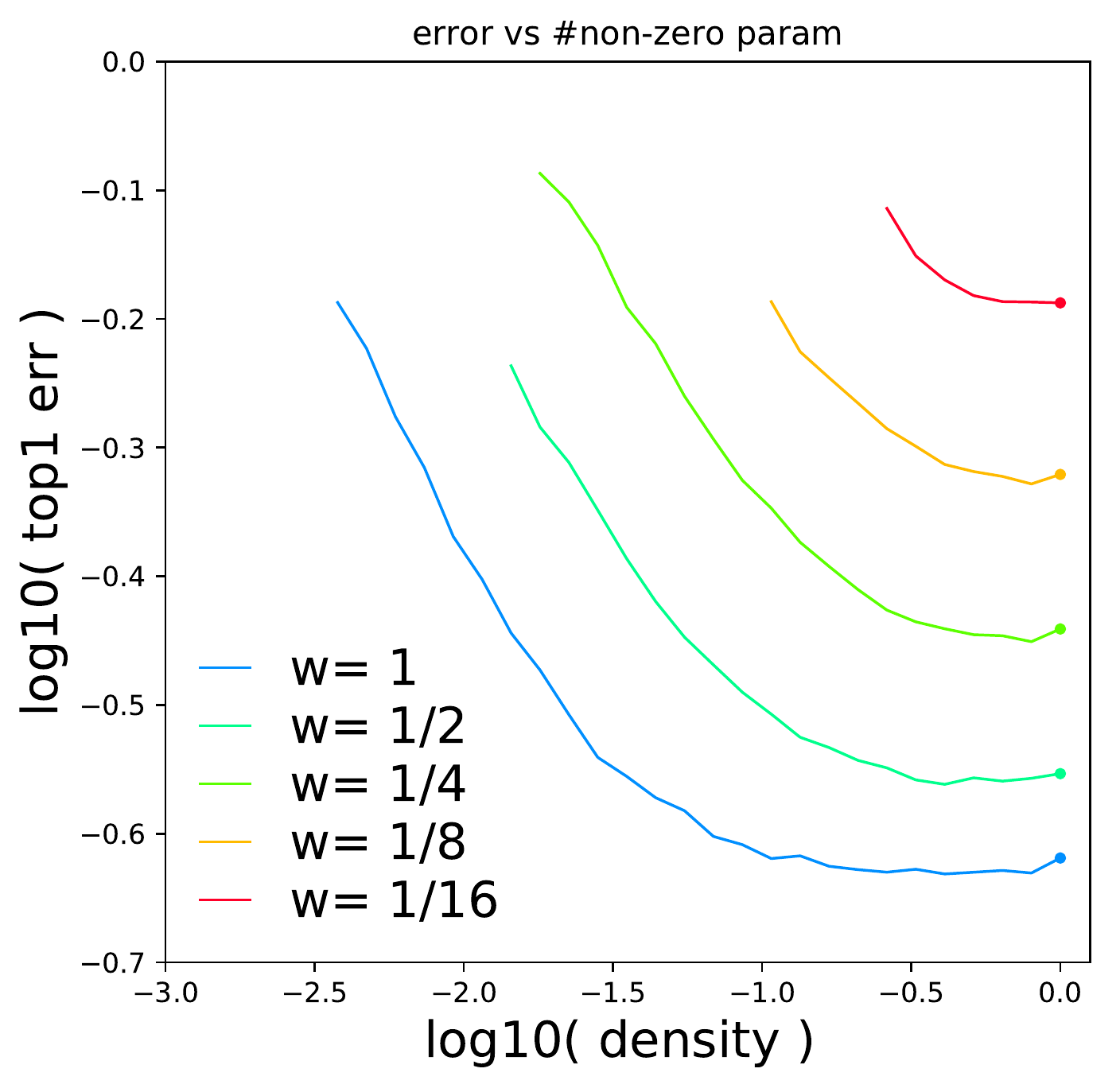}
\end{minipage}%\hfil % <-- added
\begin{minipage}{0.3\textwidth}
  \includegraphics[width=\linewidth,trim={2cm 0 0 0.6cm},clip]{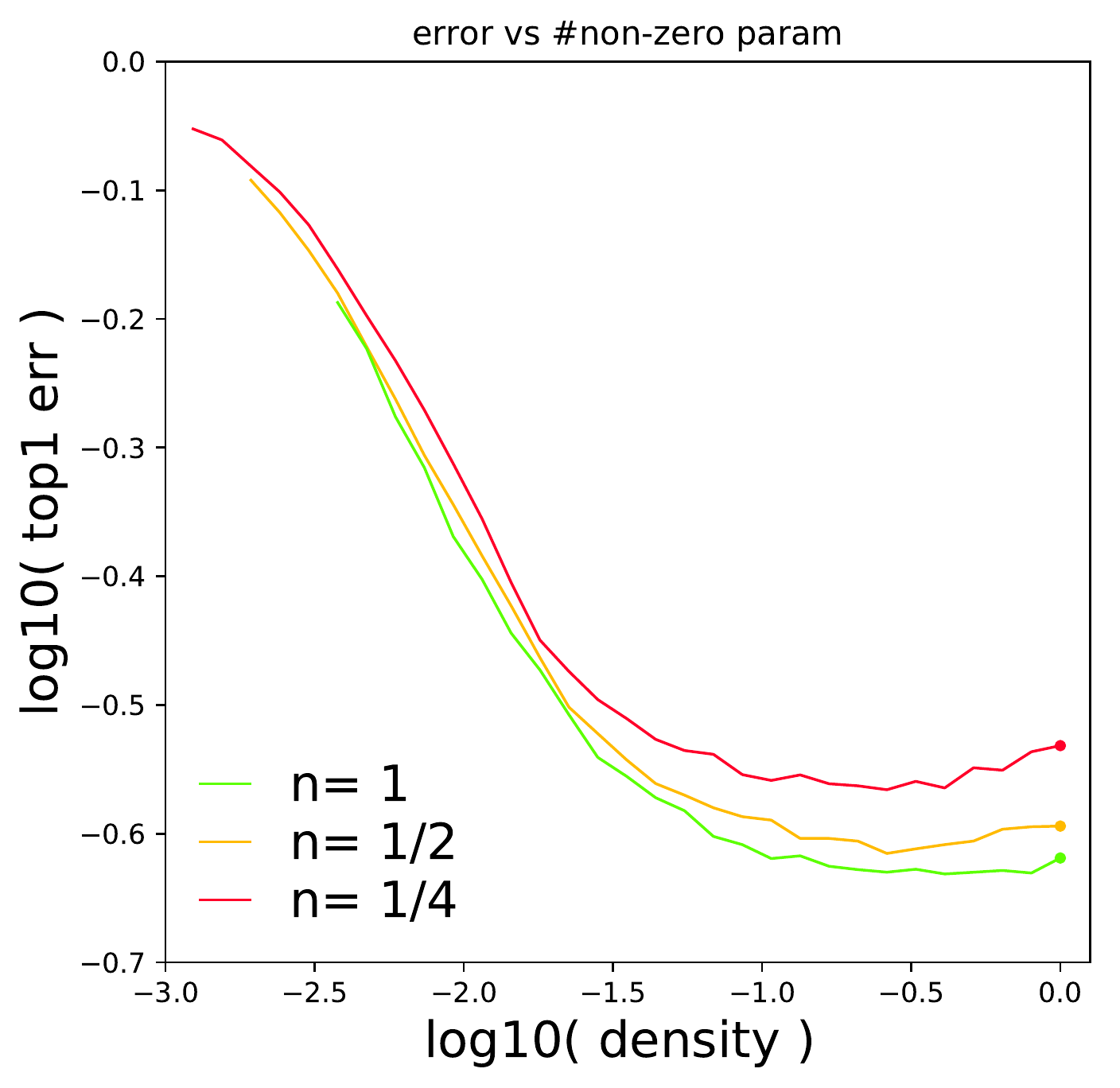}
\end{minipage}%\hfil % <-- added
% \begin{minipage}{0.29\textwidth}
%   \includegraphics[width=\linewidth,trim={2cm 0 0 0.6cm},clip]{figures/pruning_curves_data.pdf}
% \end{minipage}
\caption{Relationship between density and error when pruning Imagenet ResNet-50 and varying $w$ (left, $n=N$),  and $n$ (right, $w=1$) }
\vspace{0mm}
\label{fig:imagenet_observations_all}
\end{figure}

In this appendix, we show that our observations from Section \ref{sec:single-network} hold when varying all dimensions (depth, width, and dataset size) on both the CIFAR-10 and ImageNet ResNets for IMP.
Figure \ref{fig:cifar_observations_all} shows the error versus density when changing width (left) depth (center) and data (right). 
In Figure \ref{fig:imagenet_observations_all}, we similarly show the dependency of the error on density for Imagenet when varying width (left) and dataset size (right).

% Note that the high-error level is not explicitly reached, but rather - the power-law and lower error plateau are visible.
In Figure \ref{fig:cifar_observations_all}, we observe that all curves have a similar slope in the power-law region.
In Equation \ref{eq:prune_density}, this implies that while $\gamma$ is allowed to vary with $l$, $w$ and $n$, it is in practice approximately a constant.
Similarly, the high-error plateau $\epsilon^\uparrow$ is also shared across curves such that it too is approximately constant.
In contrast, the transition from high-error plateau to the power-law region is not constant as a function of density.
Section \ref{sec:joint} finds exactly this dependency of the transition parameter $p$.

\clearpage

\section{Partial (Projections) Fit Results for Section \ref{sec:joint}}
\label{app:more_fits}

In Section \ref{sec:joint}, we fit the error jointly as a function of all dimensions showing that Equation \ref{eq:intermediate_state} provides a good approximation to the error in practice.
In this appendix, we consider important sub-cases, such as the case when one wishes to scale only one degree of freedom while pruning. This serves both a practical scenario, but also allows for a qualitative visualization of the fit (and typical sources of error), which is otherwise difficult to perform over all dimensions jointly.
From a practical standpoint, in this case one need not estimate the parameters associated with the fixed degree of freedom.

Recall that, given the non-pruned network error $\epsilon_{np}$, all dependencies on the individual structural degrees of freedom $l,w$ are captured by the invariant $m^* \triangleq l^\phi w^\psi d$.
This means that, if one wishes to estimate the error while pruning when holding width fixed, we need not estimate $\psi$.
Similarly if depth is held constant, we need not estimate $\phi$.

\begin{figure}[h!]
    \vspace{-2mm}
         \begin{minipage}{0.34\textwidth}
            \includegraphics[width=\linewidth,trim={0.2 0 0 0.65cm},clip]{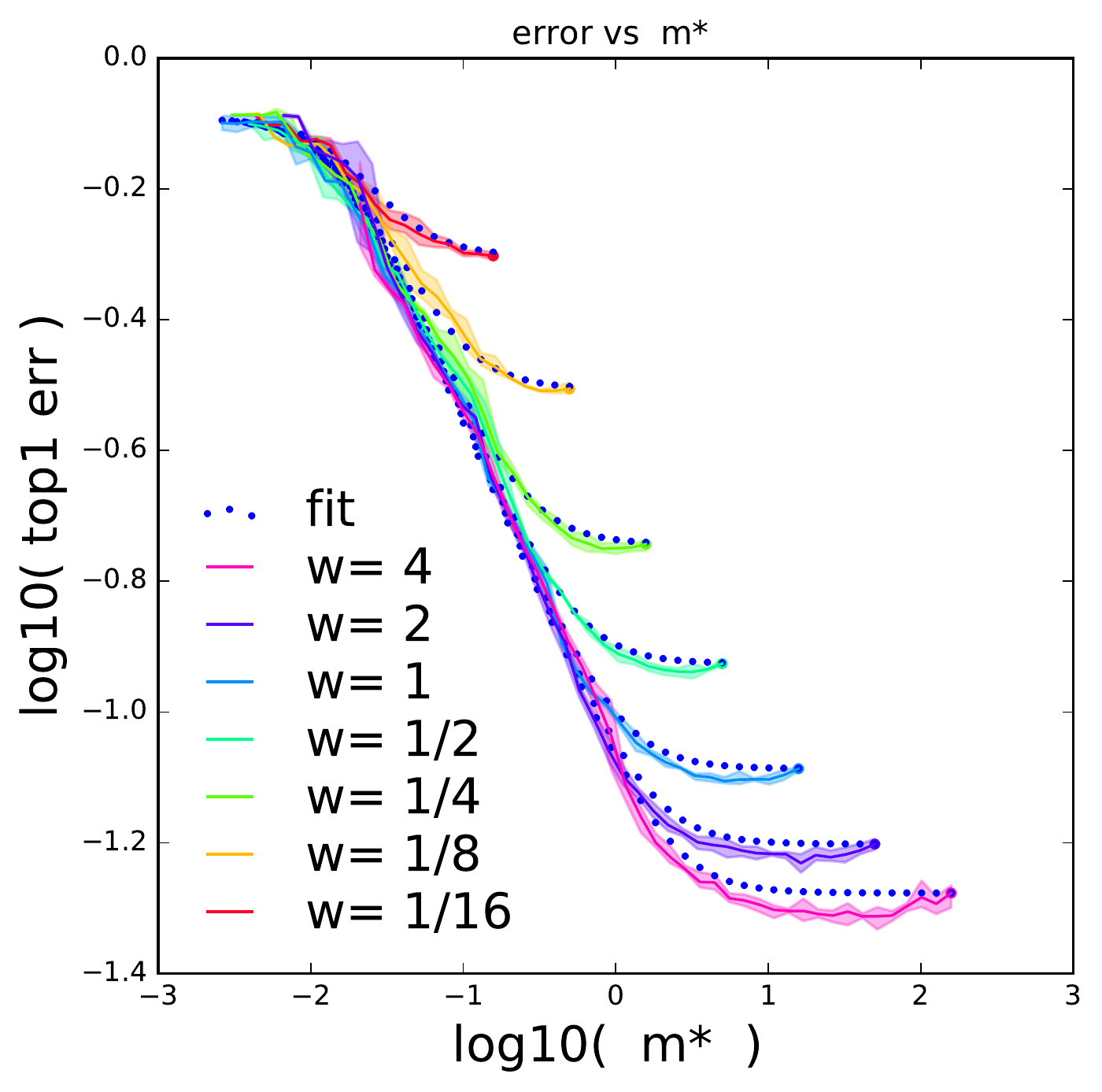}\llap{\makebox[2.15cm][l]{\raisebox{2.45cm}{\includegraphics[scale=0.17,trim={1.9cm 1.5cm 0.3cm 0.3cm},clip]{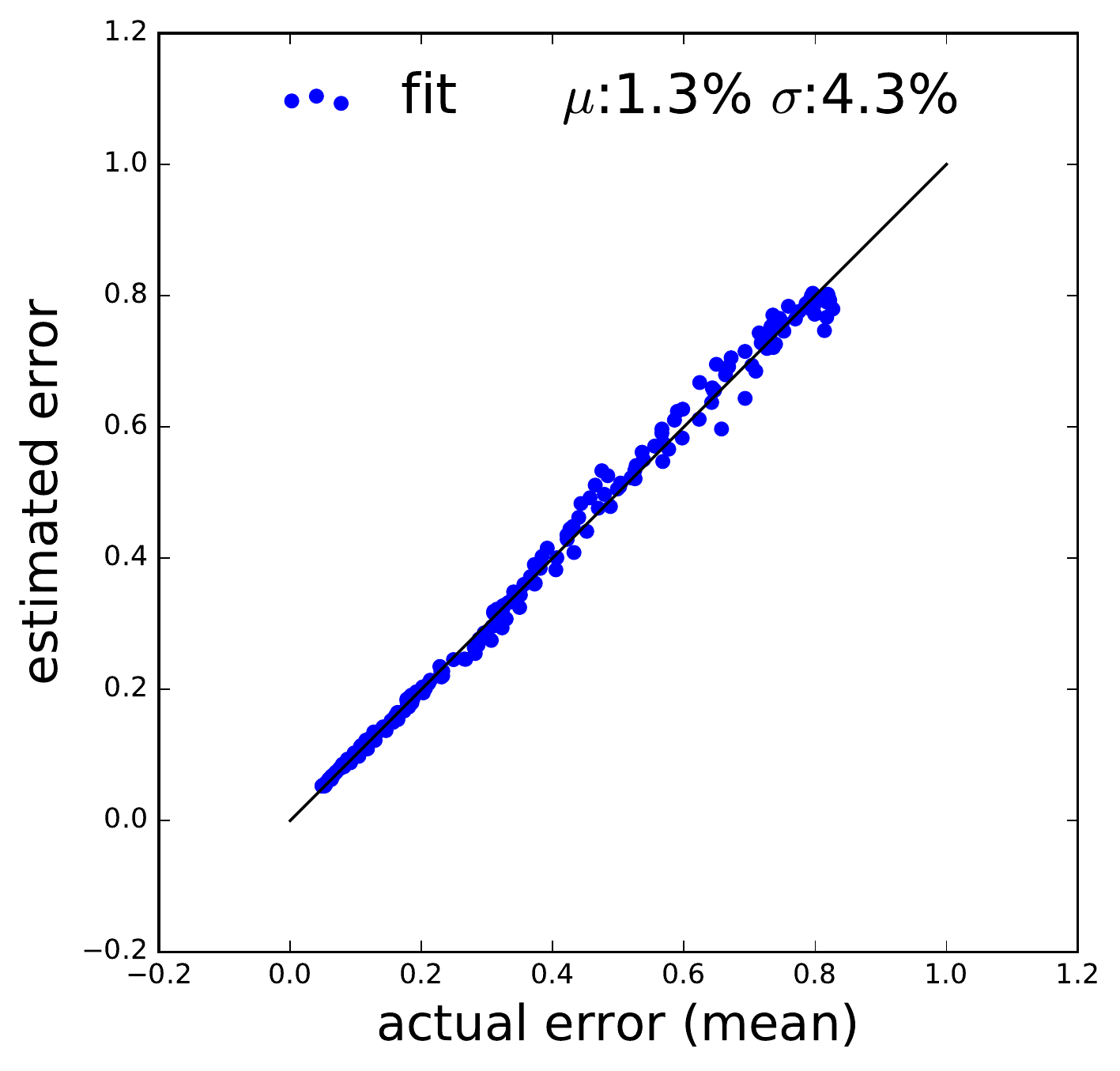}}}}
        \end{minipage}%
        \begin{minipage}{0.292\textwidth}
            \vspace{-0.0mm}
            \includegraphics[width=\linewidth,trim={2cm 0 0 0.65cm},clip]{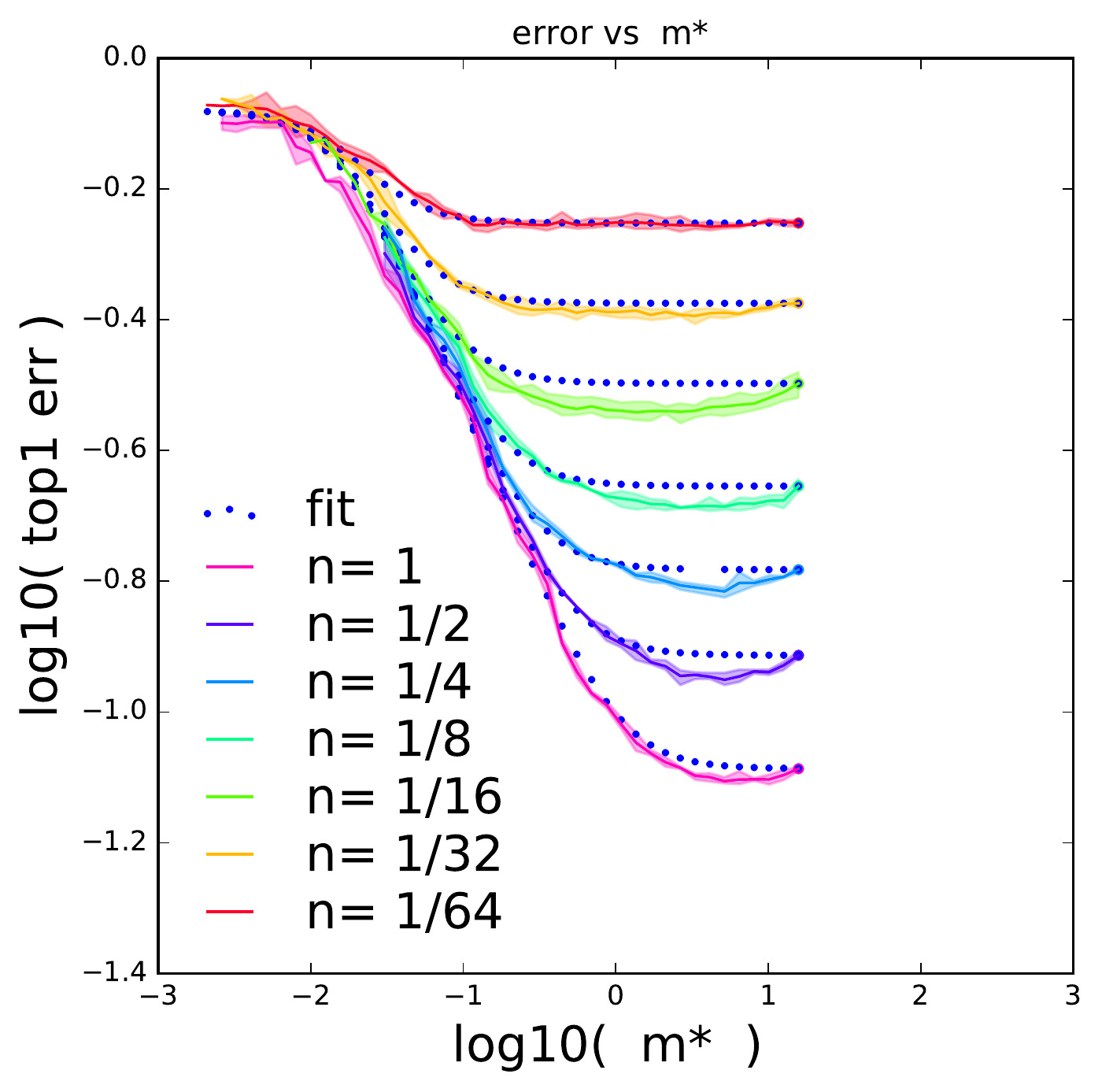}\llap{\makebox[2.25cm]{\raisebox{2.45cm}{\transparent{0.7}\includegraphics[scale=0.17,trim={1.9cm 1.5cm 0.3cm 0.3cm},clip]{figures/fit_corr_True,False,False.pdf}}}}
        \end{minipage}%
        \begin{minipage}{0.292\textwidth}
            \vspace{-0.0mm}
            \includegraphics[width=\linewidth,trim={2cm 0 0 0.65cm},clip]{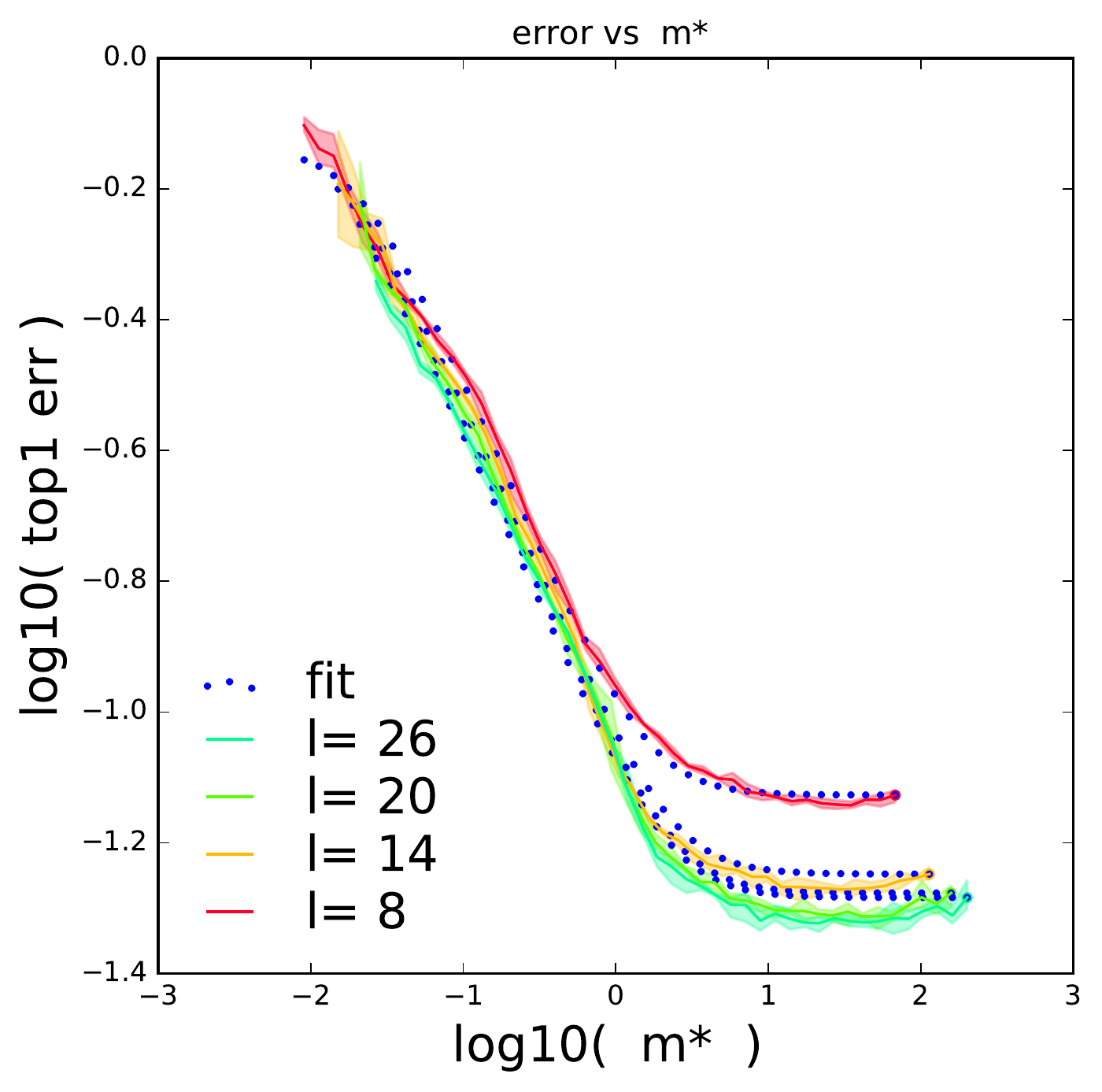}\llap{\makebox[2.25cm]{\raisebox{2.45cm}{\includegraphics[scale=0.17,trim={1.9cm 1.5cm 0.3cm 0.3cm},clip]{figures/fit_corr_True,False,False.pdf}}}}
        \end{minipage}
        
        % \vspace{-10pt}
         \begin{minipage}{0.338\textwidth}
            \includegraphics[width=\linewidth,trim={0.2 0 0.3cm 0.65cm},clip]{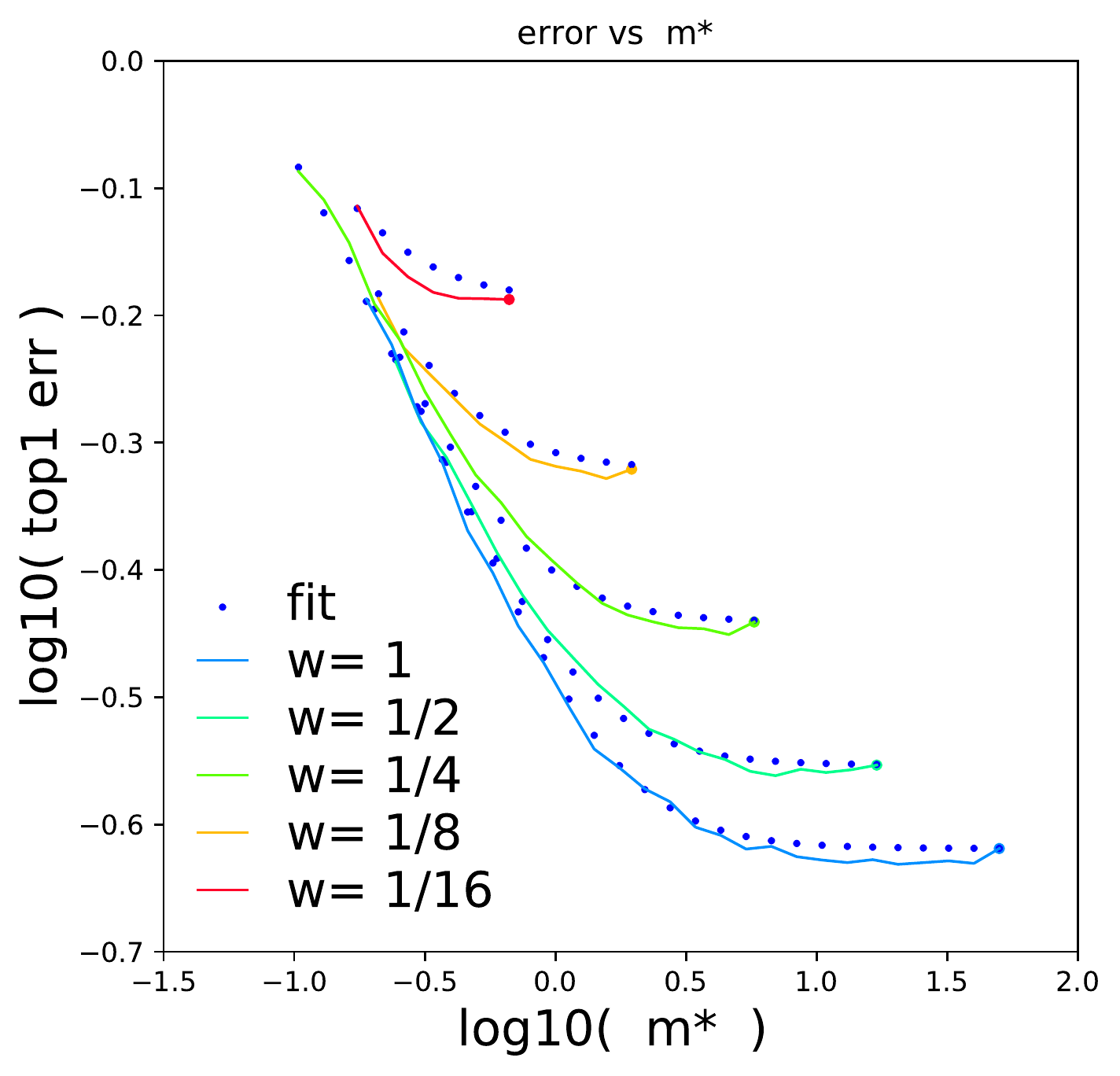}\llap{\makebox[2.1cm][l]{\raisebox{2.39cm}{\transparent{0.8}\includegraphics[scale=0.17,trim={1.8cm 1.5cm 0.3cm 0.3cm},clip]{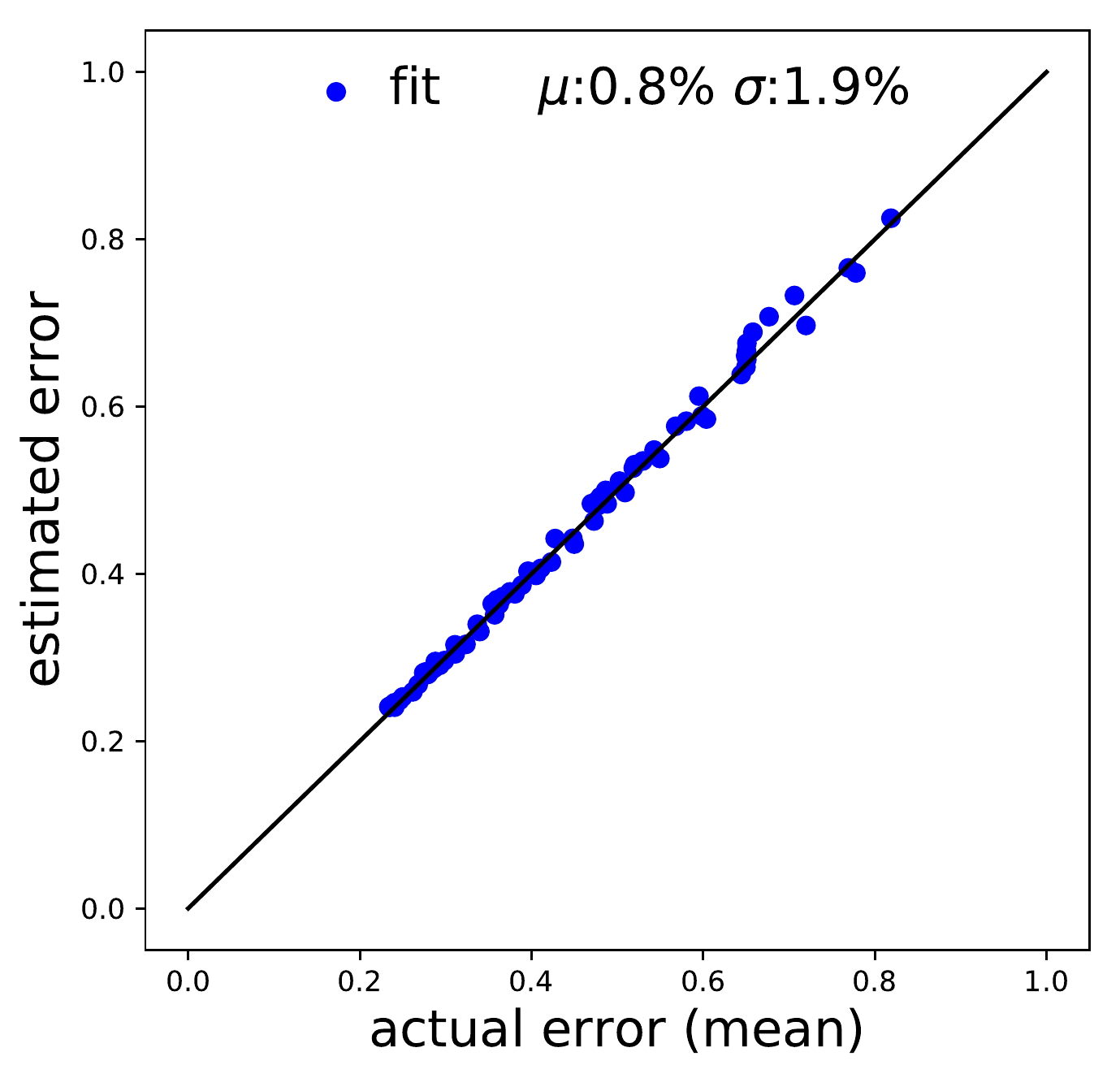}}}}
        \end{minipage}
        \begin{minipage}{0.3\textwidth}
            \includegraphics[width=\linewidth,trim={1.9cm 0 0 0.65cm},clip]{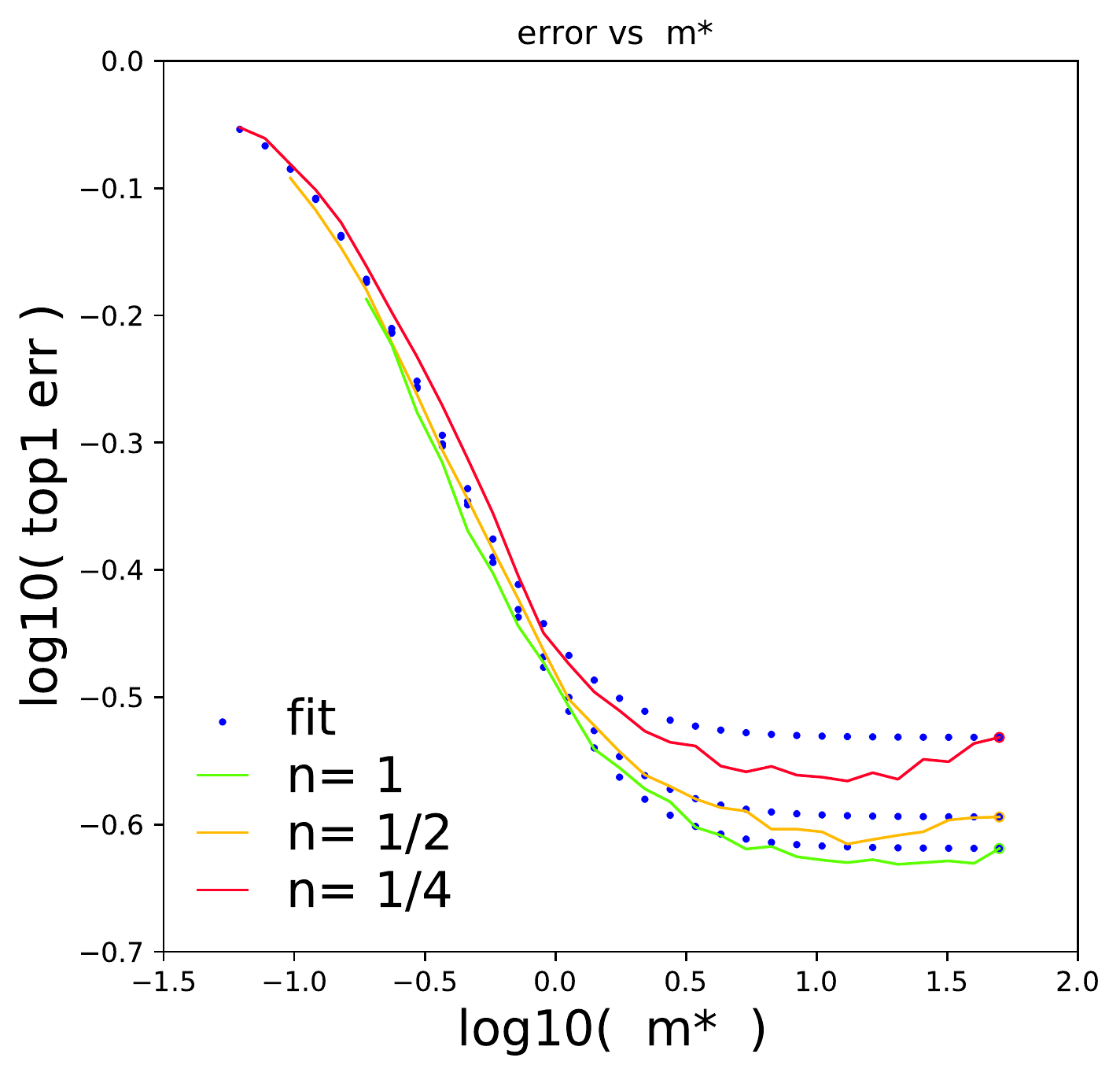}\llap{\makebox[2.2cm][l]{\raisebox{2.39cm}{\includegraphics[scale=0.17,trim={1.8cm 1.5cm 0.3cm 0.3cm},clip]{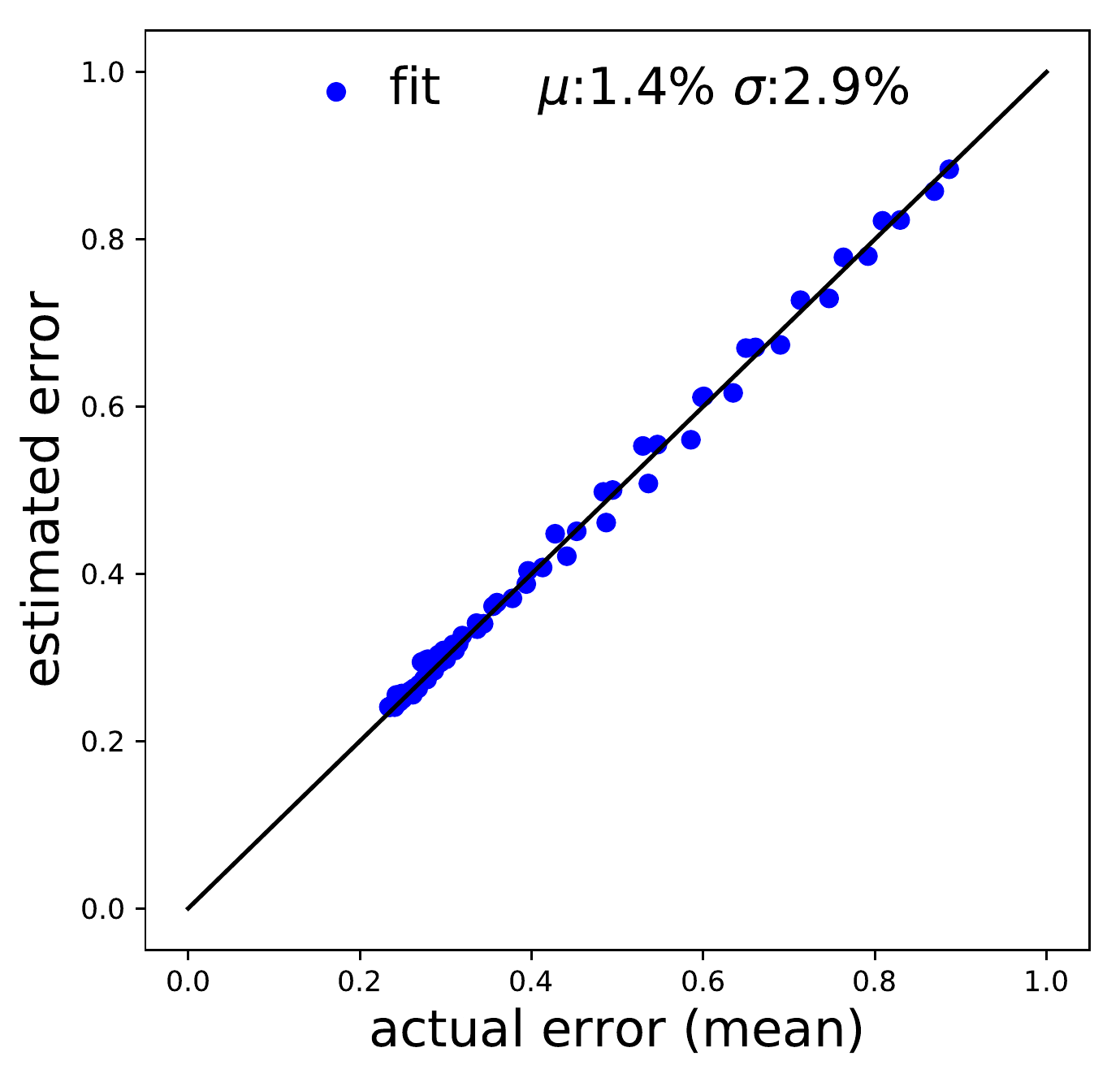}}}}
        \end{minipage}
        \vspace{-0em}
\vspace{-5pt}
\caption{Top row: CIFAR-10. Bottom row: ImageNet. Left: varying width. Center: varying dataset size. Right: varying depth. Lines are the actual error and dots are the estimated error.}
\label{fig:cifar_partial_fit}
\end{figure}

Figure \ref{fig:cifar_partial_fit} shows these partial fits.
Shown from left to right are the fits done while pruning and varying width, depth and data respectively. Correspondingly, these fits omit separately $\psi$ or $\phi$ or omit both when depth nor width are scaled.
The fits were performed with all available density points for each dimension. For CIFAR-10: 7 widths, 224 points for the width partial fit; 7 dataset fractions, 240 points for the data partial fit; 4 depths, 164 points for the depth partial fit. For ImageNet: 5 widths, 83 points for the width partial fit; 3 dataset fractions, 86 points for the data partial fit.

This exercise, apart from its practical implications, highlights the fact that there are in effect two groups of parameters comprising the estimation. The first are the parameters $\epsilon^\uparrow$, $\gamma$ and $p'$ which control the dependency as a function of density (or more generally, as a function of the invariant). The second are $\phi$ and $\psi$ which are properties of the architectural degrees of freedom captured by the invariant. 
Moreover, within the first group of parameters $\epsilon^\uparrow$, $\gamma$, can be isolated and found from a single pruning curve, as they are not a function of $l,w,n$.
%We suspect that these distinctions holds for other tasks and network types, although we leave this exploration to future work. 

\newpage

\section{Additional Architectures and Datasets}
\label{app:more_arch_alg}

In this appendix, we show that our functional form applies to additional pairs of networks and datasets: (CIFAR-10 ResNet, SVHN), (VGG-16, CIFAR-10), (VGG-16, SVHN), (DenseNet-121, CIFAR-10), (DenseNet-121, SVHN), (ImageNet ResNet-18, TinyImageNet). 

In general, we obtain good fits on CIFAR-10 and TinyImageNet.
On SVHN, fits are worse but the networks suffer from high measurement error (i.e., accuracy varies greatly between multiple runs at the same density) at low densities; nevertheless, fits are often better than measurement error because they average out some of the error.

\begin{figure}[h!]
\centering
\begin{minipage}{0.3\textwidth}
    \includegraphics[width=\linewidth,trim={0.2 0 0 0.65cm},clip]{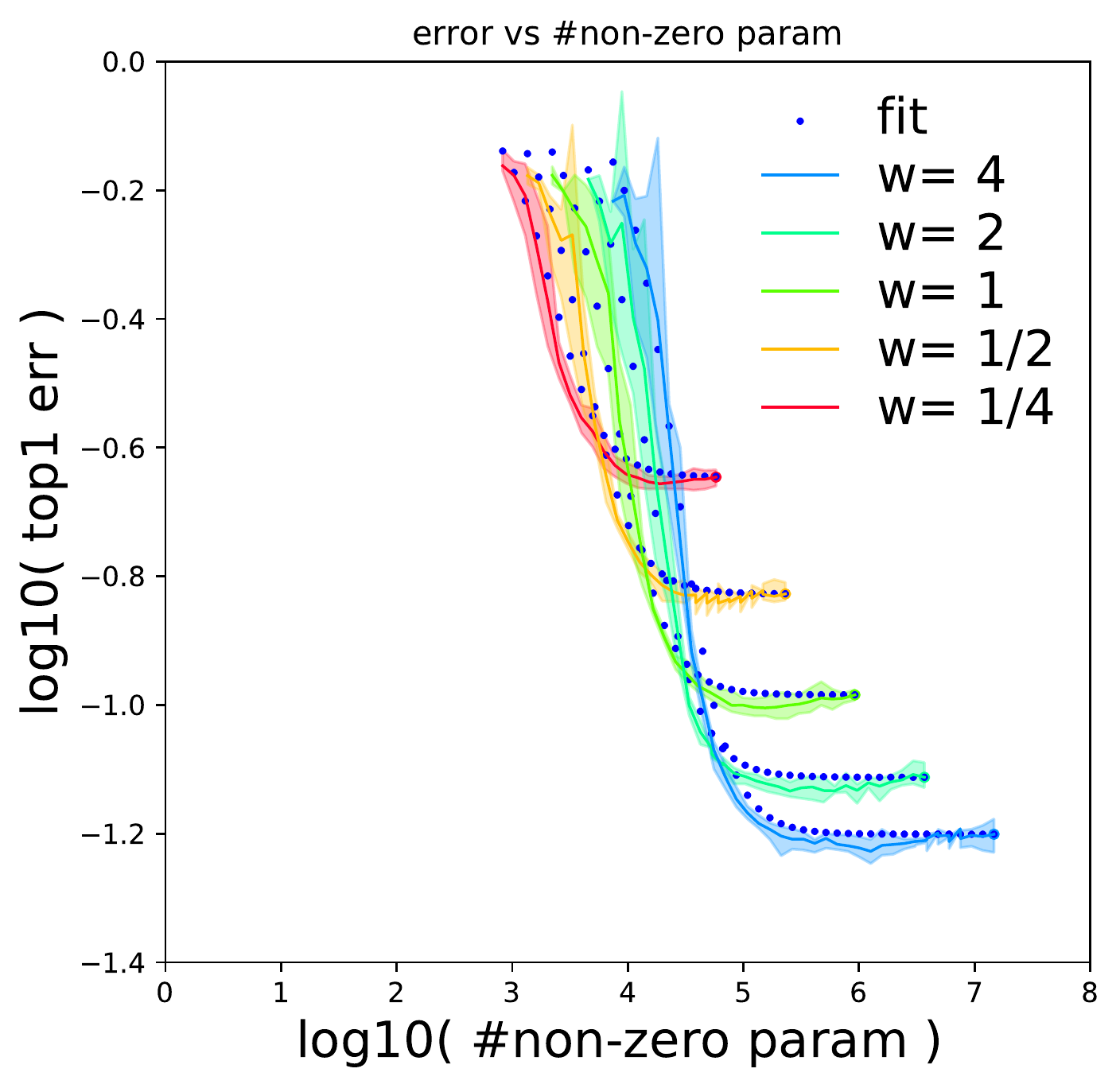}
\end{minipage}
\begin{minipage}{0.3\textwidth}
    \includegraphics[width=\linewidth]{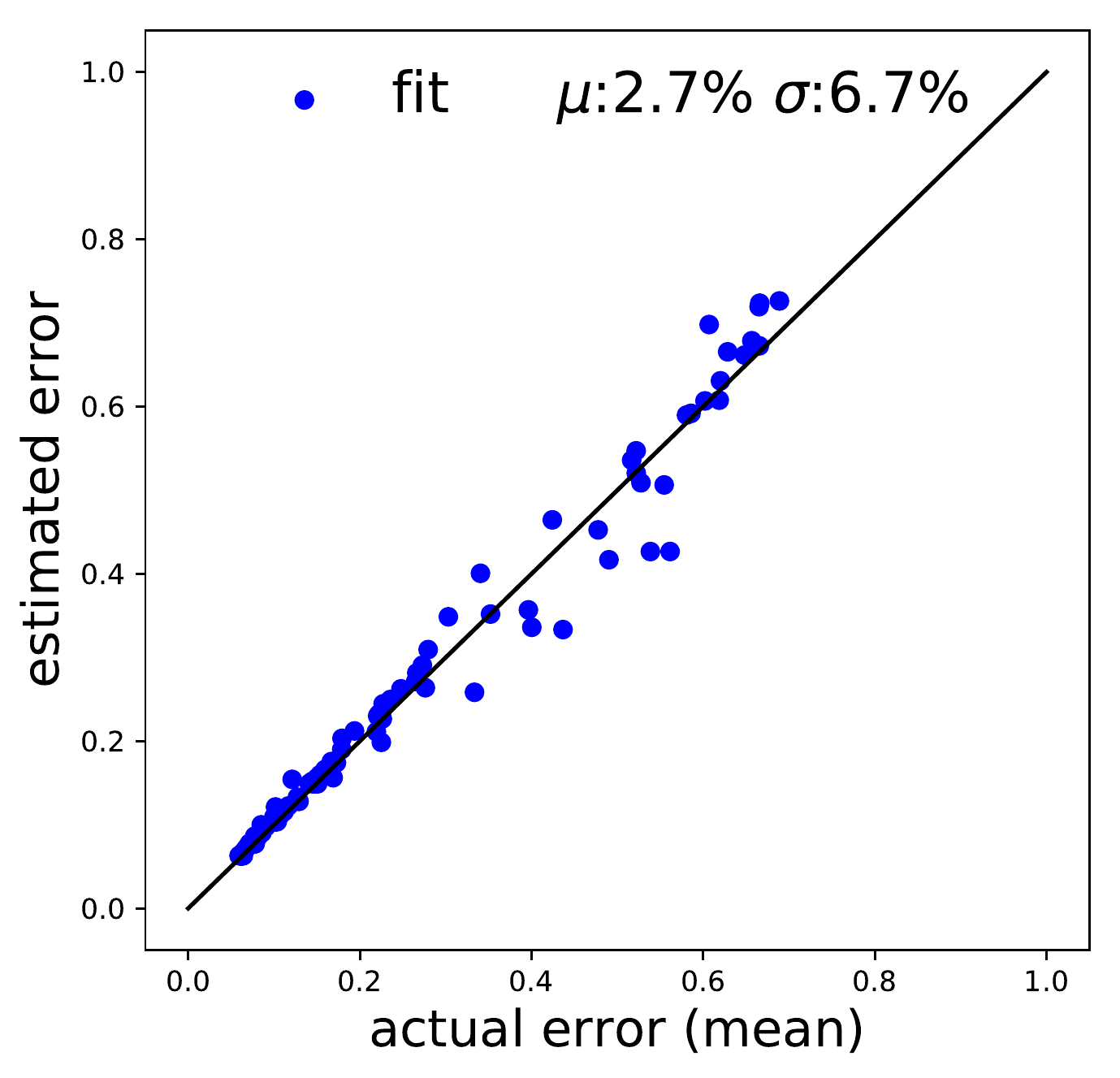}
\end{minipage}
\begin{minipage}{0.3\textwidth}
    \includegraphics[width=\linewidth]{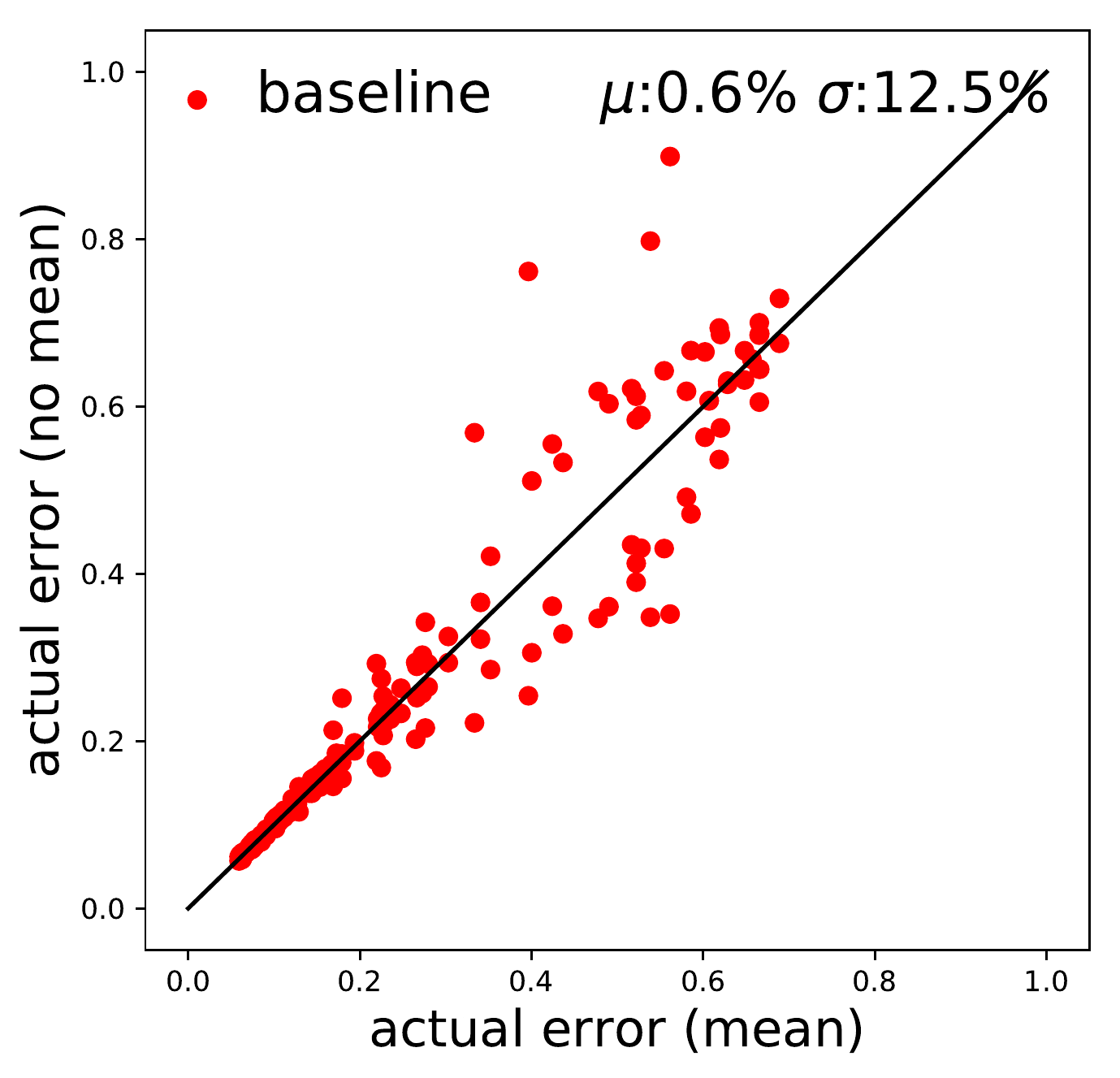}
\end{minipage}
\caption{Fit for VGG-16 on CIFAR-10 with IMP pruning.}
\vspace{0mm}
\label{fig:vgg-imp}
\end{figure}

\begin{figure}[h!]
\centering
\begin{minipage}{0.3\textwidth}
    \includegraphics[width=\linewidth,trim={0.2 0 0 0.65cm},clip]{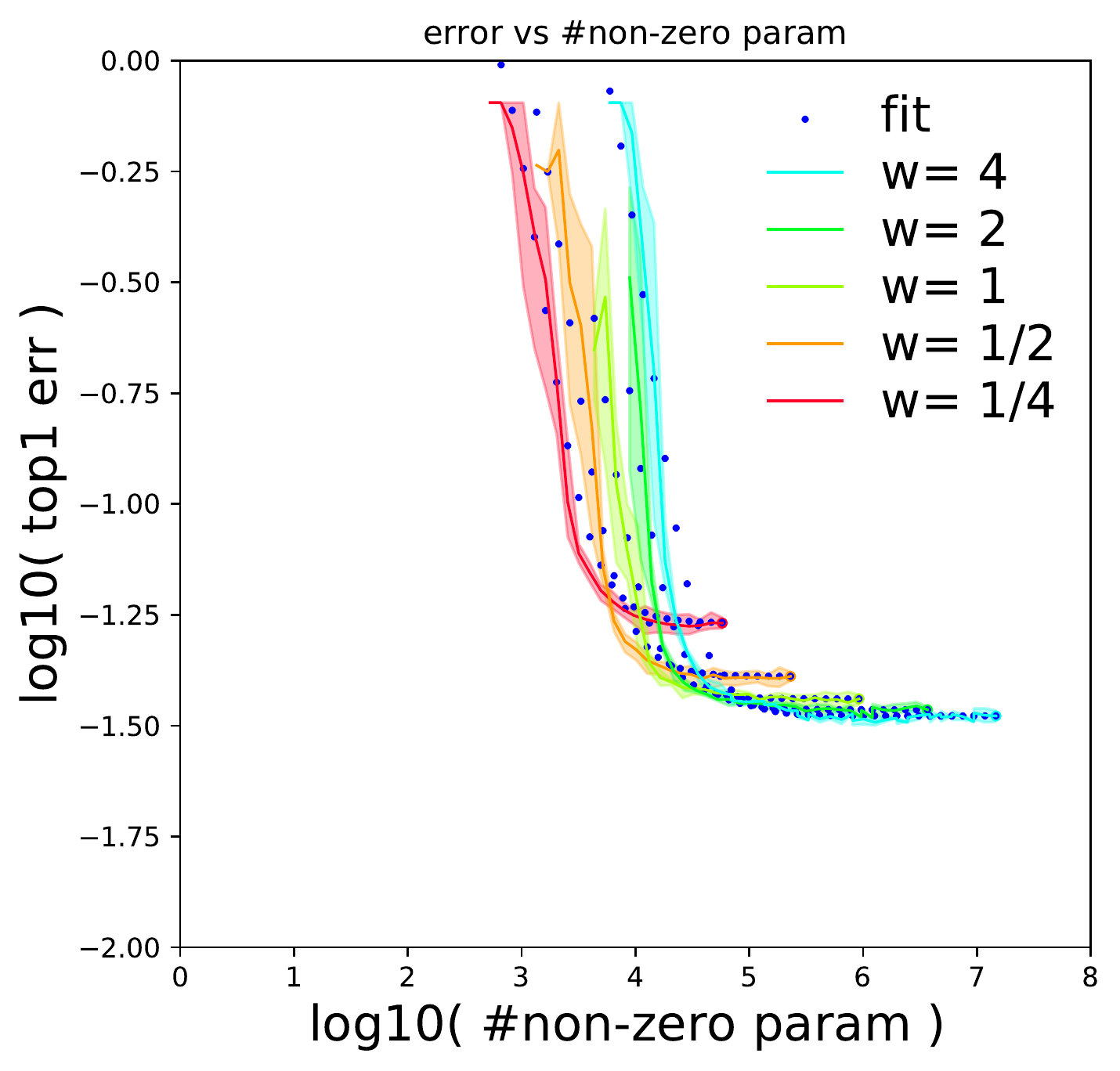}
\end{minipage}
\begin{minipage}{0.3\textwidth}
    \includegraphics[width=\linewidth]{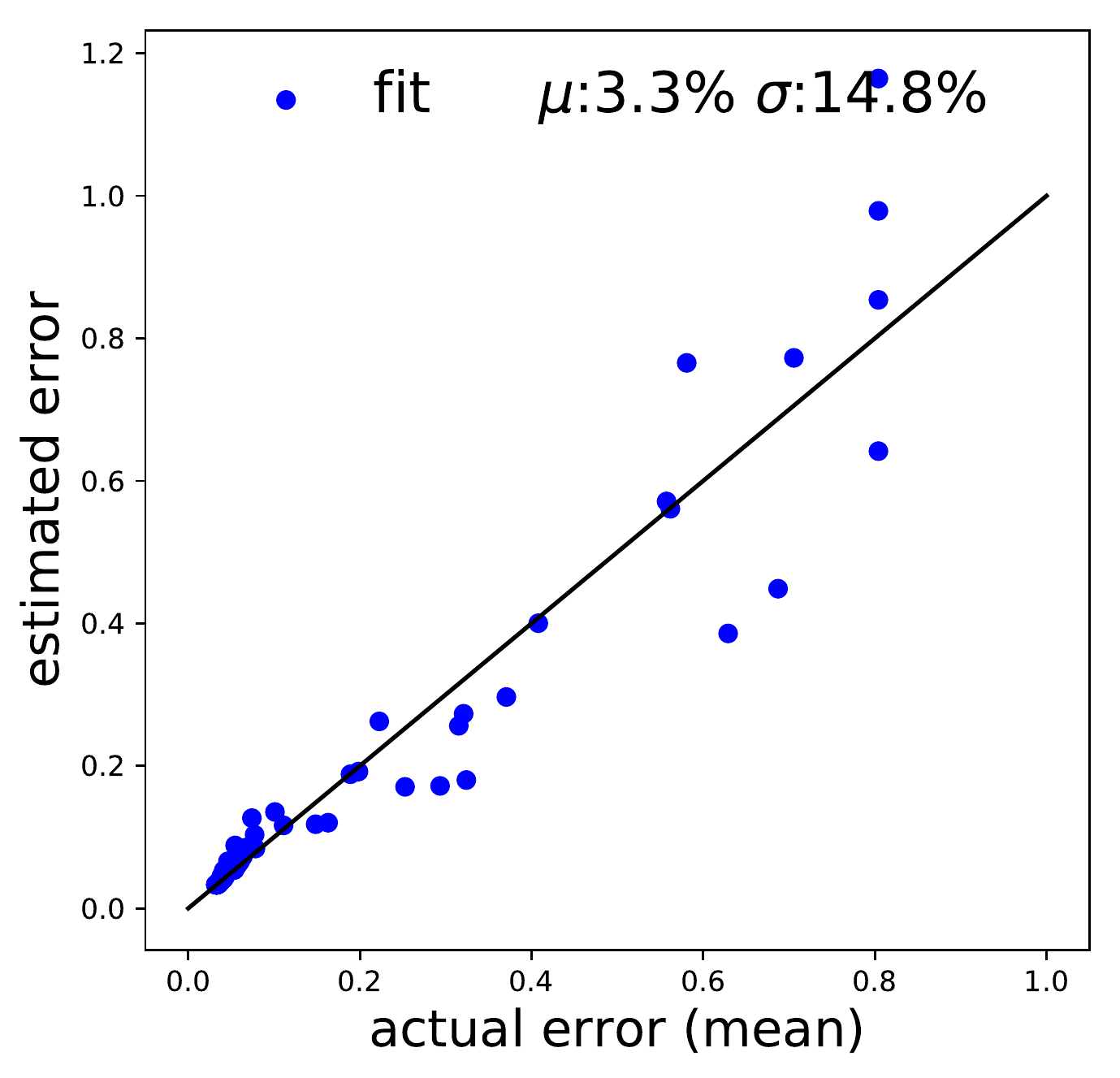}
\end{minipage}
\begin{minipage}{0.3\textwidth}
    \includegraphics[width=\linewidth]{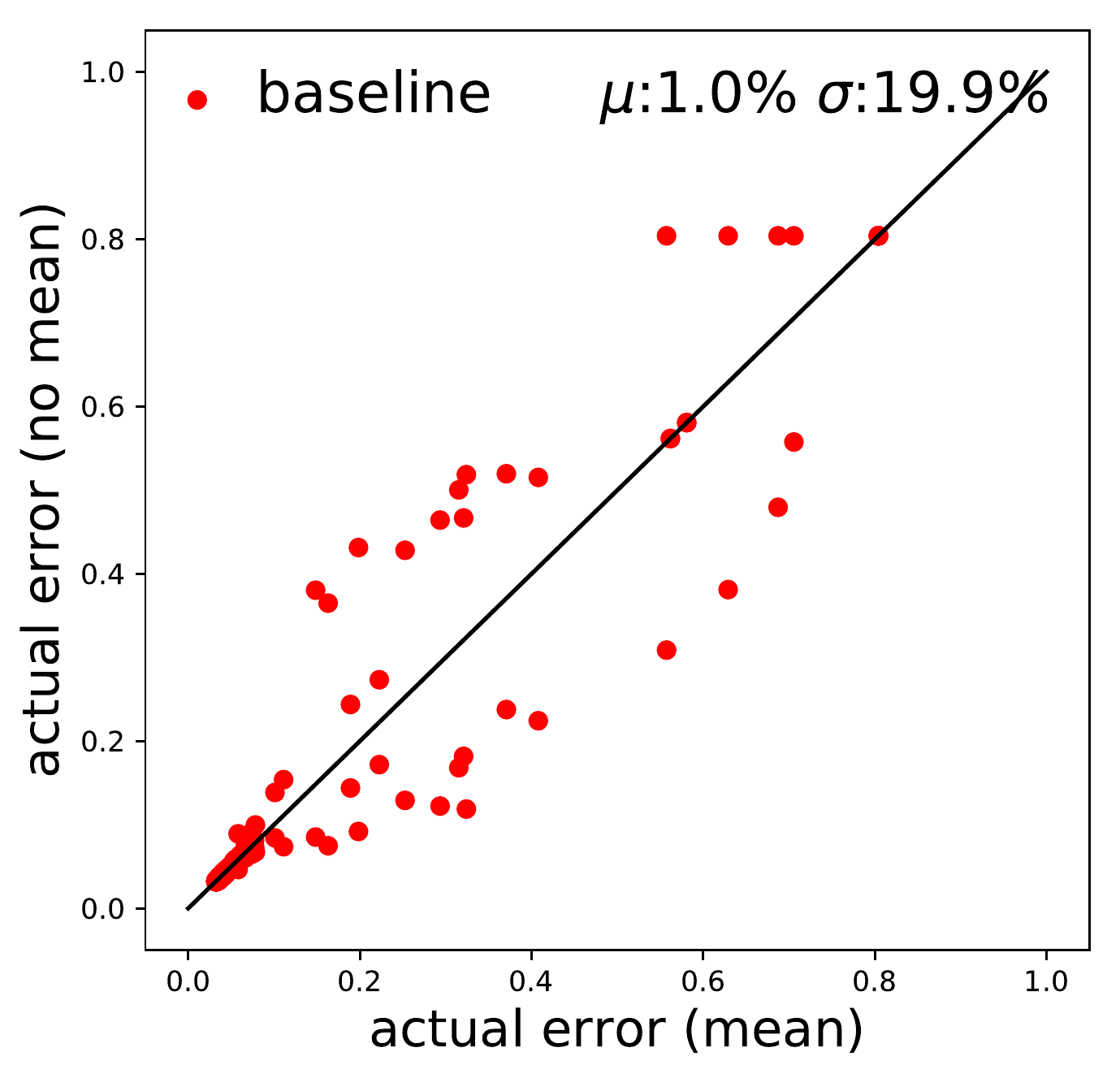}
\end{minipage}
\caption{Fit for VGG-16 on SVHN with IMP pruning.}
\vspace{0mm}
\label{fig:vgg-svhn}
\end{figure}

\begin{figure}[h!]
\centering
\begin{minipage}{0.3\textwidth}
    \includegraphics[width=\linewidth,trim={0.2 0 0 0.65cm},clip]{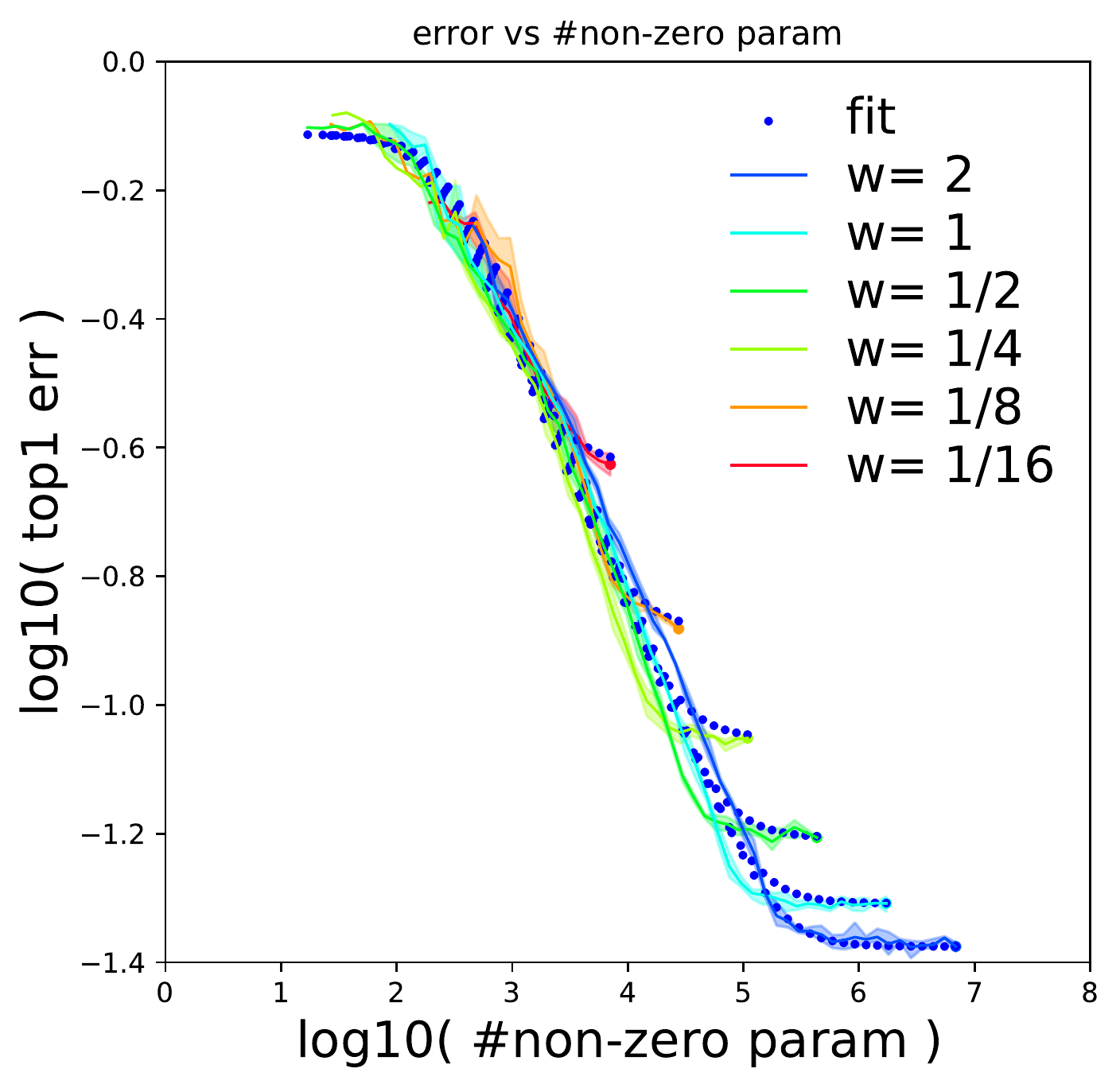}
\end{minipage}
\begin{minipage}{0.3\textwidth}
    \includegraphics[width=\linewidth]{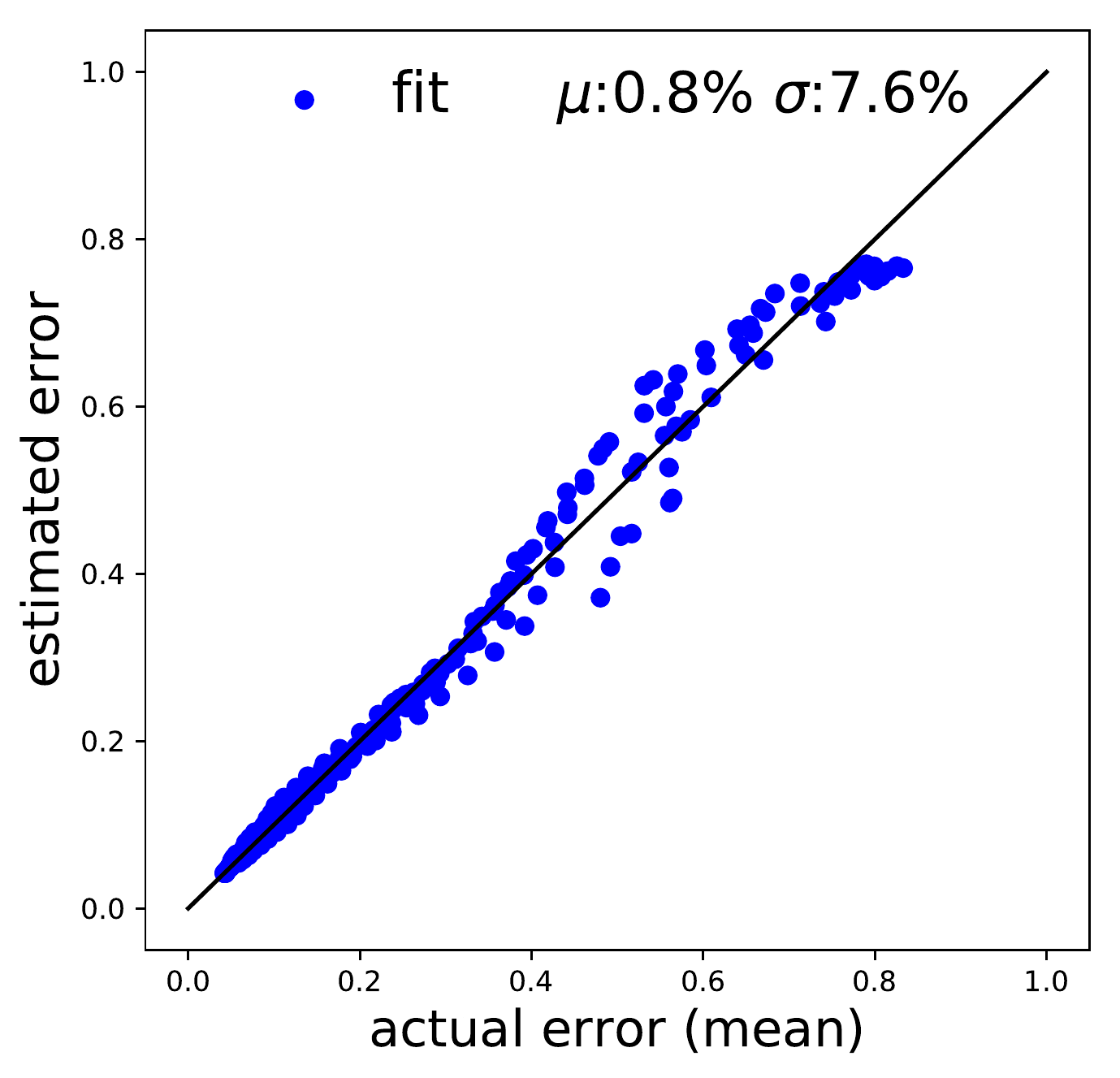}
\end{minipage}
\begin{minipage}{0.3\textwidth}
    \includegraphics[width=\linewidth]{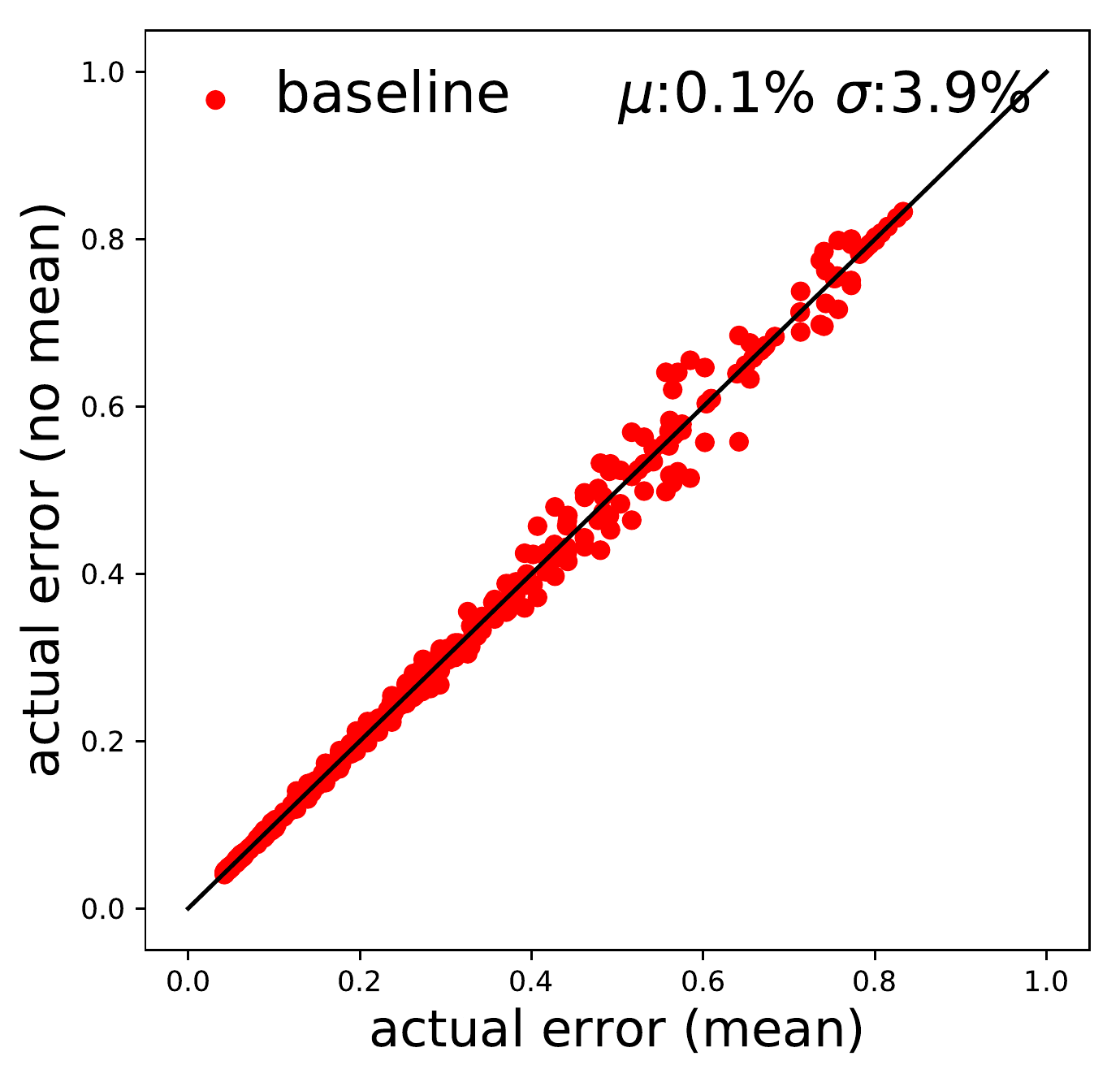}
\end{minipage}
\caption{Fit for DenseNet-121 on CIFAR-10 with IMP pruning.}
\vspace{0mm}
\label{fig:densenet-cifar}
\end{figure}

\begin{figure}[h!]
\centering
\begin{minipage}{0.3\textwidth}
    \includegraphics[width=\linewidth,trim={0.2 0 0 0.65cm},clip]{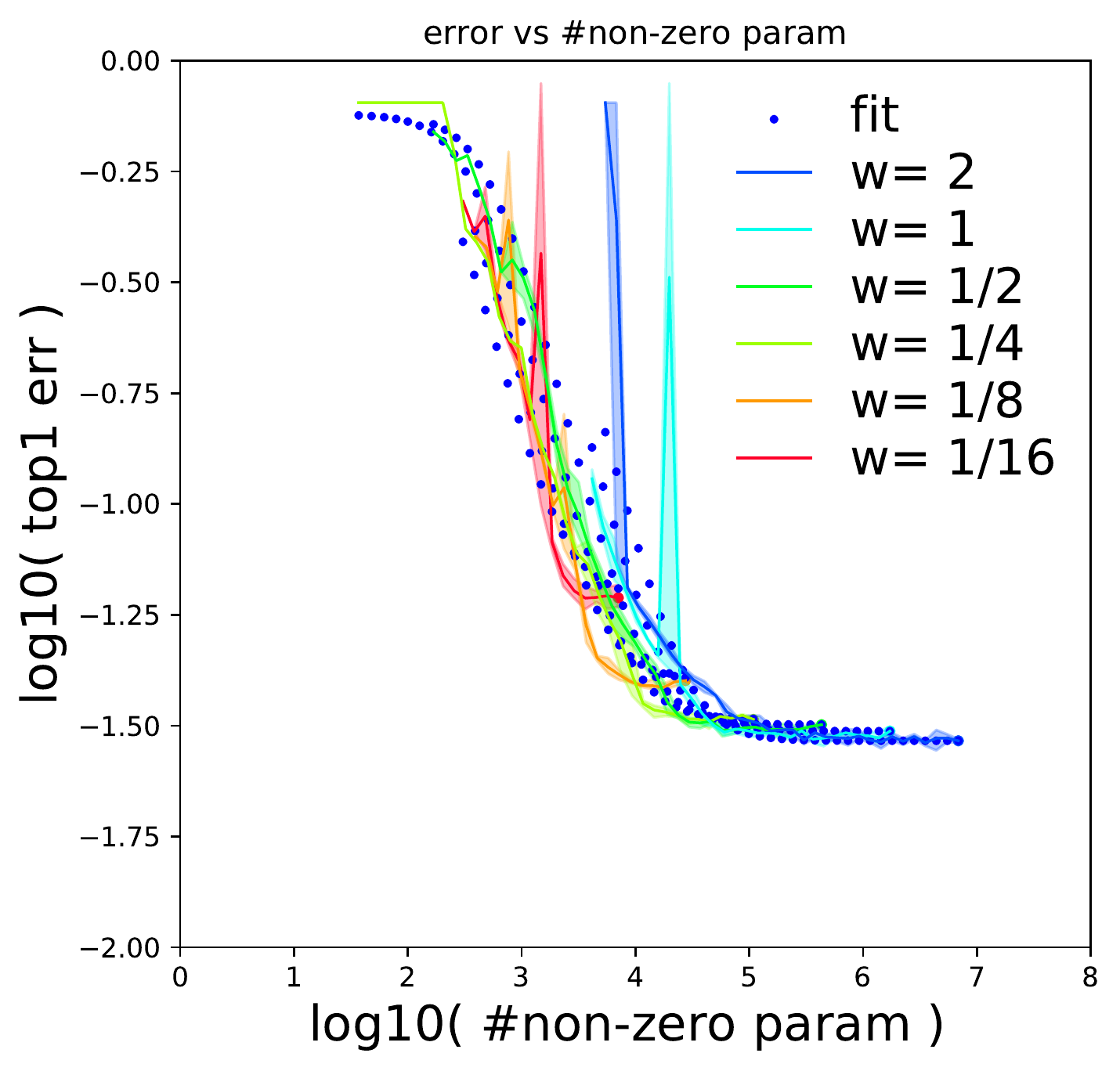}
\end{minipage}
\begin{minipage}{0.3\textwidth}
    \includegraphics[width=\linewidth]{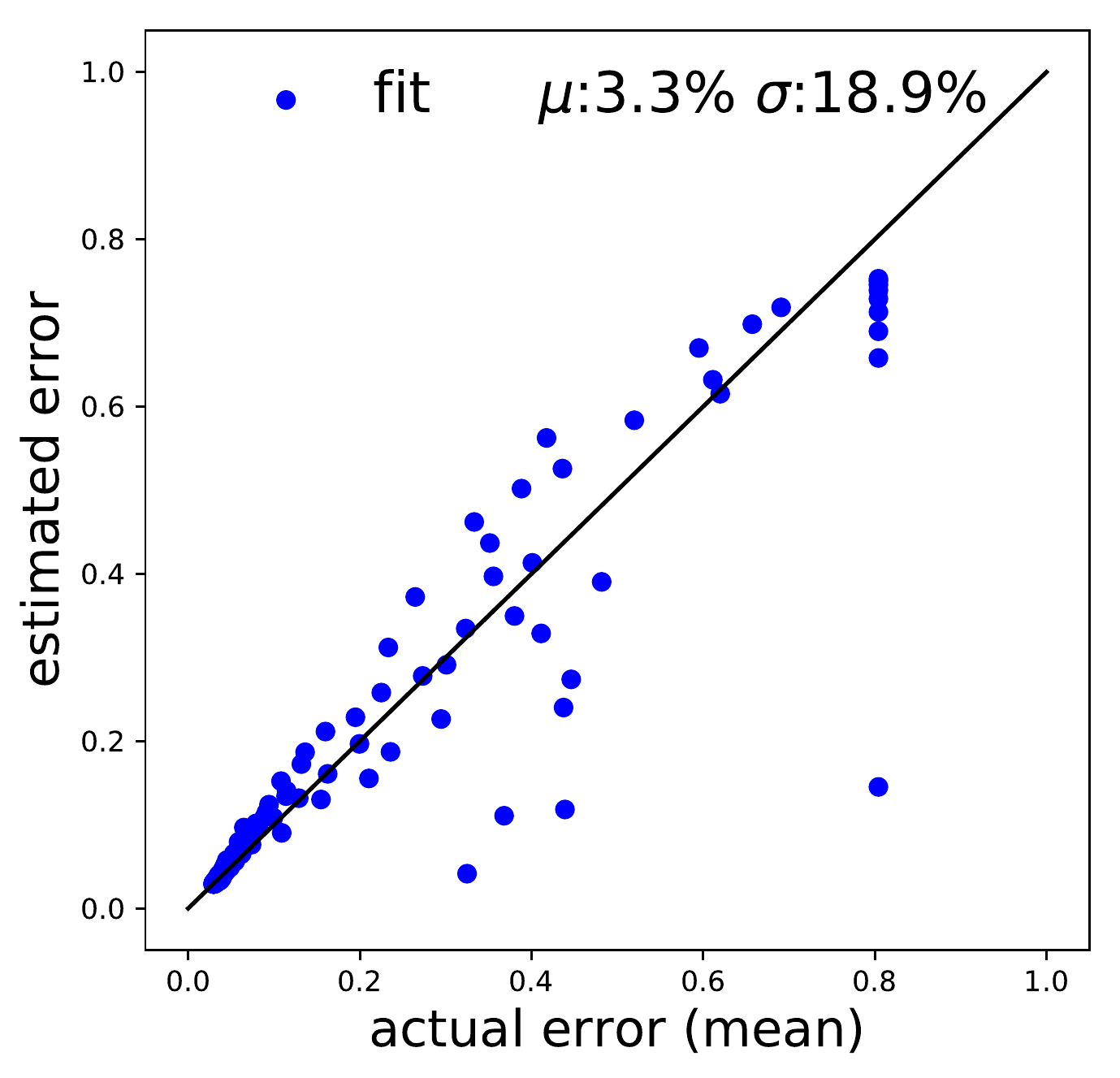}
\end{minipage}
\begin{minipage}{0.3\textwidth}
    \includegraphics[width=\linewidth]{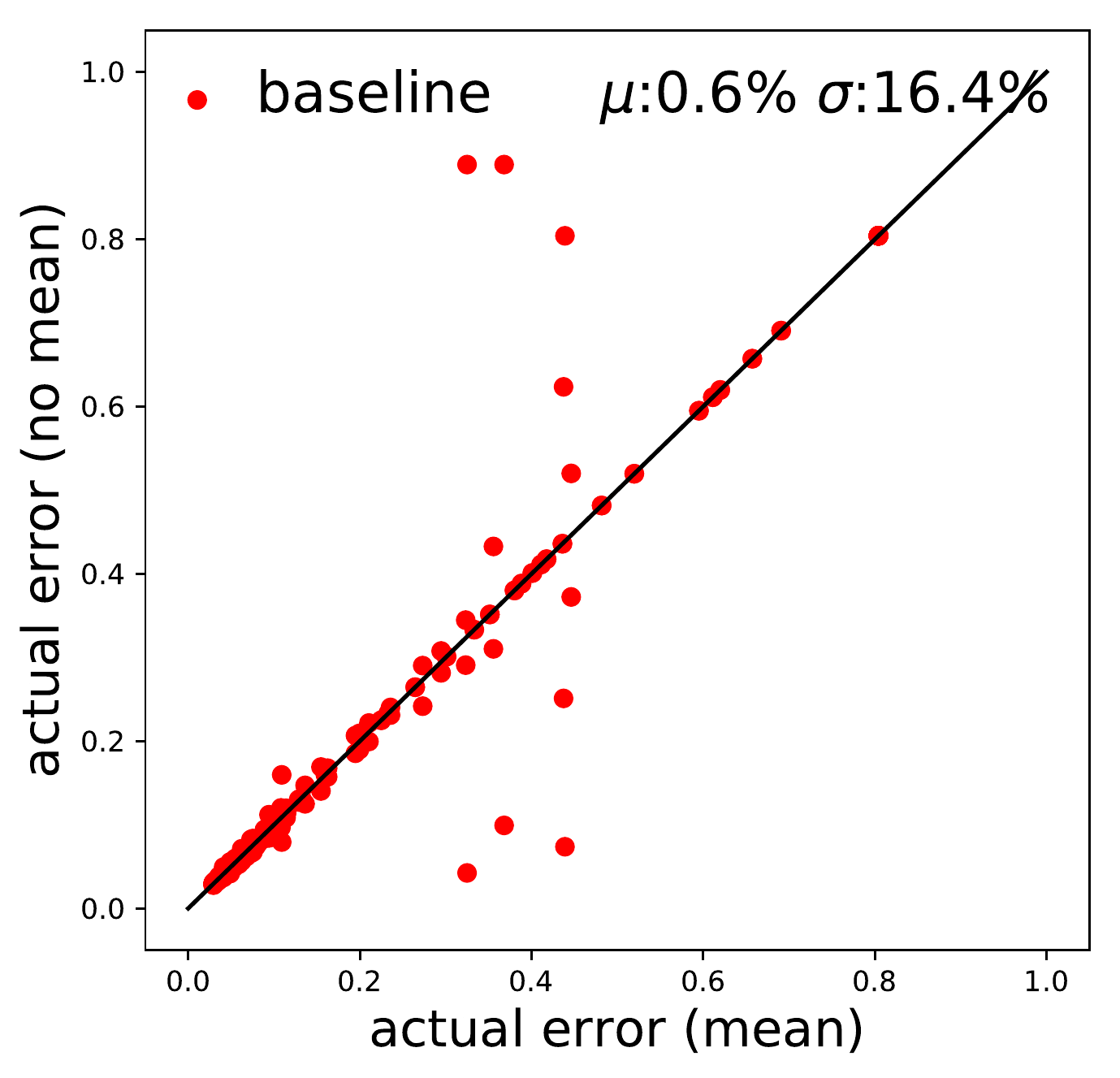}
\end{minipage}
\caption{Fit for DenseNet-121 on SVHN with IMP pruning.}
\vspace{0mm}
\label{fig:densenet-svhn}
\end{figure}

\begin{figure}[h!]
\centering
\begin{minipage}{0.3\textwidth}
    \includegraphics[width=\linewidth,trim={0.2 0 0 0.65cm},clip]{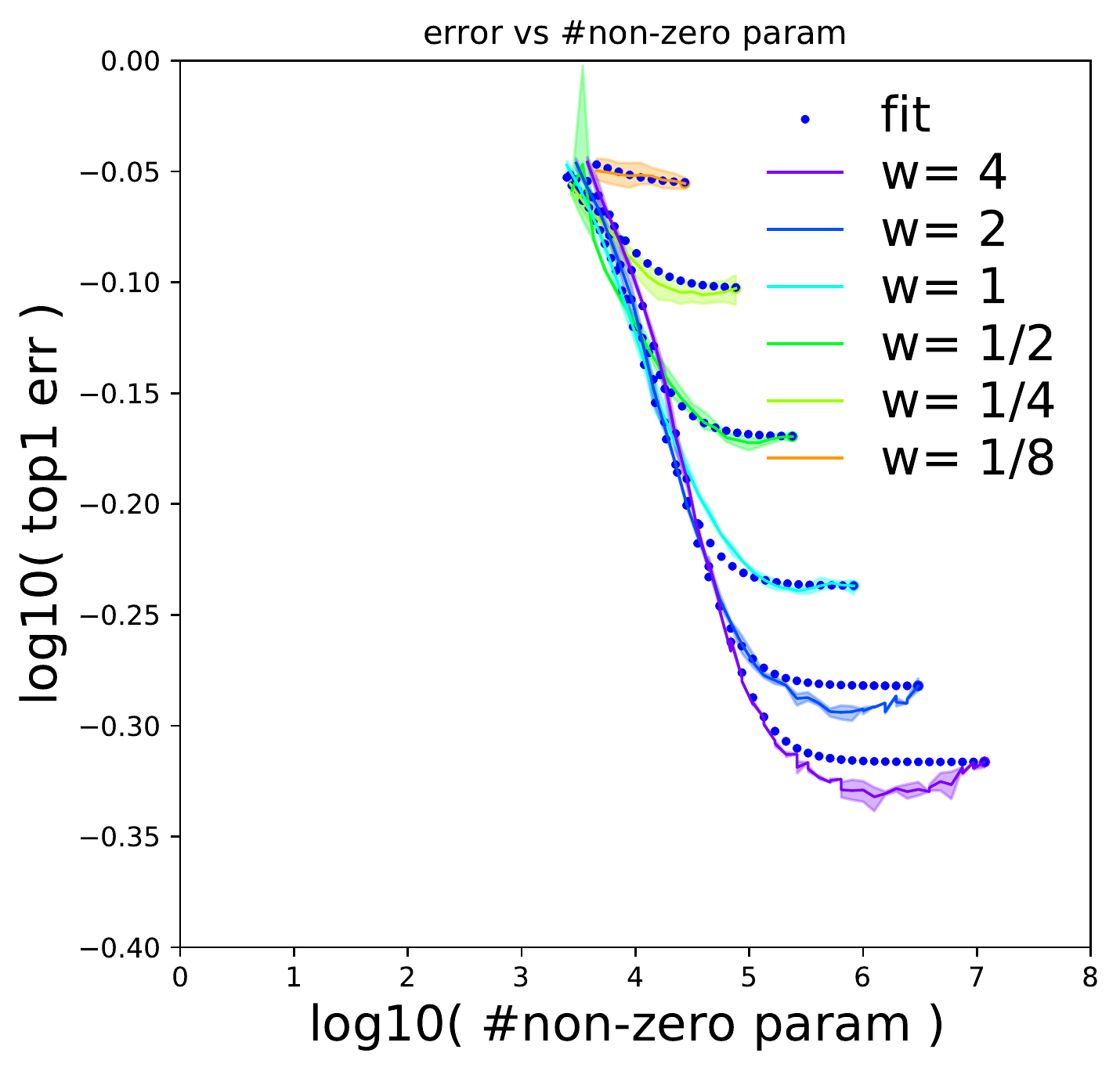}
\end{minipage}
\begin{minipage}{0.3\textwidth}
    \includegraphics[width=\linewidth]{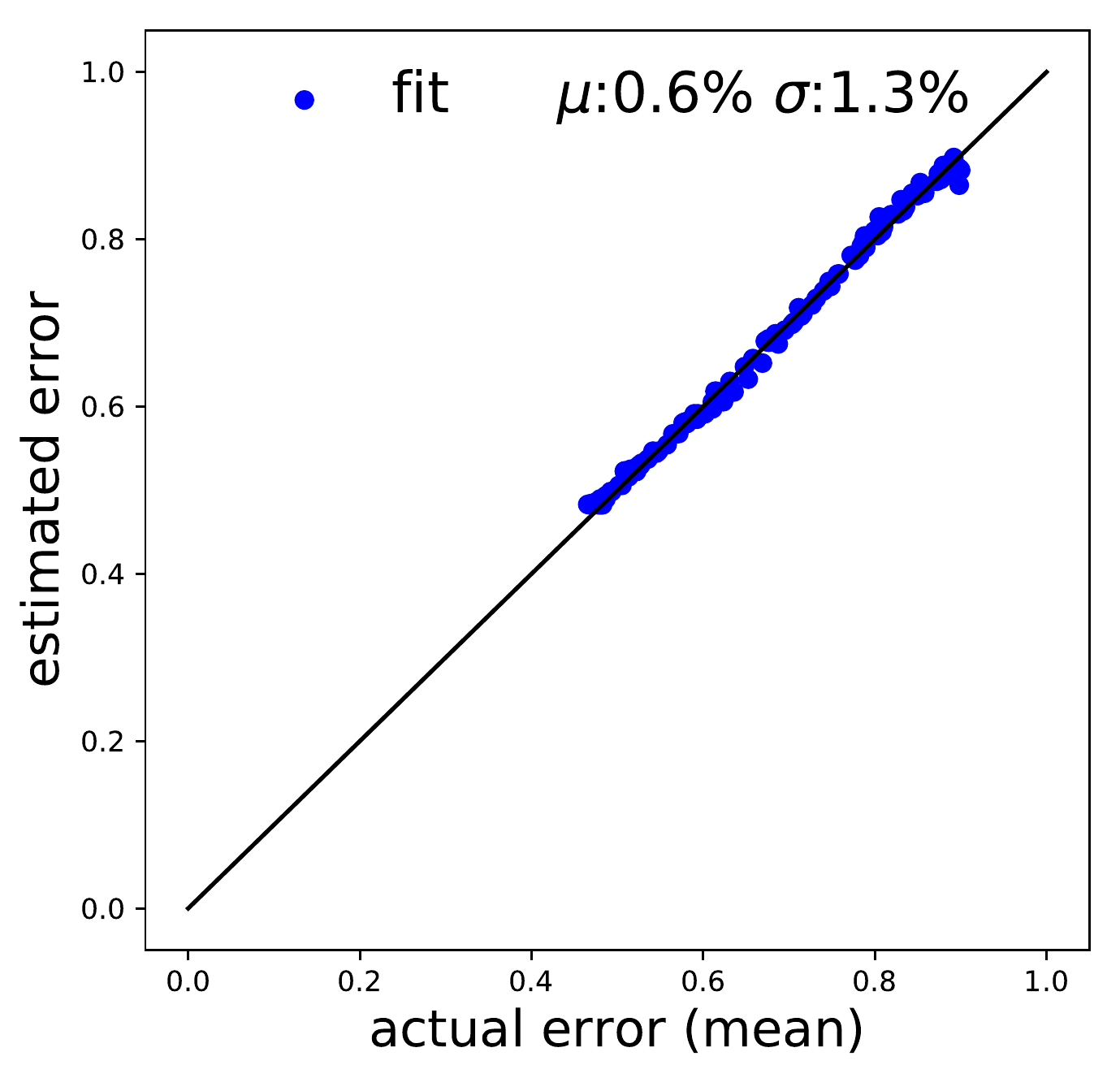}
\end{minipage}
\begin{minipage}{0.3\textwidth}
    \includegraphics[width=\linewidth]{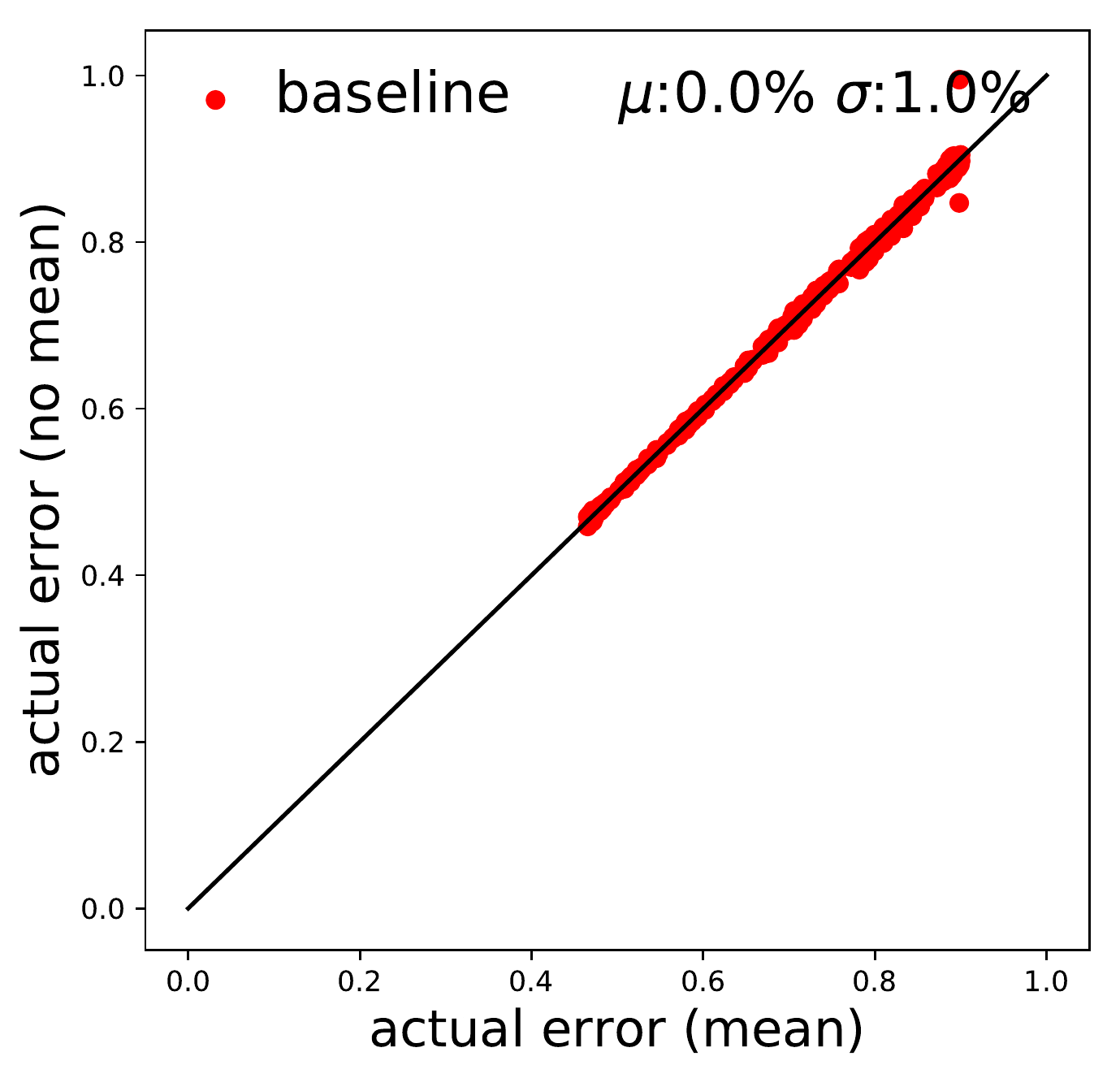}
\end{minipage}
\caption{Fit for ResNet-18 on TinyImageNet with IMP pruning.}
\vspace{0mm}
\label{fig:resnet-imagenet}
\end{figure}

We add these additional comparisons in the following Figures:

\begin{itemize}
    \item Figure \ref{fig:resnet-svhn}: ResNet-20 on SVHN with IMP as width varies. $|\mu|<2\%$, $\sigma<8\%$. 
    Notably, the measurement error in this case is large ($\sigma \sim 9.5\%$), dominating (over the approximation error) the total fit error. The fit averages out some of this error, resulting in a fit error which is lower than the measurement error. In general, experiments on SVHN are quite noisy, leading to significant measurement error.
    \item Figure \ref{fig:vgg-imp}: VGG-16 on CIFAR-10 with IMP as width varies. $|\mu|<3\%$, $\sigma<7\%$ (compared to measurement error 12\%).
    \item Figure \ref{fig:vgg-svhn}: VGG-16 on SVHN with IMP as width varies. $|\mu|<4\%$, $\sigma<15\%$ (compared to measurement error 20\%---measurement error is large at very low densities).
    \item Figure \ref{fig:densenet-cifar}: DenseNet-121 on CIFAR-10 with IMP as width varies. $|\mu|<1\%$, $\sigma<8\%$. 
    Bias is evident in the transition as we discuss in Section \ref{sec:design}.
    \item Figure \ref{fig:densenet-svhn}: DenseNet-121 on SVHN with IMP as width varies. $|\mu|<4\%$, $\sigma<19\%$ (compared to measurement error 16\%---measurement error is large at very low densities).
    \item Figure \ref{fig:resnet-imagenet}: ResNet-18 on TinyImageNet with IMP as width varies. $|\mu|<1\%$, $\sigma<1.3\%$ (compared to measurement error 1\%).

\end{itemize}

% Note that SynFlow suffers from exploding activations on deeper networks, so we do not vary ResNet depth in any of our SynFlow experiments.

\begin{figure}[h!]
\centering
\begin{minipage}{0.3\textwidth}
    \includegraphics[width=\linewidth,trim={0.2 0 0 0.65cm},clip]{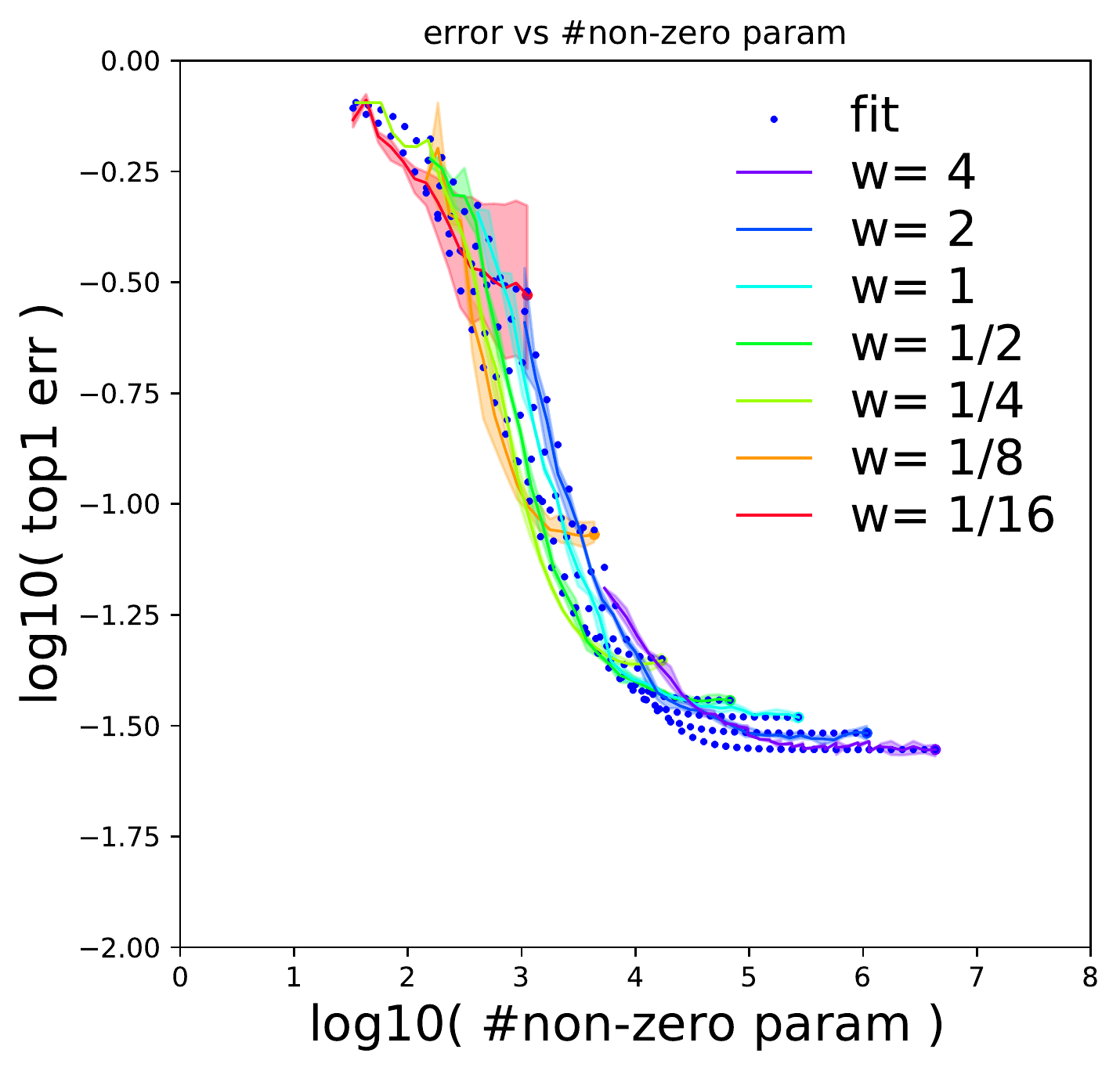}
\end{minipage}
\begin{minipage}{0.3\textwidth}
    \includegraphics[width=\linewidth]{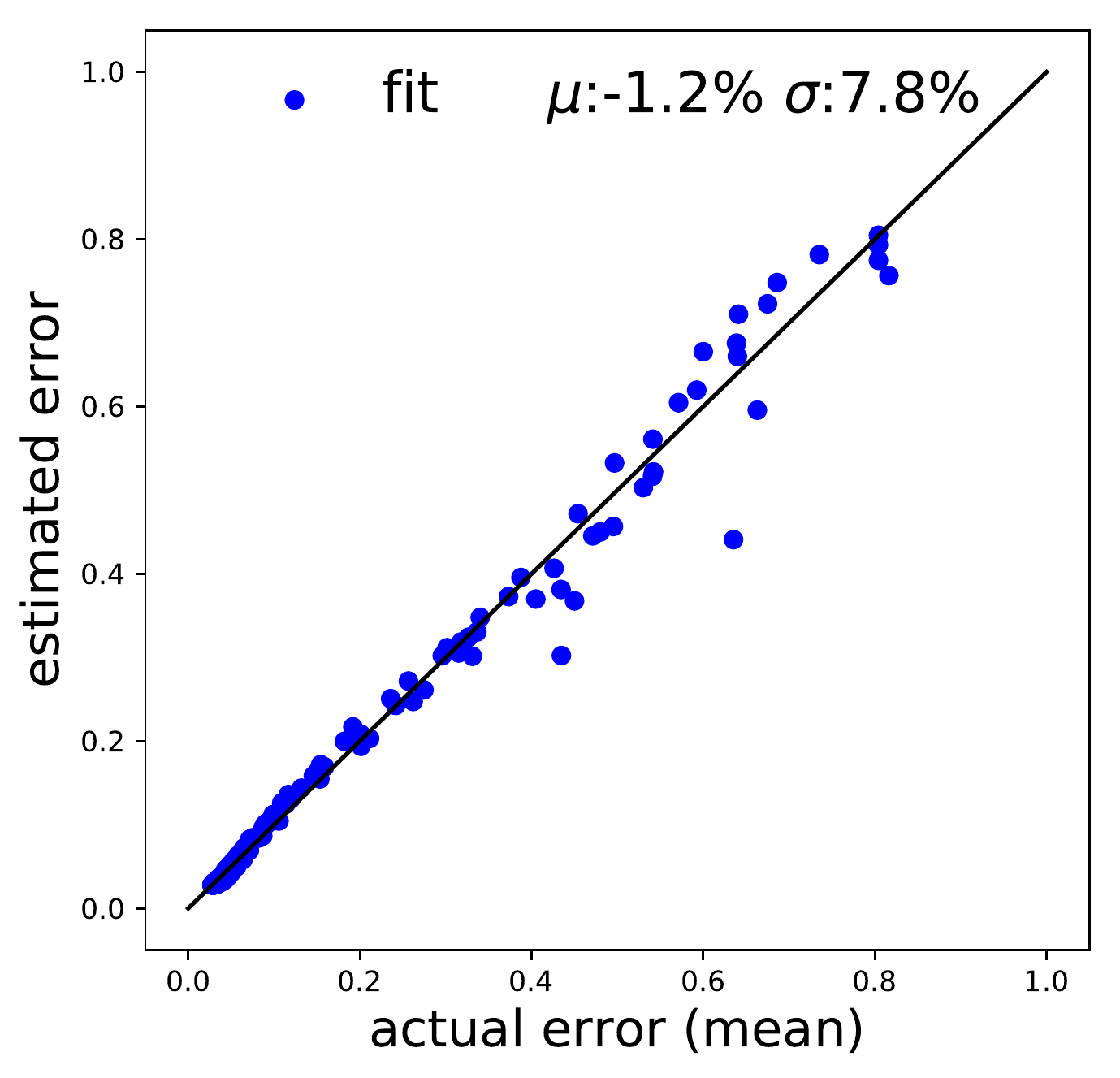}
\end{minipage}
\begin{minipage}{0.3\textwidth}
    \includegraphics[width=\linewidth]{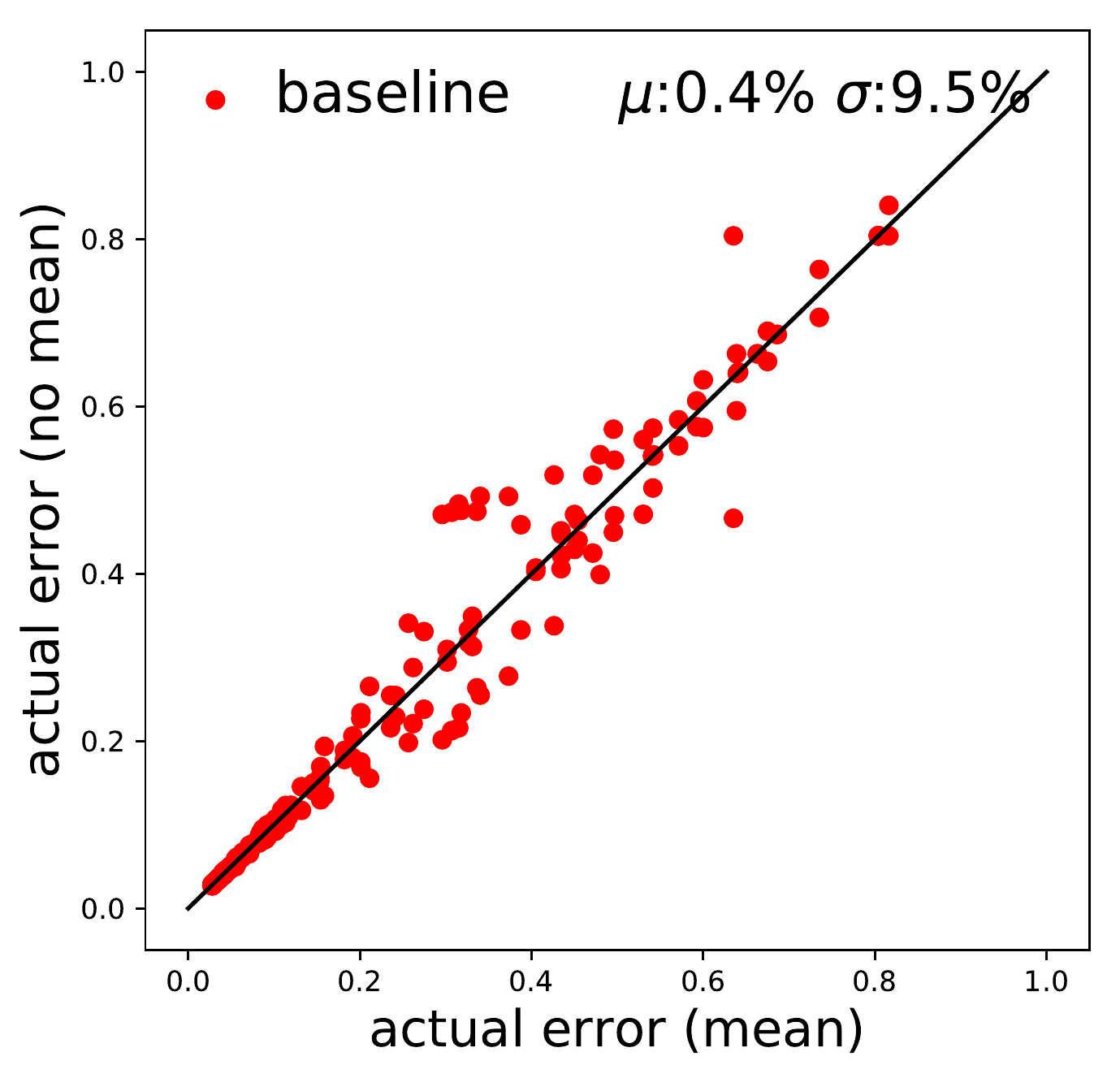}
\end{minipage}
\caption{Fit for ResNet-20 on SVHN with IMP pruning.}
\vspace{0mm}
\label{fig:resnet-svhn}
\end{figure}

% \begin{figure}[h!]
% \centering
% \begin{minipage}{0.3\textwidth}
%     \includegraphics[width=\linewidth,trim={0.2 0 0 0.65cm},clip]{figures/fit_resnet_synflow.pdf}
% \end{minipage}
% \begin{minipage}{0.3\textwidth}
%     \includegraphics[width=\linewidth]{figures/fit_corr_resnet_synflow.pdf}
% \end{minipage}
% \begin{minipage}{0.3\textwidth}
%     \includegraphics[width=\linewidth]{figures/fit_corr_minmax_resnet_synflow.pdf}
% \end{minipage}
% \caption{Fit for ResNet-20 on CIFAR-10 with SynFlow pruning.}
% \vspace{0mm}
% \label{fig:resnet-synflow}
% \end{figure}

% \begin{figure}[h!]
% \centering
% \begin{minipage}{0.3\textwidth}
%     \includegraphics[width=\linewidth,trim={0.2 0 0 0.65cm},clip]{figures/fit_vgg_synflow.pdf}
% \end{minipage}
% \begin{minipage}{0.3\textwidth}
%     \includegraphics[width=\linewidth]{figures/fit_corr_vgg_synflow.pdf}
% \end{minipage}
% \begin{minipage}{0.3\textwidth}
%     \includegraphics[width=\linewidth]{figures/fit_corr_minmax_vgg_synflow.pdf}
% \end{minipage}
% \caption{Fit for VGG-16 on CIFAR-10 with SynFlow pruning.}
% \vspace{0mm}
% \label{fig:vgg-synflow}
% \end{figure}

% \newpage

\clearpage
\newpage

\section{Towards Extrapolation}
\label{app:more_extrapolations}
%Good correspondence between the functional form and the generalization error in practice implies predictive power. That initself already allowed us to reason about pruning tradeoffs.
% - e.g. what is the minimal model possible for a target error on a task, given architecture and pruning algorithm, \hl{See further discussion \ref{}}. 

\textbf{Background.}
In the main body, we showed that our scaling law accurately fits the error of pruned neural networks.  As, such it has predictive power, allowing us to reason in a principled manner about pruning trade-offs.
Similarly, it allows to make predictions about what would happen at larger model and data scales than explored here.
Importantly, only a few experiments need be performed to find the coefficients for the scaling law (see Appendix \ref{app:interpolation}).

However, we could ask, how accurately can we estimate the scaling law parameters from even smaller scales?
That is, is it possible to fit our scaling law to data from networks with deliberately smaller depths, widths, and dataset sizes and accurately predict the error of larger-scale models? 
If so, we could make informed decisions about pruning large-scale models through small-scale experiments alone, saving the costs associated with large scale training and pruning.

% When developing a scaling law, one further property is desirable: \emph{extrapolation}.
% In practice, we are especially concerned with extrapolation from small-scale settings to large-scale settings.
% That is, is it possible to fit our scaling law to data from networks with smaller depths, widths, and dataset sizes and accurately predict the error of larger-scale models?
% This would make it possible to make informed decisions about pruning large-scale models through small-scale experiments alone.

%Moreover, one can consider - does \emph{extrapolation} become possible? In other words, now that the scaling law existence and form is known, can one find the scaling law parameters (5 free parameters) by conducting small scale experiments, and subsequently predict large scale pruning performance? 
Outside the context of pruning, the scaling laws of \cite{rosenfeld2020a} (for both language models and image classification) and \cite{kaplan2020scaling} (for predicting the expected performance of GPT-3 \citep{brown2020language} at very large scale) have been shown to extrapolate successfully in this manner.

\begin{figure}[h!]
    \centering
    % \vspace{-2mm}
         \begin{minipage}{0.34\textwidth}
            \includegraphics[width=\linewidth,trim={0.2 0 0 0.65cm},clip]{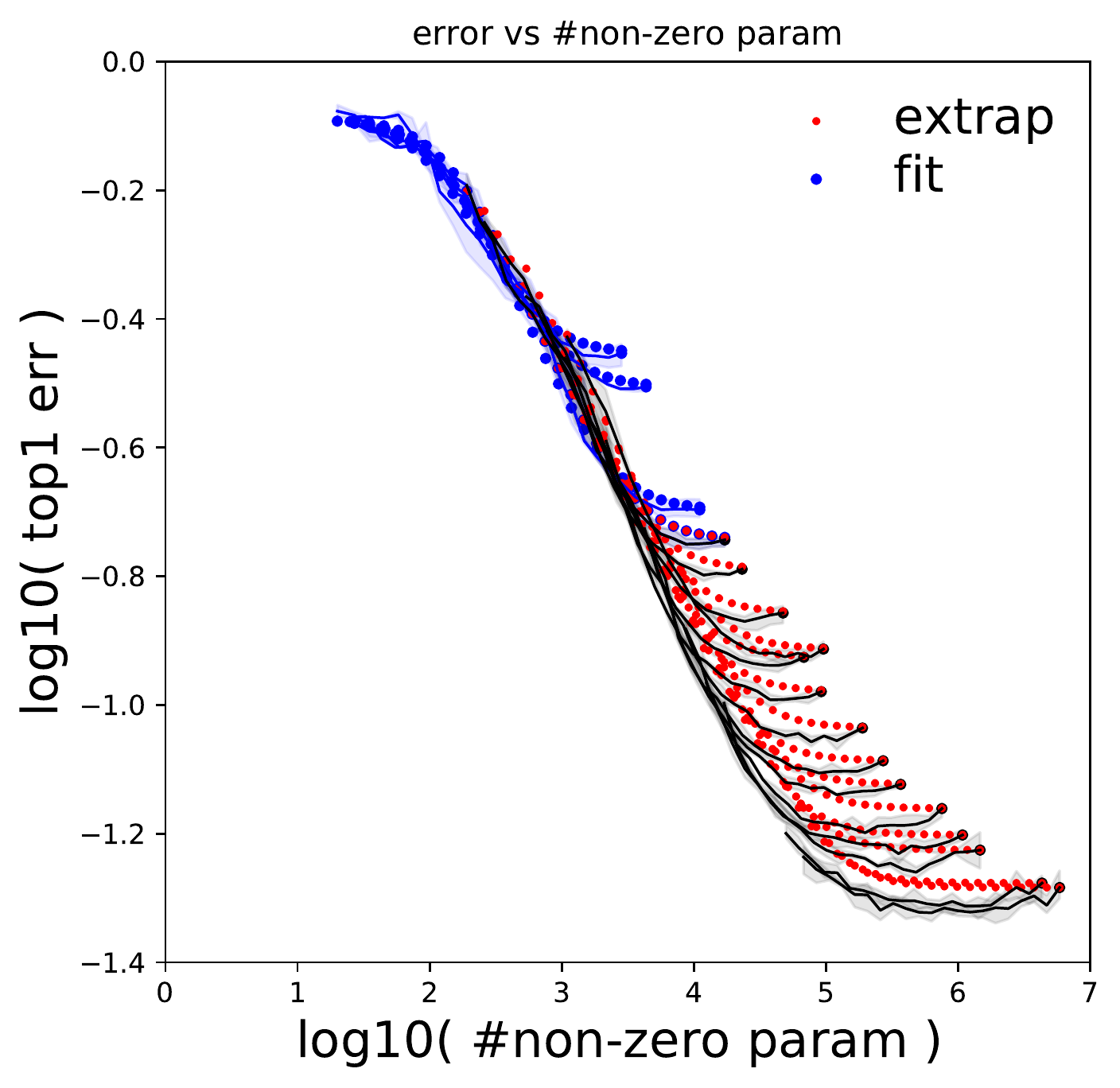}
            \label{fig:cifar_extrap}
        \end{minipage}%
        \begin{minipage}{0.335\textwidth}
            \centering
            \includegraphics[width=\linewidth,trim={0.2cm 0 0 0.3cm},clip]{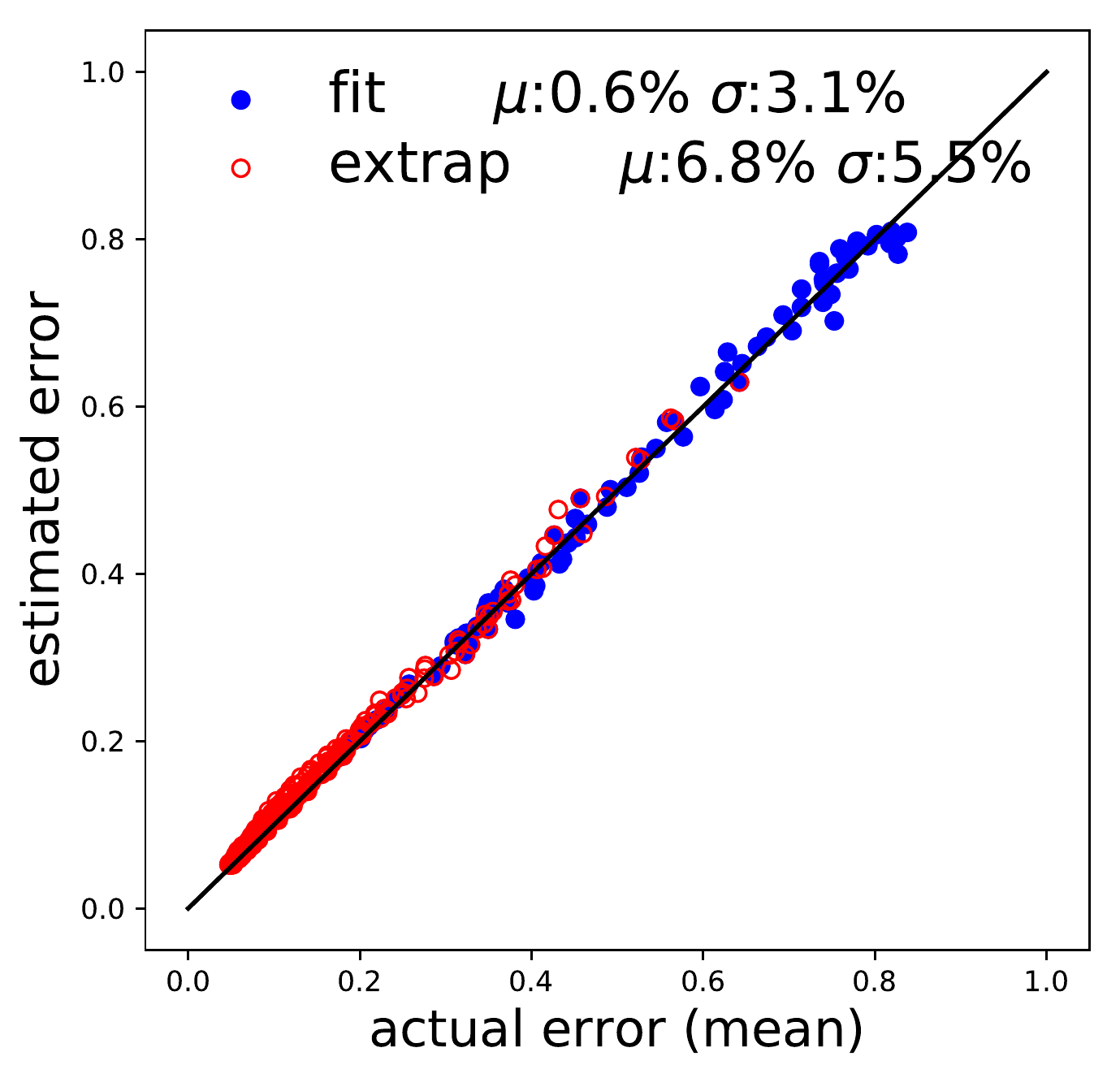}
            \label{fig:cifar_extrap_corr}
        \end{minipage}%

% \vspace{-5pt}
\caption{ Extrapolation results from four pruned networks on CIFAR10 $w=\frac{1}{8},\frac{1}{4}; l=14,20$ to all larger networks $(n=1)$. Fit results are in blue, extrapolation in red, actual in black.  Error versus number of non-zero parameters (left).  Estimated versus actual errors (right). %Insets: Corresponding Estimated versus actual mean errors in linear scale - black line is the identity line between 0 and 1.
% (right) Estimated versus mean actual error for all configurations $(d,w,l,n)$ for CIFAR-10, $(d,w,n)$ for ImageNet. (right column)
}
\label{fig:extrap}
% \vspace{-5mm}
\end{figure}

\textbf{Results on CIFAR-10.}
In Figure \ref{fig:extrap}, we show the result of extrapolating from small-scale networks on CIFAR-10 ($w = \frac{1}{8}, \frac{1}{4}$; $l = 14, 20$) to all widths and depths on CIFAR-10.
Extrapolation prediction is still accurate: $\mu<7\%$, $\sigma<6\%$ (vs. $\mu<1\%$, $\sigma<6\%$ in the main body).

\textbf{Future work.}
% As the value of $\mu$ shows, 
However, extrapolation is particularly sensitive to systemic errors.
Specifically, the transitions and the error dips can lead to large deviations when extrapolating.
For ImageNet, the error dips (especially on small dataset sizes) are especially pronounced, preventing stable extrapolation.
In order to improve extrapolation performance, future work should explore the challenges we discuss in Section \ref{sec:design}: approaches to either model or mitigate these dips and to improve the fit of the transitions.

% However extrapolation is sensitive to systematic (approximation) errors. Namely the transition modeling and error dips, when pronounced can cause large deviations when extrapolating. For Imagenet these deviations (critically, the dips in error at initial densities grows larger with dataset decimation), left unchecked, prevent robust (low sensitivity) extrapolation.
% As such, understanding the origin and modeling/mitigating  the dips as well as refining the transition modeling are important future work needed for making extrapolation robust.

% In principle, a single pruning curve, spanning the three characteristic regions of plateaus and power-law suffices to find three parameters $\gamma, p', \epsilon^\uparrow$. The remaining two parameters $\phi, \psi$, are determined by as little as a pair of pruning curves varying the corresponding dimension. In total a 4 pruning curves. However extrapolation is sensitive to systematic (approximation) errors. In other words, systematic errors translate to errors in the estimation the scaling law parameters when done at the small scale. In particular the transition modeling and error dips, when pronounced can cause large deviations when extrapolating. For Imagenet these deviations (critically, the dips in error at initial densities grows larger with dataset decimation), left unchecked, prevent robust (low sensitivity) extrapolation.
% As such, understanding the origin and modeling/mitigating  the dips as well as refining the transition modeling are important future work needed for making extrapolation robust.

\setlength{\tabcolsep}{0pt}
\renewcommand{\arraystretch}{0.5}

\newpage

% \clearpage
\section{Comparison of Pruning and Non-pruning Scaling Laws}
\label{app:comparison-to-rosenfeld}

In this appendix, we contrast the behavior of the error when pruning with the behavior of the error in the non-pruning setting. \citet{hestness2017deep} show the the error follows a saturating power-law form when scaling data (with both low and high-error plateaus) but does not model them.
\citet{rosenfeld2020a} unify the dependency on data and model size while approximating the transitions between regions; they propose the following form:

\begin{equation} \label{eq:norm_err} 
\vspace{-2pt}
    \tilde{\epsilon}(m,n) = an^{-\alpha} + bm^{-\beta} + c_\infty  
\end{equation}

\begin{equation} \label{eq:envelope}
    \hat{\epsilon}(m,n) = \epsilon_0 \left\Vert \frac{ \tilde{\epsilon}(m,n)}{\tilde{\epsilon}(m,n)-j\eta} \right\Vert 
\end{equation}

where $m$ is the total number of parameters and $n$ is the dataset size. $a,b,\alpha,\beta,c_\infty,$ and $\eta$ are constants, and $\epsilon_0$ plays the role of $\epsilon^\uparrow$ in our notation.

\citeauthor{rosenfeld2020a} model the upper transition---from power-law region to the high-error plateau---by a rational form in a fashion similar to the approach we take.
The key difference is that we consider a power of the polynomials in the numerator and denominator of the rational form, where in Eq. \ref{eq:norm_err} the power is hidden in the term $\tilde\epsilon$.

The biggest difference arises when considering the lower transition (between the low-error plateau and the power-law region).
This transition is captured by Eq. \ref{eq:norm_err}.
Considering either the width or depth degrees of freedom $x \in \{w,l\}$, Eq. \ref{eq:norm_err} can be re-written as:

\begin{equation} 
\vspace{-2pt} \label{eq:adapted}
    \tilde{\epsilon}(x) = b_xx^{-\beta_x} + c_x
\end{equation}

Where $b_x$ and $\beta_x$ are constants and $c_x$ is a constant as a function of $x$ (it is only a function of the data size $n$).

Figure \ref{fig:mismatch_app} (right) shows the error versus depth for different dataset sizes.
In grey is the actual error, while in red is the best fit when approximating the error by Eq. \ref{eq:adapted}.
Qualitatively, one sees that the fit using Eq. \ref{eq:adapted} does indeed closely match the error in practice.

Recall that we are interested in comparing the errors as a function of the density. 
A requirement from any functional form used to model the dependency on the density is to degenerate to the error of the non pruned model $\epsilon_{np}$ at $d=1$.
We adapt Eq. \ref{eq:adapted} by solving the relation between $b_x$ and $c_x$ meeting this constraint, to arrive at:

\begin{equation} 
\vspace{-2pt} \label{eq:adapted_density}
    \tilde{\epsilon}(x) = b_xx^{-\beta_x} + \epsilon_{np}-b_x
\end{equation}

Contrast Eq. \ref{eq:adapted} with the functional form we propose in Eq. \ref{eq:prune_density}, re-written here for convenience:
{
\small
\begin{equation}
\label{eq:prune_density_2}
    \hat{\epsilon}(d,\epsilon_{np}~|~l, w, n) = \epsilon_{np} \left\Vert \frac{d-jp\left(\frac{\epsilon^\uparrow}{\epsilon_{np}}\right)^{\frac{1}{\gamma}}}{d-j p} \right\Vert^\gamma
    \mbox{where } j = \sqrt{-1}
\end{equation}
}

This can be simplified to capture only the lower transition---far enough from the upper transition ($d\gg p$)---to:

{
\small
\begin{equation}
\label{eq:prune_density_3}
    \hat{\epsilon}(d,\epsilon_{np}~|~l, w, n) = \epsilon_{np} \left\Vert \frac{d-jp\left(\frac{\epsilon^\uparrow}{\epsilon_{np}}\right)^{\frac{1}{\gamma}}}{d} \right\Vert^\gamma
\end{equation}
}

Figure \ref{fig:mismatch_app}  (left) shows error versus density for different widths.
In blue is the fit with Eq. \ref{eq:prune_density_3} which follows closely the actual error (black) while in red is the fit with Eq.  \ref{eq:adapted_density} which deviates noticeably in comparison.

\label{app:difference_in_powerlaws}
{

\begin{figure}[H]
    \centering
    \begin{minipage}{0.325\textwidth}
        \includegraphics[width=\linewidth,trim={0 0 0 0.7cm},clip]{figures/pruning_curves_comparison.pdf}
    \end{minipage}%
    \begin{minipage}{0.325\textwidth}
        \includegraphics[width=\linewidth,trim={0 0 0 0.7cm},clip]{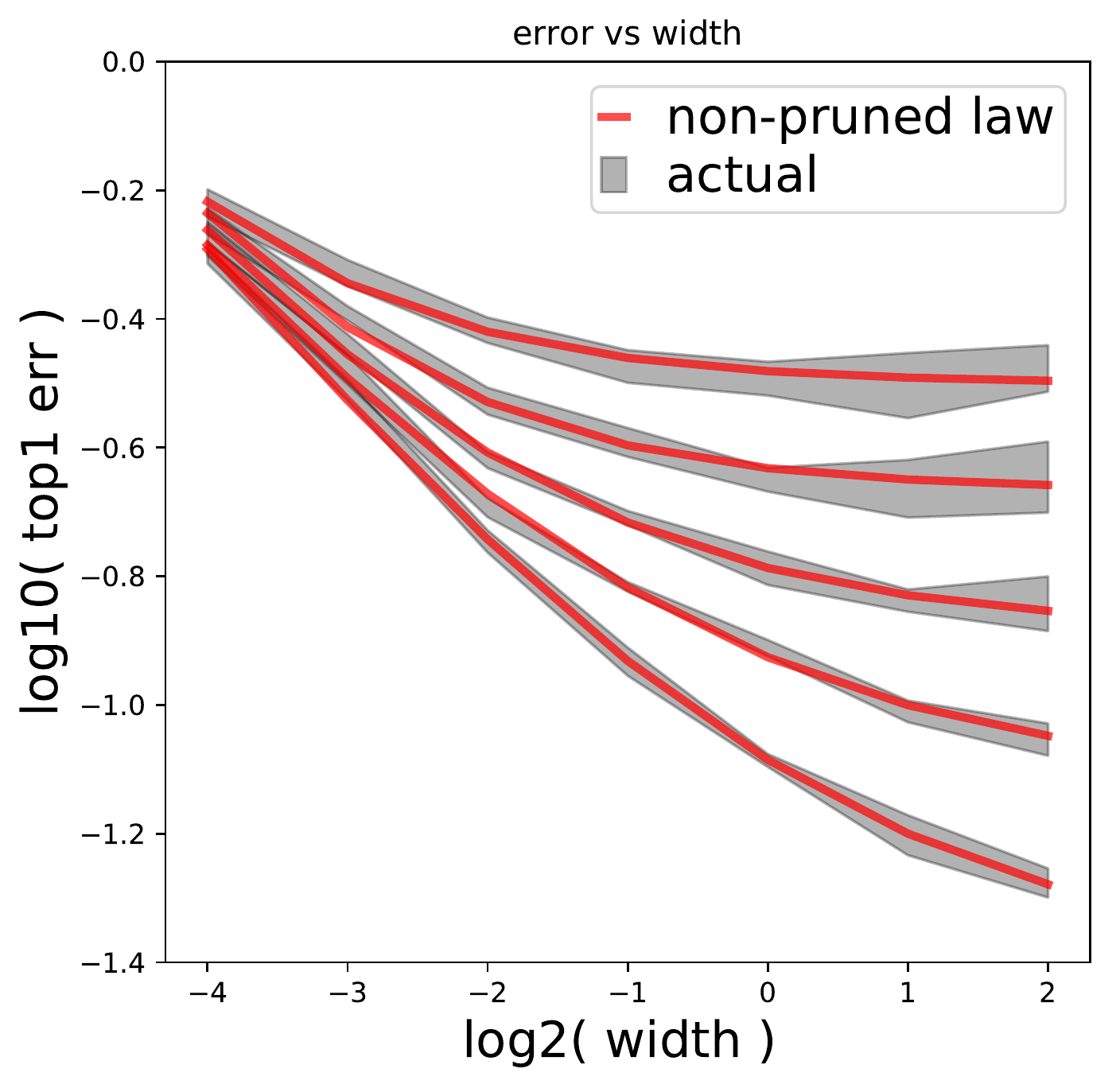}
    \end{minipage}%

    \caption{(Left) Error versus density for different widths. In blue is the fit \eqref{eq:intermediate_state} follows closely the actual error (black) while in red is the fit for the adapted from \citet{rosenfeld2020a} which deviates noticeably in comparison. (Right) error of non-pruned networks versus width for different data, fit shown (solid red) for the non-pruning scaling from  \citet{rosenfeld2020a}.
    }
    \label{fig:mismatch_app}

\end{figure} 

We have seen that in practice that the form of Eq. \ref{eq:adapted_density} does not match well the pruning case, where the mismatch originates from lower transition shape.
We have thus reached a phenomenological observation distinguishing the pruning and non-pruning forms; we leave the study of the origins of this phenomenon for future work. 

}

\newpage

\section{The effect of error dips on estimation bias}
\label{app:magic-one-percent}
In this appendix, we consider the effect of the error dips on our estimator as discussed in Section \ref{sec:joint}.
As we mention in that section, when pruning a network, error often dips below $\epsilon_{np}$ during the low-error plateau.

Recall that we find the parameters in our estimator (Equation \ref{eq:intermediate_state}) by minimizing the MSE of relative error $\delta$.
Our estimation has bias if $\mathbb{E}\left(\hat\epsilon -\epsilon\right) \neq 0 $ where the expectation is over all model and data configurations. Equivalently, the relative bias is $\mu \triangleq \mathbb{E}\delta = 0$ iff the estimator is unbiased. 
The Estimator captured by the joint form in Equation \ref{eq:intermediate_state} is a monotonically increasing function of the density.
It is also constrained such that at density $d=1$ it is equal to the non-pruned error $\epsilon_{np}$. 
It thus, can not reduce The MSE to zero, as it can not decrease to match the actual error dips. This results in the bias of the relative error $\mu$ which in practice is $\sim 1\%$.

\end{document}